\documentclass[12pt,a4paper]{book}

\usepackage{listings}
\usepackage{hyperref}
\usepackage{calc}   %
\usepackage{xcolor} %
\usepackage{ifthen}
\usepackage{fancyhdr}
\usepackage{subcaption}

\usepackage{graphicx}
\usepackage{xxcolor}
\usepackage{url}
\usepackage{bbm}
\usepackage{amsthm}
\usepackage{amssymb}
\usepackage{amsmath}
\usepackage{amsfonts} 
\usepackage{amsbsy}
\usepackage{fancyvrb}
\usepackage{lipsum}  

\usepackage{algorithm,algpseudocode}
\usepackage[printonlyused]{acronym}
\usepackage{natbib}
\usepackage{latexsym}
\usepackage{booktabs} 
\usepackage{tabularx}
\usepackage{colortbl}
\usepackage{tikz}
\usepackage{svg}
\usetikzlibrary{shapes,arrows,petri,calc,shadows}
\usetikzlibrary{decorations.pathreplacing}
\usetikzlibrary{snakes,automata,backgrounds,positioning}
\usepackage{tikz-dependency}
\usepackage{xspace}
\usepackage{rotating}
\usepackage{pifont}
\usepackage{times}
\usepackage{comment}
\usepackage{multirow}
\usepackage{multicol}
\usepackage{pgfplots}
\usepackage{enumerate}
\usepackage{verbatim}

\usepackage[utf8]{inputenc}
\usepackage{anyfontsize}
\usepackage{gtrcrd}
\usepackage[scale=0.78]{geometry}
\usepackage{footnote}

\newcommand{\attr}[1]{{\tt \small #1}}

\ifnum \thechapter>0
\fi
\fancypagestyle{plain}{%
\fancyhead{} 

}

\newcommand*{\cventry}[7][.25em]{%
  \cvitem[#1]{#2}{%
    {\bfseries#3}%
    \ifthenelse{\equal{#4}{}}{}{, {\slshape#4}}%
    \ifthenelse{\equal{#5}{}}{}{, #5}%
    \ifthenelse{\equal{#6}{}}{}{, #6}%
    .\strut%
    \ifx&#7&%
    \else{\newline{}\begin{minipage}[t]{\linewidth}\small#7\end{minipage}}\fi}}

\newcommand*{\cvitem}[3][.25em]{%
  \begin{tabular}{@{}p{\hintscolumnwidth}@{\hspace{\separatorcolumnwidth}}p{\maincolumnwidth}@{}}%
    \raggedleft\hintstyle{#2} &{#3}%
  \end{tabular}%
  \par\addvspace{#1}}

\newcommand{\triple}[3]{(\textit{#1}, \textit{#2}, \textit{#3})}

\newlength{\hintscolumnwidth} 
\newlength{\separatorcolumnwidth}
\newlength{\maincolumnwidth}
\newcommand{\coref}{coref}
\newcommand{\ellipsis}{ellipsis}
\newcommand{\inc}{inc}
\newcommand{\pronoun}{pronoun}

\definecolor{pinegreen}{rgb}{0.0, 0.47, 0.44}

\definecolor{color0}{rgb}{0,0,100} 
\newcommand*{\hintfont}{\mdseries} 
\newcommand*{\hintstyle}[1]{{\hintfont\textcolor{color0}{#1}}}

\DeclareMathOperator*{\argmax}{argmax}

\makeatletter
\newcommand\footnoteref[1]{\protected@xdef\@thefnmark{\ref{#1}}\@footnotemark}
\makeatother

\begin{document}
\thispagestyle{plain}

\begin{titlepage}
\begin{center}
\includegraphics[scale=0.3]{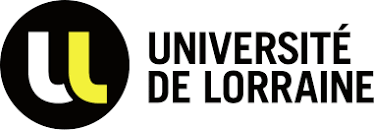}\\
\vspace*{0.8in}
{\LARGE
	\textbf{HDR: Talking to Machines: do you read me?}
   
}
{
    \textbf{Part I: Scientific contributions}
}
\par
\vspace{0.2in}
{\Large
	\textbf{Lina María ROJAS BARAHONA}
}
\vspace{0.8in}
\par
{\Large MEMOIRE}\\
{\normalfont  pour l'obtention de l'}\\
{\Large Habilitation de l’Université de Lorraine}\\\normalfont {(Spécialité Informatique)}\\
\vspace{0.2in}
{\Large Ecole doctorale IAEM Lorraine}
\vfill
\par

\vspace{0.2in}
\bigskip\vfill
  \begin{tabular}{cccc}
	\large  &&& \\
	&&&\\
	\textbf{\large Rapporteurs }&&&\\\\
	\large  Prof. Olivier Pietquin. Université de Lille/Cohere. &&& \\
	\large  DR.  Chloé Clavel. INRIA-Paris. &&& \\
	\large  Prof. Giusseppe Riccardi. University of Trento &&& \\\\
	\textbf{\large Examinateurs}\\\\
	\large Prof. Yannick Toussaint. LORIA/Université de Lorraine. &&&\\
	\large  Prof. Slim Ouni. LORIA/Université de Lorraine. &&& \\
	\large  MC. Jean-Charles Lamirel. LORIA/Université de Lorraine. &&& \\
	\textbf{\large Invités}\\\\
	\large  DR. Claire Gardent. LORIA/CNRS. &&& \\
	\large  CR. Christophe Cerisara. LORIA/CNRS. &&& \\	
	
  \end{tabular}


	
\vfill
\par
\vspace{1.5in}
\par
\end{center}
\end{titlepage}

\frontmatter

\chapter*{}

\textit{To my daughter Sophía, the bright light of my life. To my husband Hervé Vaultrin, who has been always embracing me with his love, support and patience.  To my brother Oscar and my sister Beatriz who have been there for me and gave me my beloved nephews and niece: Alejandro, José Francisco and Maríapaz. To my aunt Beatriz Barahona, who has always stood by my side being my second mother. I also dedicate this work as many others in the past to my parents, Aquiles Rojas and Ruth Barahona, who are always beside me, supporting me and giving me their strength even though they are enjoying the splendid experience of eternity. Thank you all for being the source of my inspiration.}

\textit{To my beautiful country, my beloved Colombia.}
\tableofcontents
\chapter*{List of Acronyms}

\begin{acronym}[DTMFFF]
 \acro{DS}{Dialogue System}
 \acro{TOD}{Task-Oriented Dialogues}
 \acro{DSs}{Dialogue Systems}
 \acro{MDP}{Markov Decision Process}
  \acro{POMDP}{Partially Observable Markov decision process}
 \acro{CLs}{Computational Linguistics}
 \acro{NLP}{Natural Language processing}
 \acro{SLU}{Spoken Language Understanding}
 \acro{NLU}{Natural Language Understanding}
 \acro{DST}{Dialogue State Tracking}
 \acro{NLG}{Natural Language Generation}
 \acro{ASR}{Automatic Speech Recognition}
 \acro{TTS}{Text to Speech}
 \acro{DM}{Dialog Manager}
 \acro{ML}{Machine Learning}
 \acro{QA}{Question Answering}
 \acro{GNNs}{Graph Neural Networks} 
 \acro{IL}{Imitation Learning} 
 \acro{AdaRTE}{Adaptable dialogue architecture and runtime engine.}
  \acro{VoIP}{voice over Internet Protocol}
 \acro{BDI}{Beliefs, Desires and Intentions}
 \acro{LR}{Logistic Regression}
 \acro{SVM}{Support Vector Machines}
 \acro{CNN}{Convolutional Neural Network}
 \acro{LSTM}{Long-Short Term Memory}
 \acro{GNN}{Graph Neural Network}
 \acro{DIP}{Domain Independent Parametrisation}
 \acro{IL}{Imitation Learning}
 \acro{IRL}{Inverse Reinforcement Learning}
 \acro{GNN}{Graph Neural Network}
 \acro{QA}{Question Answering}
 \acro{CoQA}{Conversational Question Answering}
 \acro{CoQAR}{Conversational Question Answering with  Rewriting}
 \acro{KGConv}{Knowledge-base Conversations}
 \acro{KG}{Knowledge Graph}
 \acro{CQA}{Conversational Question Answering}
 \acro{CS4GK}{Converstional Search for General Knowledge}
 \acro{DIANA}{DIalogue in NAtural Language}
 \acro{CBT}{Cognitive Behavioural Therapy}
 \acro{CIFRE}{Conventions industrielles de formation par la recherche}
 \acro{NL4XAI}{Interactive Natural Language Technology for Explainable Artificial Intelligence}
 \acro{COBRA}{Conversational Brains}
 \acro{AMR}{Abstract Meaning Representations}
 \acro{ITN}{Innovative Training Networks}
 \acro{RDF}{The Resource Description Framework}
 \acro{RL}{Reinforcement Learning}
 \acro{LLMs}{Large Language Models}
 \acro{VQA}{Visual Question Answering}
 \acro{RAG}{Retrieval Augmented Generation} 
 \acro{AI}{Artificial Intelligence} 
 \acro{IQ}{Interaction Quality}


 \end{acronym}

\mainmatter
\chapter{Introduction}

\begin{flushright}
\parbox{.8\textwidth}{`Desde entonces no me gané un centavo que no fuera con la máquina de escribir, y esto me parece más meritorio de lo que podría pensarse, pues los primeros derechos de autor que me permitieron vivir de mis cuentos y novelas me los pagaron a los cuarenta y tantos años, después de haber publicado cuatro libros con
beneficios ínfimos. Antes de eso mi vida estuvo siempre perturbada por una maraña de trampas, gambetas e ilusiones para burlar los incontables señuelos que trataban de convertirme en cualquier cosa que no fuera escritor.' 
--- \emph{Gabriel García Marquez, Vivir para Contarla.}}
\end{flushright}

\begin{flushright}
\parbox{.8\textwidth}{`Since then I haven't earned a penny other than with the typewriter, and this seems to me more meritorious than one might think, since the first author rights that allowed me to live from my tales and novels were paid to me when I was in my forties, so many years after having published four books with just tiny benefits. Before that my life was always disturbed by a tangle of traps, tricks and illusions to circumvent the countless lures that tried to turn me into anything other than a writer.' 
--- \emph{Gabriel García Marquez, Living to tell the tale.}}
\end{flushright}

In this dissertation I would like to guide the reader to the research on dialogue but more precisely the research I have conducted during my career since my PhD thesis. Starting from modular architectures with machine learning/deep learning and reinforcement learning to end-to-end deep neural networks. Besides my work as research associate, I also present the work I have supervised in the last years. I proposed four PhD thesis \ac{CIFRE} that Orange accepted to fund. Therefore, I could co-supervise four PhD candidates: Timothy Garwood supervised by Claire Gardent at CNRS, Thibault Cordier supervised by Fabrice Lefevre at the University of Avignon, Sebastien Montella supervised by Alexis Nasr at the university of Aix-Marseille, Léo Jacqmin supervised by Benoit Favre at the University of Aix-Marseille. During 5 years I was head of the industrial research project on dialogue, \ac{DIANA}, which gave me the opportunity of supervising the work of the young researcher Quentin Brabant, other experimented researchers, a developer as well as students in internship and apprenticeship. The deliverables of DIANA project gather open-sourced datasets and neural models as well as scientific publications.

I review briefly the state of the art and highlight the open research problems on conversational agents in Chapter~\ref{c:resondial}. Afterwards, I present my contribution to \ac{TOD} in Chapter~\ref{c:contributionsTOD}, both as research associate and as the industrial supervisor of \ac{CIFRE} theses. I discuss conversational \ac{QA} in Chapter~\ref{c:contributionsConvQA}. Particularly, I present the work of two PhD candidates Thibault Cordier and Sebastien Montella; as well as the work of the young researcher Quentin Brabant. I present the scientific project in Chapter~\ref{c:project}. Finally I present the conclusions in Chapter~\ref{c:conclusion}.



\chapter{A Glance to the Research on Dialogue}
\label{c:resondial}

\begin{flushright}

\parbox{.8\textwidth}{`Please, then,' said Alice, `how am I to get in?'

`There might be some sense in your knocking,' the Footman went on without attending to her [...]' He was looking up into the sky all the time he was speaking, and this Alice thought decidedly uncivil. `But perhaps he can't help it,' she said to herself; [..] --How am I to get in?' she repeated, aloud. 

-[..] the Footman continued in the same tone, exactly as if nothing had happened.
`How am I to get in?' asked Alice again, in a louder tone.

`Are you to get in at all?' said the Footman. `That's the first question, you know.' 

It was, no doubt: only Alice did not like to be told so. `It's really dreadful,' she muttered to herself, `the way all the creatures argue. It's enough to drive one crazy!' 
--- \emph{Lewis Carroll} }
\end{flushright}
\vspace{0.5cm}

The origins of dialogue systems date from 1966 when the Eliza chatterbox was presented \citep{ELIZA}. Eliza was the automated psychoanalyst that let us dream about intelligent systems able to converse as humans. However, Eliza was a simple \textit{template-based} approach, limited by its poor understanding as well as the lack of expressivity and adaptability. The research on dialogue has come a long way since then. Several solutions have been proposed from \textit{regular expressions} and \textit{symbolic approaches} (formal grammars and formal logics)~\citep{Traum00informationstate,McTear2002} to \textit{statistical approaches}~\citep{pieraccini1992speech,williams2007partially,young2010hidden,zhang-etal-2020-dialogpt}, which are data-driven techniques that use either machine learning or deep learning. 

The availability of big data as well as the advances in processing units made deep learning approaches feasible and promising~\citep{cuayahuitl2015strategic, daubigney2013model,vinyals2015neural,sordoni2015hierarchical,wen2016network,serban2017hierarchical}, reviving the dream of creating artificial agents that can easily converse to people.   We have already obtained promising results.  For instance, we know that machines can learn optimal strategies for simple tasks in small domains (\ac{TOD})\citep{weisz2018sample,zhu2020convlab}. Moreover, we are already treating open domain dialogues by asking questions to online encyclopaedias such as Wikipedia (conversational reading comprehension) \citep{choi2018quac, reddy2018coqa} and we are able to predict the best answer in chitchats (end to end neural approaches).  %
The very recent breakthrough ChatGPT~\citep{ouyang2022training} confidently generates apparently coherent responses for a great amount of domains and tasks. A variety of methods has been studied by the research community on dialogue during the last decade. From modular architectures to end-to-end neural networks. In this chapter I will describe some these approaches.

\section{Why is human conversation difficult?}
\label{s:difficulties}

Who has not experienced the frustration of talking to an automatic system in a call centre. Typically, these systems struggle to understand. They are repetitive because they are unable to rectify misunderstandings. Users then must start all over from scratch. In the worst-case users need to call again, then they try hard to fool the system until the call is finally answered by a human. Conversations with automatic systems are unnatural because they do not deal correctly with misunderstandings, they do not adapt to novel situations and they constraint humans' great communication skills.

\paragraph{Dialogue-Acts and discourse obligations:} A dialogue can be seen as a sequence of turns, in which every speaker takes a turn to speak and to contribute to the conversation. The philosopher \citep{austin62} stated that speakers perform actions while conversing and named these actions speech acts. Examples of these actions are : informing, requesting, offering, promising, answering, persuading, convincing, etc. The research community nowadays call these actions Dialogue Acts or Communicative Acts~\citep{bunt}.
Adjacency-pairs are pairs of dialogue-acts in conversation. For example, after a question in a conversation, the speaker is waiting for an answer. After an offer the speaker is waiting for an acceptance or a rejection. These represent discourse obligations in human conversation.


\paragraph{Coreferences and Ambiguity:} Moreover, humans can refer to concepts that were  mentioned previously in the conversation. Indeed, a fluent conversation avoid repetitions. For instance, if you are talking about the president of France, you can say ``Emmanuel Macron" the first time you mentioned him, later you can choose to say \textit{``the president"}. Moreover, if you want to further give your opinion about a recent political proposition he has made, you can say ``I disagree with \textit{his} retirement policy". These are well studied linguistic phenomena that made conversation difficult.

Another aspect is the ambiguity, the same sentence can mean different things in different contexts. ``It's cold in here" can be understood as a request to close the window inside a closed room or it may mean ``It's cool in here" in summer. In winter instead it could mean ``I can't stand the weather; it is too cold".

\paragraph{Grounding: } speakers are always checking that they are following each other. For instance, let us suppose you are receiving instructions about where to place a box in a room. If there are many similar boxes, and the instruction is ``move the box to the right of the desk", you will probably ask ``which one?". Then the instructor will provide more precise information such as ``the yellow rectangular box". This coordination or mutual agreement is known as grounding, mutual knowledge or shared knowledge\citep{clark1991grounding}.  

\paragraph{Planning:} The model Beliefs, Desires and Intentions (BDI) as the primary mental attitudes of an agent was first introduced by Bratman~\citep{Bratman87}. The beliefs are the agent’s model of the world. Desires, in turn, represent how the agent would like the world to be in the future; while intentions are the structured plan the agent has decided to perform. The agent interacts with the world by performing actions and by perceiving aspects of it, including changes which result from its own actions. Perceptions will influence the beliefs of the agent, while actions may change aspects of the world. This model was at the origin of modern \ac{NLU}, in which the aim is to detect user's intentions or \textit{intents}. However, the term intent usually means a dialogue-act with a set of concepts or a combination of them in semantic labels.

All these inherent characteristics in natural dialogue make implementing automated systems a very difficult task.

\section{Preliminary Approaches}


A range of approaches emerged in the history of \ac{DSs}, they were classified in conformity with their \ac{DM}~\citep{allen01towards,churcher97}. According to this classification, ordered by increasing complexity, the simplest of these is the \emph{finite-state scripts}, also called dialogue grammars, followed by \emph{slot-filling}, \emph{plan-based} and \emph{agent-based models}. In a finite-state script the dialogue is represented as a script of prompts for the user. In slot-filling, questions are asked in order to enable the system to fill the necessary slots to perform a task. Conversely, plan-based theories claim that utterances infer acts that are part of a plan, thus, the system tries to identify users' underlying plan, collaborates in accomplishing that plan and responds appropriately. Agent-based models are at the highest level of complexity. They consider planning, executing, and monitoring operations in a dynamically changing world, possibly involving multi-modality. Examples of agent-based models are: the logic-based approaches, which uses inference engines of a higher complexity that in some cases are semi-decidable (in some cases the system will never halt), as well as reinforcement learning approaches, which need a large number of interactions to converge.

\section{Task Oriented Dialogue Systems}
\label{c:intros:tod}
Conversational agents have gained great interest in both academy and industry in the last decades. Typically, available conversational agents have been designed for the task of information-seeking.  These agents act as a natural language interface to a database. First, the system tries to fill the constrains to query a database by inquiring the user. Then it retrieves the items that fulfill users' constraints and finally it communicates the results to the user in natural language. For instance, a person could call the system to check train timetables, she would provide the departure and arrival city as well as the departure date and time. Then, the system would inform her about the available trains. 

\ac{TOD} main goal is to complete a task in collaboration with the user~\citep{pieraccini1992speech, young2002talking,rieser2011reinforcement,young2013pomdp}. Examples of tasks are to search information about a restaurant, to reserve a hotel or to buy train tickets.

\subsection{Definitions}
\label{c:intross:def}

As introduced in Section~\ref{s:difficulties},  \textit{dialogue-acts} or communicative acts are the actions performed by the speakers when uttering sentences (e.g. Informing, Asking, Confirming, Greeting, etc.)~\citep{austin1975things}.  \textit{A domain} is formally defined in an \textit{ontology} as a list of \textit{slots} with their valid \textit{values}. The most common task, the information seeking task, is usually modelled as a \textit{slot-filling} data-query problem in which the system requests constraints to the user and proposes items that fulfil those constraints in a database.  \textit{The action} or \textit{intention} is composed by  a predicate: the dialogue-act, and a set of arguments: the slot-value pairs.  For instance, let us suppose the user has uttered ``I would like a restaurant in the center of town please", this will be translated in the semantic form: $\mbox{inform}(\mbox{type}=\mbox{restaurant}, \mbox{area}=\mbox{center})$. This semantic representation is usually called \textit{flat-semantics} because there is not hierarchy in the concepts of the ontology.

\subsection{Statistical Dialogue Systems}
\label{c:intross:ssds}
One approach to automatic dialogue is to use Reinforcement Learning (RL) to select the system's action~\citep{levin2000stochastic,litman-etal-2000-njfun}. As in a Chess game, a dialogue involves two players, in which each of them takes turns to play. The system should decide its move by considering the environment (the other player's moves) and the rewards is either win or lose the game.  Dialogue is then formulated as an optimisation problem, in which the environment is the user action and the user's feedback is the reward.  The final goal of the system is then to maximise the accumulated reward at long run~\citep{rieser2011reinforcement}. The optimal {\it policy},$\pi$, is a function that takes as argument the current state $s$ and returns the optimal action $a$.

\subsubsection{Markov decision process}
Dialogue can be formalised as a \ac{MDP}, which is a tuple $M=(S,A,T,\gamma,R)$ where:
 \begin{itemize}
  \item $S$: A set of possible states that represent the dynamic environment.
  \item $A$: A set of possible actions.
  \item $T:S \times A \times S \to [0,1]$ is a transition probability function. For any action $a \in A(s)$ taken in a state $s \in S$, the probability of transiting to the next state $s'$ is given by $T(s,s')$.
  \item $\gamma$: A discounting factor in the range of [0, 1], which controls the prediction horizon of the algorithm.
  \item $R$: The reward function that specifies the reward gained at every state. It contains the information that guides the agent towards the goal. $R$ is a function of the state that is bounded in absolute value by $R_{max}$.
 \end{itemize}
A stationary \textit{policy} is a map $\pi: S \to A$ and the discounted infinite-horizon expected reward for starting in state $s$ and following policy $\pi$ thereafter is given by the value function $V^{\pi}(s)$ that
satisfies the following \textit{Bellman Equation}:
\begin{equation}
V^{\pi}(s)= R(s)+\gamma\sum_{s'}T(s,\pi(s),s')V^{\pi}(s') 
\end{equation}
The discounted infinite-horizon expected reward for starting in state $s$, taking action $a$ and following policy $\pi$ thereafter is given by the \textit{Q-function} $Q^{\pi}(s,a)$ that satisfies the following equation:
\begin{equation}
Q^{\pi}(s,a)= R(s)+\gamma\sum_{s'}T(s,a,s')V^{\pi}(s') 
\end{equation}

A policy $\pi$ is optimal in $M$ if, for all $s \in S$:
\begin{equation}
  {\pi}(s) = \operatorname*{arg\,max}_{a \in A} Q^{\pi}(s,a)
\end{equation}
Likewise $Q^*(s,a,\textbf{R})$ is the optimal Q-function of the optimal policy $\pi^*$ for a known reward function $\textbf{R}$.

\subsubsection{Partially Observable Markov decision process}

Dialogues can be modelled as an optimisation problem with \ac{POMDP}s. It simulates the inherent dynamic behaviour of human conversations while deals with the uncertainty of spoken language~\citep{roy2000spoken,williams2007partially, young_pomdp-based_2013}.

A \ac{POMDP} can be seen as a continuous-space Markov decision process (MDP) in terms of policy optimisation where the states are the belief states, which is partially observable. POMDPs have been proposed for spoken dialogue systems because the system is never sure about the user beliefs because of speech recognition errors due to noisy or spoken language disfluencies and hesitations~\citep{roy2000spoken,young2013pomdp}. Since the state is uncertain, it is called the belief state $b(s)$.  An example of a POMDP dialogue system is presented in Section~\ref{cintross:archi}.  The task of predicting the $b(s)$ at a given time $t$ is known as the task of \ac{DST}. The policy learning algorithm receives as input the $b(s)$ and returns the optimal policy $\pi^*$ and a given time.

The belief state $b_t$ is a vector encoding a probability distribution over the different goals, dialogue acts and concepts that are discussed in the dialogue. In the same way, the dialogue action $a_t$ is a vector encoding a probability distribution over the possible agent dialogue actions.

\subsubsection{Hierarchical Reinforcement Learning}
MDP models have been proven to be inefficient for solving complex tasks. These models have trouble overcoming
the cold start problem and/or suffer from the curse of dimensionality~\citep{barto2003recent}. This pattern was also observed with models proposed recently~\citep{mnih2013playing, duan2016benchmarking}. To overcome this issue, ~\citep{parr1998reinforcement} proposed to specify a hierarchy of tasks and to reuse parts of the state space across many sub-tasks, which can greatly improve both learning speed and agent performance.

The notion of temporal abstraction, in which a policy can be decomposed into sub-policies by calling temporally extended sub-tasks was first proposed by~\citep{sutton1999between}. In order to consider hierarchical architectures with temporally extension, we have to generalise the MDP to the semi-Markov Decision Process (SMDP)~\citep{parr1998reinforcement} where actions can take a variable amount of time to complete. This creates  a  division between primitive actions that span over only one action and composite actions that involve an execution of a sequence of primitive actions. This introduces a policy $\mu$ over options that selects option $o$ in state $s$ with probability $\mu(s,o)$, $o$'s policy might in turn select other options until $o$ terminates and so on.  The value function for option policies can be defined in terms of the value functions of the flat Markov policies~\citep{sutton1999between}.

\citep{cuayahuitl2009hierarchical} was the first to propose Hierarchical Reinforcement Learning (HRL)  based on the MAXQ algorithm for dialogue decomposition, making use of hierarchical abstract  machines~\citep{parr1998reinforcement}. However, the tabular approach of this algorithm prevents the efficient approximation of the state space and the objective function. To overcome this limitation ~\citep{budzianowski2017sub}, uses Gaussian process, which provides uncertainty estimates which can be used to speed up learning and achieve more robust performance. 
Some recent work such as \citep{tang-etal-2018-subgoal} aims to  discover automatically sub-goals hierarchy in dialog.

\subsubsection{Reward Functions for Dialogue Systems}
\label{rwfordial}
Previous works on RL for learning dialogue strategies typically use reward functions that penalise long dialogues, returning a final positive reward for task completion or user satisfaction~\citep{levin2000stochastic, litman-etal-2000-njfun, roy2000spoken, young2010hidden, rieser2011reinforcement}. This might be an intuitive reward function for slot-filling applications, such as train ticket or restaurant reservation, 
in which usually customers know exactly what they want, and they expect to be accurately informed by the system as fast as possible. 

However, this reward function might be inappropriate in other situations or for distinct users. For instance, a user might want more advice without caring about the duration of the call. This is especially true in tutorial dialogues, where learners usually have to complete a task and may not know exactly how to do it. 

\subsubsection{Architecture}
\label{cintross:archi}
\begin{figure}[h!]
	\centerline{\includegraphics[scale=0.45]{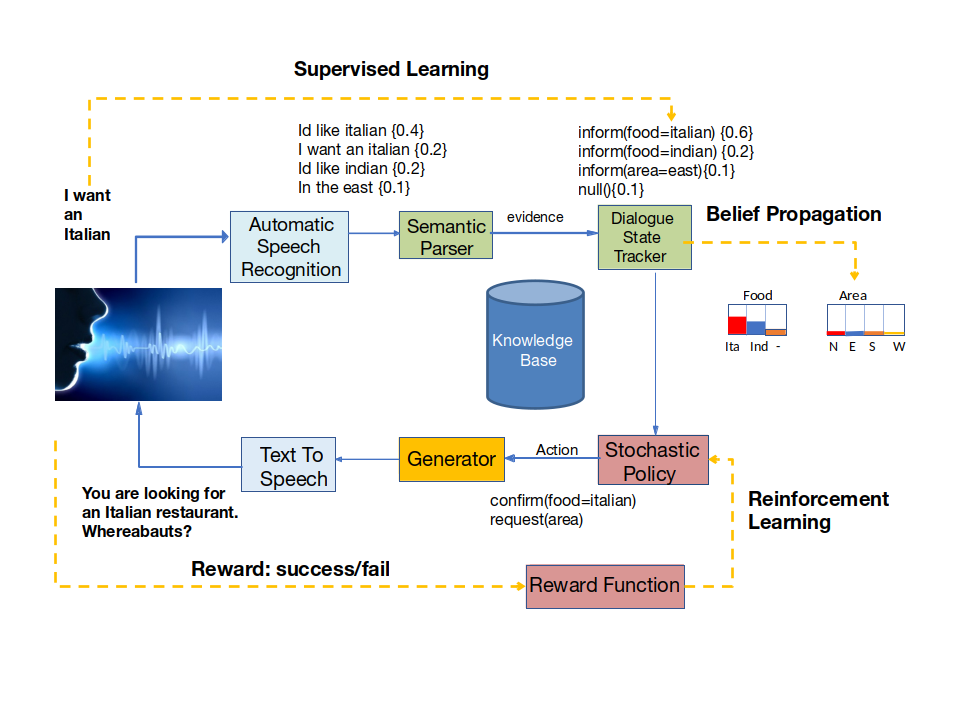}}
	\caption{\small\noindent Basic elements of a statistical spoken dialogue system}
	\label{f:achictecture}
\end{figure}

The basic elements of a RL based spoken statistical dialogue system are shown in Figure~\ref{f:achictecture}. The words recognised by the speech recognition are converted to an abstract representation (the user dialogue acts) by the semantic parser, also known as semantic decoder or \ac{NLU}.  These user dialogue acts are then processed by a belief-state tracker which maintains
a {\it dialogue state} $s$.  This is typically a set of variables denoting the {\it slots} that the system must fill-in to complete the user's goal. For example, in a restaurant information system the slots might 
be \attr{food} for the type of food offered and \attr{area} for the location, and the state $s$ might record the current value and confidence level of each slot. From the state, a {\it belief state} $b$ (usually just a sub-set of the state vector) is extracted and an action $a$ is decided based on a dialogue {\it policy}.  The set of possible actions will include requesting new slot values, confirming already filled slot values and accessing the application for information. Once the appropriate action is determined, it is converted to a textual message $m$ and then rendered by a speech synthesiser.

\vspace*{2mm}

Deep Learning (DL) has been used to implement dialogue components such as semantic decoder or \ac{NLU}~\citep{rojasbarahona2016exploiting}, belief tracker~\citep{daubigney2013model,mrkvsic2016neural} and generator~\citep{wen2015semantically}. Deep Reinforcement Learning is used for policy learning (or \ac{DM})~\citep{cuayahuitl2015strategic}, such as deep q-network (DQN)~\citep{casanueva2017benchmarking}, the actor-critic algorithm~\citep{su2017sample} and actor-critic with experience replay (ACER)~\citep{weisz2018sample}.  Benchmarks comparing these algorithms across different domains and different environments have been published in~\citep{casanueva2017benchmarking}. It is worth noting that ACER has been trained on around 1000 action spaces and not on summary actions as the others~\citep{weisz2018sample}. These algorithms are available in the open-source dialogue framework PyDial~\citep{ultes2017pydial}.

\subsection{End-to-End Task-Oriented Systems}
One preliminary approach to end-to-end dialogue is pipeline-based. It replicates the classical dialogue architecture (Figure~\ref{f:achictecture}), with the main difference that each module is a deep neural network. One example of this approach is the seminal work of~\citep{wen2016network}. This neural pipeline was later improved in \citep{wen2017latent} through a latent variable model for learning the distribution of system actions. Although, authors claimed this approach to be end-to-end, modules are not trained jointly into a single learning unit that can be optimised by gradient-based methods such as back-propagation. Conversely, each module is trained separately in a cascade fashion where the outputs of one model are the inputs to the next one.


\section{End-to-End Dialogues}
Three approaches to end-to-end dialogue system are identify: retrieval-based, generative-based and the combination of both. These approaches are truly end to end, which means the model learns through gradient-based optimisation.
\textit{Retrieval-based} approach treats dialogue as an information retrieval problem~\citep{lowe2015ubuntu, wu2018response}, in which there is a set of candidate responses from which one  is selected as system response, here dialogue is evaluated as an information retrieval problem (in terms of precision/recall). \textit{Generative-based} models~\citep{vinyals2015neural,sordoni2015hierarchical, serban2016building, goo2018abstractive} on the contrary use natural language generation to generate the system response and dialogue is evaluated as as a generation problem (in terms of comparison with multi-references for instance, with BLEU score).
Both approaches have been used for chit-chats or casual conversations. Retrieval-based models however have been also used for task-oriented solutions\citep{lowe2015ubuntu}.
The combination of both, can use generation to paraphrase the retrieved answer. Another way is to compare generative and retrieved responses. The interested reader can find more details about recent end-to-end approaches in~\citep{ni2022recent}. 

\begin{figure}[h!]
	\centerline{\includegraphics[scale=0.55]{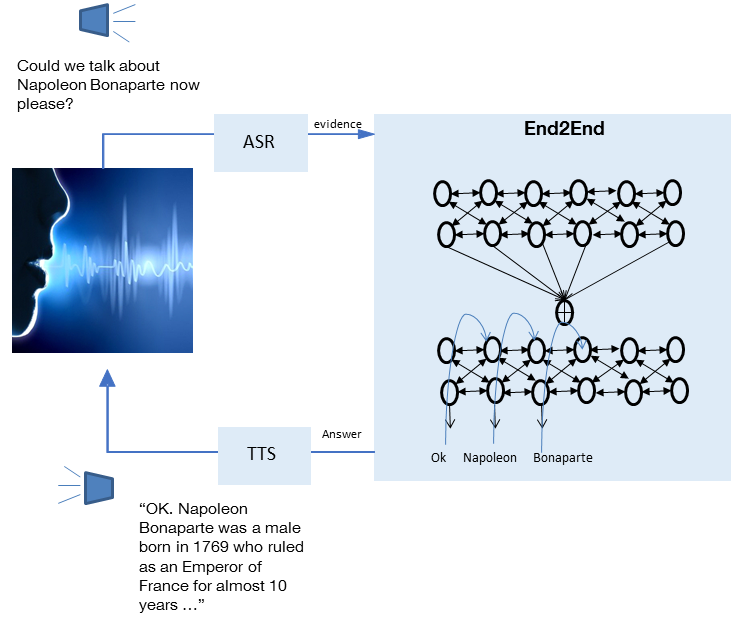}}
	\caption{\small\noindent E2E neural architecture}
	\label{e2earch}
\end{figure}

Yet these approaches neglect the fact that dialogue is dynamic and highly dependent on the environment. Moreover, they do not consider the metrics usually used for evaluating dialogues such as task completion and user satisfaction. Generative approaches usually generate fluent answers, which are sometimes incoherent with the dialogue context. They can generate hallucinations, distortions or repetitions. Retrieval-based approaches are limited to a list of candidate responses, which needs to be created in advance, yielding answers without any context agreement. To overcome these limitations, \citep{Sankar_2019} proposes a combination of reinforcement-learning and generative models. The successful ChatGPT is an example of generative model, carefully trained to follow instructions that also learns to rank its responses.

\section{Pre-trained Language Models}

With the success of Language Models, Deep Learning and Transformers~\citep{vaswani2017attention}, recent solutions proposed pre-trained models with self-supervision. These models outperformed the state-of-the-art in  different \ac{NLP} tasks. They have been initially trained on a large quantity of texts. For instance, BERT has been trained on 800M words and 2,500M words of the BooksCorpus and Wikipedia respectively. GPT-3 is a huge pre-trained model with 175B of parameters, trained on large quantity of data.

Recently, these pre-trained models have been also trained on conversations. For example, BlenderBot~\citep{shuster2022blenderbot} is an encoder decoder which has been trained on 1.5B of Reddit comments. DialoGPT is just a decoder that has trained on 147M of dialogues extracted from Reddit.~\citep{santra2021representation}.  ConveRT has been trained on 727M of dialogs. ChatGPT is presumably the result of finetuning GPT3 with carefully curated instructions with reinforcement learning for correctly rank the generated responses (i.e. learning to rank)~\citep{ouyang2022training}.

The trend nowadays is to initialise deep learning models with a pre-trained one and then fine-tune them for a specialised task. However, fine-tuning with little data will often degrade the initialised weights. Therefore, recent optimisation for fine-tuning are : prompting~\citep{lester-etal-2021-power},  prefixes~\citep{li2021prefixtuning} and adapters~\citep{hu2021lora}.

Despite very promising, the great limitation is that pre-trained models are still static. If there are changes in the language or in the World, these changes will not be reflected in the model. Think about the BERT models pre-trained before the Covid pandemic, they do not contain any representation for the bunch of new vocabulary that emerged during the pandemics: Covid-19, Moderna, Pfizer, sanitary vaccination pass, test PCR, etc. It is worth noting that training these large models is computationally costly and they require Hyper-performance Computing (HPC). A research path is how to keep these models up to date avoiding catastrophic forgetting and reducing the carbon impact necessary during training. 

\section{Conversational Question Answering}
\label{c:intros:cqa}

Research on conversational question answering has gained increasing interest \citep{saha2018complex,reddy2018coqa,choi2018quac}. It consists in sequences of question-answer pairs related to a document or to a knowledge graph.
Complex sequential question answering (CSQA) contains open-domain conversations that treat complex linguistic phenomena such as co-references, ellipses, incompleteness (or under specification) as well as logical, comparative and quantitative reasoning~\citep{saha2018complex}. Two corpora containing discussions about a paragraph of a Wikipedia document have been made public, namely question answering in context (QuAC)~\citep{choi2018quac} and a conversational question answering challenge (CoQA)~\citep{reddy2018coqa}. In addition, a workshop devoted to this topic has first created in 2017, search conversational artificial intelligence (SCAI), with the participation of academics and industrials\footnote{\url{https://scai.info/}}. 

Paragraph-based Question-Answering can be seen as a problem of reading comprehension, in which given a candidate document and a question, it finds the correct answer in the document. These systems use attention mechanisms~\citep{seo2016bidirectional} and memory networks~\citep{DBLP:journals/corr/WestonCB14}. 

\section{Positioning My Contributions in the State-of-the-Art}

Most of my work concerns task-oriented dialogue, which involves the different components presented in Section~\ref{c:intros:tod}. I worked on \ac{NLU}, \ac{DM} and more recently on \ac{NLG}. I have adapted neural models for distinct dialogue components that are presented in Chapter~\ref{c:contributionsTOD}. Since the past five years, I have also explored open-domain \ac{CQA} (Section~\ref{c:intros:cqa}) which I present in Chapter~\ref{c:contributionsConvQA}, with predictive and generative (encoder-decoder) models for question rewriting, reading-comprehension and knowledge-graph question generation. As an extension of the last approach, I have also explored graph-embeddings (Section~\ref{ss:graphemb}) and graph verbalisation with language models~\citep{montella2023investigating}. The contributions published the second half of this year or the work in progress are not included in this document.

My work has always followed the state-of-the-art at the time of publication. For a detail comparison of each contribution summarised in this manuscript with the related work of its time, we invite the curious reader to check the corresponding publications.
To provide a brief positioning of some contributions, the work on data collection, NLU, dialogue management and human evaluation I made for the EmoSpeech corpus was the first of its kind: a set of dialogues (12 distinct types of dialogue) in a Serious Game and in French.  
Moreover, we were among the first to propose in 2013 data-augmentation with back-translation and distributional representations for balancing biased models~\citep{gardent:hal-00905405}.

The work on inverse reinforcement learning followed on the seminal work of~\cite{ng2000algorithms}. It differed from previous work~\cite{paek2008automating, chandramohan2011user,el2012reward, boularias2010learning} because of the Bayesian \ac{IRL} algorithm that was applied for learning the tutor (i.e. system) reward function from experts. Furthermore, experts were taken from twelve distinct types of conversations in a serious game that were Human-Human (i.e. The EmoSpeech corpus) and not Machine-Machine (not from simulated conversations). 

I proposed to enrich a subset of the corpus CoQA~\citep{reddy_coqa_2019} releasing the corpus CoQAR\footnote{\url{https://github.com/Orange-OpenSource/COQAR}} with up to three out-of-context question paraphrases per question in conversations (Chapter~\ref{c:contributionsConvQA}, Section \ref{s:qr}). Unlike previous work, these paraphrases were made by professional English native annotators instead of using crowd-sourcing, guaranteeing fair earning and working conditions. We compared CoQAR to CANARD~\citep{elgohary_can_2019} that provided only one question rewriting per question in QuAC~\citep{choi2018quac}. I experimented with RoBERTa for answer extraction in both datasets CoQA and CoQAR with a without rewriting. Surprisingly solving the context through rewritten questions confuses RoBERTa, which is already good to solve co-references by itself, specially in short dialogue contexts as in these datasets.

I also utilised contextual embeddings (DistilBERT~\citep{sanh_distilbert_2020}, TransformersXL~\citep{dai2019transformer}) for estimating the reward function in long dialogues. I pointed out at that time the limitation of BERT-like models to deal with long contexts, which, besides the notable improvements of recent years, is still an open research problem. We also explored neural generation models such as BART~\citep{lewis2019bart} and T5~\citep{raffel2020exploring} for contextual question generation, question rewriting (Section \ref{s:qr}) and graph verbalisation~\citep{montella2022transfer,montella2023investigating}. I explored together with Sebastien Montella and Johannes Heinecke structural adapters for graph verbalisation~\citep{montella2023investigating} just after low rank emerged~\citep{hu2021lora} as a recommended way to optimally fine-tune LLMs.

After the advent of \ac{LLMs}, I am now questioning the performance of these models in complex tasks that required planning, such as dialogue. First, we need an evaluation methodology to assess the performance of LLMs in these tasks. Then we need to compare LLMs based reasoning~\citep{wei2022chain,yao2023tree, yao2022react} with \ac{RL} approaches, and explore recent trends for learning complex strategies such as algorithm distillation~\citep{laskin2022context}. I talk about these research paths in Chapter~\ref{c:project}.

\chapter{Contributions to Task-Oriented Dialogues}
\label{c:contributionsTOD}


\begin{flushright}
\parbox{.8\textwidth}{`José Arcadio Buendía pasó los largos meses de lluvia encerrado en un cuartito que construyó en el fondo de la casa para que nadie perturbara sus experimentos. Habiendo abandonado por completo las obligaciones domésticas, permaneció noches enteras en el patio vigilando el curso de los astros, y estuvo a punto de contraer una insolación por tratar de establecer un método exacto para encontrar el mediodía.' 
--- \emph{Gabriel García Marquez, Cien años de soledad.}}\\
\end{flushright}
\begin{flushright}
\parbox{.8\textwidth}{
 `José Arcadio Buendía spent the long months of the rainy season shut up in a small room that he had built in the rear of the house so that no one would disturb his experiments. Having completely abandoned his domestic obligations, he spent entire nights in the courtyard watching the course of the stars and he almost contracted sunstroke from trying to establish an exact method to ascertain noon.'
 --- \emph{Gabriel García Marquez, One Hundred Years of Solitude.}
 }
\end{flushright}
 

As introduced in Section~\ref{c:intros:tod}, task-oriented dialogue systems search to accomplish a task. This task can be for instance, \textit{information seeking}, in which the system search for items in a database according to the constraints obtained through natural language interaction. Thus, these constrains can be given by the user (\textit{`I am looking for a restaurant'}) or can be enquired by the system  (\textit{`which price range?'}, \textit{`where about?'}). Once the desired items are retrieved, the system informs the results back to the user. This kind of dialogue has rich interactions with dialogue acts such as: \textit{informing}, \textit{requesting}, \textit{clarifying}, \textit{rectifying}. State-of-the-art systems are centred on the information seeking task for a variety of domains: hotels, restaurants, touristic attractions, trains, flights, taxis, etc. Unfortunately, these dialogues are very specific and have difficulties to generalise to new domains and to more complex tasks (i.e. beyond information seeking).

I present in this Chapter my contributions to \ac{TOD} dialogues (Figure~\ref{f:achictecture}). I start by presenting my work on \ac{NLU} in Section~\ref{c:conts:nlu} and on Dialogue Management in Section~\ref{c:conts:dm}. My contributions to  \ac{NLU} concerns the definition of an annotation schema for the \textit{French corpus EmoSpeech}, in which 12 distinct dialogues were integrated in a \textit{serious game}. The dialogues involved various characters representing \textit{the system} and \textit{the player} and they were triggered at different levels of the game quest. Dialogues were modelled as an information seeking task, in which the system is always providing information to the player. The set of dialogue acts and goals are introduced in Section~\ref{c:conts:nlu}, as well as the models that were trained. Data augmentation through paraphrases via back translation, dictionaries, lexical resources and distributional semantics is also presented in Section \ref{ssec:paraphrases}. Moreover, the task of spoken language understanding is studied by using deep neural models and few-shot learning through risk minimisation in Section \ref{ss:slu}. Finally, I present the work of the PhD candidate Sebastien Montella on graph embeddings in Section~\ref{ss:graphemb}.

The contributions on Dialogue management are presented in Section~\ref{c:conts:dm}. I first present my own work on inverse reinforcement learning to find the implicit reward followed by humans in the EmoSpeech dialogue corpus. I also present a reward estimation by using deep learning. Finally, I present the work of the PhD candidate Thibault Cordier on hierarchical imitation learning.

\section{Language Understanding }
\label{c:conts:nlu}
Natural language understanding (NLU) is the task of mapping natural language sentences to semantic concepts. As a component of a spoken system, it would map utterances to a semantic representation that describes user intentions. This representation can be a  combination of  dialogue acts\footnote{dialogue acts are the performative action underlying a utterance\citep{austin1975things}} (e.g., greeting, request, inform, acknowledgement, confirm, etc.) and concept-value pairs, which are usually defined in a knowledge-base that describes the domain (e.g, Depart\_City="New York"). 
This section presents the semantic annotation of the French EmoSpeech corpus, the proposed Machine learning models, and the ways to improve the performance of these models via data augmentation. This section also includes the contributions to spoken language understanding, in which the input corresponds to the output of the \ac{ASR}. I show how the N-Best hypothesis are included as inputs to a \ac{CNN} that generates the sentence representation and how the context is handle by a \ac{LSTM}. In addition, I introduce risk minimisation for zero-shot learning on this task. Finally, I present the work of the PhD candidate Sebastien Montella on graph embedding. In order to user richer representations as input to neural models, the hyperbolic space was explored for treating temporal relations.

\subsection{NLU in a Serious Game}
\label{ss:nluemospeech}
\ac{ML} approaches such as logistic regression classifiers and conditional random fields were the state-of-the art back in 2011. I could collect a corpus through Wizard-of-Oz experiments, define an annotation scheme and train \ac{LR} and \ac{SVM} multi-class classifiers for the task of NLU. These models were integrated within distinct dialogues in a serious game.  I will start by describing briefly this work that gave origin to four publications~\citep{rojas-barahona2012building, rojas-barahona2012end, rojas-barahona2012should, gardent:hal-00905405}.

The serious game is a multiplayer quest where the players (3 teenagers) seek to
build a video game joystick in order to free their uncle trapped in
the game. To build this joystick, the players must explore a 
factory and achieve 17 mandatory goals (find the plans, get the
appropriate mould, retrieve some raw material from the storing shed, etc).
In addition, they can increase their score by achieving optional goals
which, when reached, provide them with extra information about the industry (therefore increasing their knowledge). In total, the players can achieve up to 28 goals by
conducting 12 separate subdialogs in various parts of the virtual
world. That is, dialogs in the game are long dialogs involving multiple players
in various settings.

\begin{table*}
\begin{center}
\begin{tabular}{|l|l|l|l|l|}
\hline
\scriptsize{Id}&\scriptsize{VC}&\scriptsize{Player}&\scriptsize{Mandatory Goals}&\scriptsize{Location}\\\hline
\scriptsize{1}&\scriptsize{Lucas}&\scriptsize{Ben}&\scriptsize{Find the address of the enterprise.}&\scriptsize{Unlce's place.}\\
\scriptsize{2}&\scriptsize{M.Jasper}&\scriptsize{Lucas}&\scriptsize{The manufacturing first step}&\scriptsize{Enterprise reception}\\
\scriptsize{3}&\scriptsize{Samir}&\scriptsize{Julie}&\scriptsize{Find the plans of the joystick}&\scriptsize{Designing Office}\\
\scriptsize{4}&\scriptsize{Samir}&\scriptsize{Julie}&\scriptsize{Find out what to do next}&\scriptsize{Designing Office}\\
\scriptsize{5}&\scriptsize{Melissa}&\scriptsize{Lucas}&\scriptsize{Manufacturing process ... }&\scriptsize{Plant}\\
\scriptsize{6}&\scriptsize{Melissa}&\scriptsize{Lucas}&\scriptsize{Find the right machine}&\scriptsize{Plant}\\
\scriptsize{7}&\scriptsize{Melissa}&\scriptsize{Lucas}&\scriptsize{Find out what to do next}&\scriptsize{Plant}\\
\scriptsize{8}&\scriptsize{Operator}&\scriptsize{Julie}&\scriptsize{Knowing about the material space ...}&\scriptsize{Material Space}\\
\scriptsize{9}&\scriptsize{Serge}&\scriptsize{Ben}&\scriptsize{Perform quality tests}&\scriptsize{Laboratory Tests}\\
\scriptsize{10}&\scriptsize{Serge}&\scriptsize{Ben}&\scriptsize{Find out what to do next}&\scriptsize{Laboratory Tests}\\
\scriptsize{11}&\scriptsize{Sophia}&\scriptsize{Julie}&\scriptsize{Find the electronic components.}&\scriptsize{Finishing}\\
\scriptsize{12}&\scriptsize{Sophia}&\scriptsize{Julie}&\scriptsize{Finishing process}&\scriptsize{Finishing}\\
\hline
\end{tabular}
\caption{Description of the subdialogs in the MP Game.}
\label{t:dials}
\end{center}
\end{table*}

Table~\ref{t:dials} summarises the characteristics of the subdialogs
conducted within the game highlighting three distinguishing
features of game dialogs.  First, the dialog participants vary whereby both the game agent and the player can change. Thus in the game, the player alternatively plays any of the three children involved in the quest while the game agent is successively, Lucas, M. Jasper, Samir, Melissa, an operator, Serge and Sophia.  Second, game dialogs are task-driven whereby each subdialog is related to a step in the game and each dialog turn aims to achieve a game goal and improve the player score.  Third, the context in which each subdialog takes place varies as the player moves around the world.

NLU annotation schema for dialogue is not necessarily
dictated by speech act theory alone but might also consider more practical issues namely, how well it will support interpretation and/or dialogue. To enhance learning, the annotation schema designed for the game combines core communicative acts~\citep{bunt} with domain specific information. The domain specific information specifies the goals being pursued/discussed/achieved etc. while the communicative act can be viewed as specifying how the current information state is updated by the speaker's utterance. 

\begin{table*}[htbp]
\begin{center}
\begin{tabular}{|l|l|l|l|l|l|}
\hline
\small{Dialogue Act}&\small{Label}&\small{Gloss}&\small{Speaker}  \\\hline
\small{Welcome greeting}&\small{greet}& Welcome greeting & P,S\\
\small{Farewell greeting}&\small{quit} & Farewell geeting & P, S\\
\small{Adress Request}&\small{ask(Goal)} & Request to pursue Goal & S\\
\small{Adress Request}&\small{help} & Request for help & P \\
\small{Confirm}&\small{yes} & Confirms previous query & P \\
\small{Disconfirm}&\small{no} & Disconfirms previous query & P \\
\small{Provide Information}&\small{inform(do(Goal))} & Provides information about how to achieve Goal & S \\
\small{Provide Information}&\small{Goal} & Provides information about the goal Goal & P\\
\small{Positive Feedback}&\small{ack}& Acknowledges understanding of preceding turn & S \\
\small{Propositional Question}&\small{ask(do(more(X)))} & Asks whether other topics should be discussed & S\\
\small{Set Question}&\small{ask(topic(X))} & Asks which other topics should be discussed & S \\
\small{Out of Context}&\small{other}& Out of context turn & P,S\\
\small{Misunderstanding}&\small{reqRep}& Request for rephrasing & S\\
\hline
\end{tabular}
\caption{The Annotation Scheme. P and S stands for Player and System respectively.} 
\label{tab:ann}
\end{center}
\end{table*}

\begin{figure}[ht]
\noindent
\begin{tabular}{ll}
\scriptsize{\textbf{dialog: 01\_dialogDirecteur-Tue Jun 14 11:04:23 2011}}&\\
\scriptsize{\textbf{M.Jasper:}Bonjour, je suis M.Jasper le directeur. } \\[-1ex]
    \scriptsize{Hello, I am the  director, Mr. Jasper. }&
    \scriptsize{$\rightarrow$\textbf{\textit{greet}}}\\
\scriptsize{\textbf{M.Jasper:} Qu'est-ce que je peux faire pour vous?} \\[-1ex]
\scriptsize{\textit{What can I do for you?}}&
    \scriptsize{$\rightarrow$ \textbf{\textit{ask(task(X))}}}\\
\colorbox{blue!20}{\scriptsize{\textbf{Lucas:}je dois sauver mon oncle}}\\[-1ex]
\colorbox{blue!20}{\scriptsize{\textit{I must rescue my uncle}}}&
    \scriptsize{$\rightarrow$ \color{pinegreen}\textbf{\textit{first\_step}}}\color{black} \\
\scriptsize{\textbf{M.Jasper:} Pour faire votre manette, il vous faut des plans. Allez voir dans le bureau d'études, 
ils devraient y être.}\\[-1ex]
\scriptsize{\textit{To build the joystick you will need the plans. You will find them in the Designing Office.}}&
  \scriptsize{$\rightarrow$  \textbf{\textit{inform(do(first\_step))}}} \\
\scriptsize{\textbf{M.Jasper:} J'aurais aimé continuer à répondre 
à vos questions mais je dois reprendre mon travail! 
Bonne Chance!}\\[-1ex]
\scriptsize{\textit{I have to go back to work! Good Luck!}}&
  \scriptsize{$\rightarrow$  \textbf{\textit{quit}}} \\
\end{tabular}

\caption{Excerpt from a dialogue in the EmoSpeech corpus. The corresponding user semantics is shown highlighted on the right.}
\label{f:emodial}
\end{figure}
Table \ref{tab:ann} gives the full list of Dialogue Acts used for annotation
together with the corresponding dialog acts and a gloss of their
meaning.  As can be seen the labels used are very specific to the game to facilitate the integration within the game (same goals as defined in the serious game) and to bypass much of the pragmatic reasoning necessary to associate a dialog turn with a communicative function. For instance, in the dialog above, the turn {\it je dois sauver mon   oncle (I must rescue my uncle)} does not explicitly state that the player (i) is seeking to achieve the game goal ``rescueing one's uncle'' and (ii) is asking the game agent for the first step towards achieving that goal. 

\subsubsection{Experimental setup}
\label{subsec:xpsetup}

We experimented with both an SVM and an LR\footnote{We used MALLET \citep{McCallumMALLET} for the LR classifier with L1 Regularisation.} classifier using different sets of features on
different data sets with and without TF*IDF (term frequency*Inverse
Document Frequency) filtering. 

\begin{table}[htbp]
\begin{center}
\begin{tabular}{|l|l|l|l|l|}
 \hline
&\multicolumn{2}{|c|}{\scriptsize{Whole Dialog}}&\multicolumn{2}{|c|}{\scriptsize{Subdialogs}}\\\cline{2-3}\cline{4-5}
&\scriptsize{w/o Tf*Idf}&\scriptsize{w/ Tf*Idf}&\scriptsize{w/o
  Tf*Idf}&\scriptsize{w/ Tf*Idf}
\\\hline
\scriptsize{LR}&\scriptsize{79.74}&\scriptsize{{\bf 90.26}}&\scriptsize{86.41}&\scriptsize{88.22}
\\\hline
\scriptsize{SVM}&\scriptsize{78.79}&\scriptsize{88.55}&\scriptsize{76.45}&\scriptsize{83.99}
\\\hline
\scriptsize{SVM (P)}&\scriptsize{}&\scriptsize{}&\scriptsize{78}&\scriptsize{83.55}
\\\hline
\end{tabular}
\end{center}
\caption{Global Results for the Logistic Regression (LR), the SVM (SVM)
  and the SVM Classifier with Penalisation (SVM(P)) }
\label{t:results}
\end{table}

We compared a single classifier on the whole dataset (the whole game) against 12 distinct classifiers, one for each subdialog. In both cases the
categories to be learned are restricted to the speaker's intent
(greet,quit,inform(Goal), ack, ask(do(more(X))), ask(topic(X)), other
in Table \ref{tab:ann}). Taking into account the game goals, the total
number of categories to be learned is 27. When learning on subdialogs, the number of categories to
be learned is smaller but so is the size of the training set. 
The features for the machine learning models were bag of words, in which stop words were filtered out, utterances were deaccented and converted to lower-case. In addition, we experimented with
various context length using as features the 0 to 4 previous dialogue acts. Subdialog identifiers were also used when training the classifier on the whole dialogue. More details are given in~\citep{rojas-barahona_building_2012}.


We also experimented using tf*idf filtering to limit the impact of frequent uninformative
words. Moreover, we experimented penalising those categories with more training instances, since the data was highly skewed. Dialogue acts that relate to optional goals were often not followed up by the players resulting in data sparseness. 

\subsubsection{Results}
\label{subsec:results}

Table \ref{t:results} shows the results for the 6 main configurations:
training on the whole dialog or on subdialogs, with and without
tf*idf filtering and using LR, SVM or SVM with
penalisation. The best results are obtained using the LR classifier on the whole dataset with tf*idf filtering. 
Penalising improved slightly the accuracy of the SVM when classifing without tf*idf filtering or when having a reduced context (0 or 2 previous acts in Table~\ref{tab:context}).

\textit{Impact of the tf*idf filtering}. 
Globally, the tf*idf filtering has a positive impact leading to an increase in accuracy ranging from 2.81 to 11.52 points. For the SVM classifier, the tf*idf filtering consistently lead to better results. However, for the LR classifier the filtering
adversely impacts performance on short
subdialogs (6 and 7), where one unique goal is being discussed. We conjecture that for these cases, the tf*idf filtering removed words which helped the classifier distinguish between turns about the unique goal from other turns. SVM with penalisation yields worse results with the tf*idf filtering than without, thus suggesting overfitting. In the next section we present how can we exploiting synonyms to improve generalisation.

\textit{Impact of contextual features}. Having a notion of context
is crucial for correctly interpreting dialog acts. As mentioned
above, we use the dialog acts of the previous turns to model
context. However the further back we look into the previous turns, the
more features there will be to train on. In other words, depending on the
number of previous turns considered, the data to learn from will be
more or less sparse. We experimented with 3 setups: a null context,
the dialog acts of the two previous turns and the dialog acts of the
four previous acts. Table \ref{tab:context} shows the results. 

\begin{table}[htbp]
\begin{center}
\begin{tabular}{|l|l|l|l|l|l|l|}
 \hline
&\multicolumn{3}{|c|}{\scriptsize{Whole
    Dialog}}&\multicolumn{3}{|c|}{\scriptsize{Subdialogs}}
\\\cline{2-4}\cline{5-7}
&\scriptsize{0}&\scriptsize{2}&\scriptsize{4}&\scriptsize{0}&\scriptsize{2}&\scriptsize{4}
\\\hline
\scriptsize{LR}&\scriptsize{88.43}&\scriptsize{90.26}&\scriptsize{90.26}&\scriptsize{84.43}&\scriptsize{87.59}&\scriptsize{88.22}
\\
\scriptsize{SVM}&\scriptsize{84.36}&\scriptsize{86.76}&\scriptsize{88.55}&\scriptsize{78.04}&\scriptsize{82.06}&\scriptsize{83.99}
\\
\scriptsize{SVM(P)}&\scriptsize{}&\scriptsize{}&\scriptsize{}&\scriptsize{79.32}&\scriptsize{83.12}&\scriptsize{83.55}
\\
\hline
\end{tabular}
\end{center}
\caption{The impact of context on accuracy. 0,2 and 4 indicates that
  the context is captured by having as features the dialog acts of 0, 2 and 4
previous turns respectively}
\label{tab:context}
\end{table}

\textit{Impact of dialog acts}. The accuracy varies per dialog acts from 48\% to
99\%. with most of the acts having an accuracy above
80\%. Unsurprisingly, the acts with lowest accuracy are also the
acts with fewest training data. The data is split randomly for the 30-fold evaluation with the risk of having insufficient data for optional goals. 

\paragraph{Estimating the User Satisfaction}: We applied the evaluation framework Paradise~\citep{walker-etal-1997-paradise} to assess these  dialogues in~\citep{rojas-barahona2012should}. We compared a rule-based dialogue manager and a dialogue manager that picks the answer randomly from a set of candidate responses. We found out that users prefer to talk to the second system because conversations are more fluid and there are less repetitions and misunderstandings.  However, with the second dialogue manager some times the player got stuck in the game and needed to repeat the dialogue with the virtual agent, impacting negatively the user satisfaction.

\subsubsection{Data Augmentation}
\label{ssec:paraphrases}

This work on data augmentation was published in \citep{gardent:hal-00905405}. We explored four ways of modifying the content features used for
classification: lemmatising the training and the test data; augmenting the training data with automatically acquired paraphrases; and substituting unknown words with synonyms or its distributional neighbours at run-time.

For Lemmatisation, we used the French version of
Treetagger\footnote{\scriptsize{\url{http://www.ims.uni-stuttgart.de/projekte/corplex/TreeTagger/}}}
to lemmatise both the training and the test data. Lemmas without any filtering were used to train classifiers. 
We then compare performance with and without lemmatisation. As we shall see, the lemma and
the POS tag provided by TreeTagger are also used to lookup synonym
dictionaries and EuroWordNet when using synonym handling at run-time. 
 
We were among the first to exploit automatically acquired paraphrases and to use these not only to increase the size of the training corpus but also to better balance
it\footnote{The Emospeech data is highly skewed with some classes
  being populated with many utterances and others with few.}. We
proceed as follows.

First, we generated paraphrases using a pivot machine translation
approach where each user utterance in the training corpus (around
$3610$ utterances) was translated into some target language and back
into French. Using six different languages (English, Spanish, Italian, German, Chinese and Arabian), we generated around $38000$
paraphrases. We used Google Translate API for translating.

Second, we eliminate from these paraphrases, words that are likely to
be incorrect lexical translations by removing words with low normalised term
frequency ($<$ 0.001) across translations i.e., lexical translations given by few
translations and/or translation systems. 
We then preprocessed the
paraphrases in the same way the utterances of the initial training
corpus were preprocessed i.e., utterances were unaccented, converted
to lower-case and stop words were removed, the remaining words were
filtered with TF*IDF. After preprocessing, duplicates were removed.

Third, we added the paraphrases to the training data seeking to
improve the balance between dialog acts per dialog.  The process to balance data was guided by the deviation of the category with lowest examples compared to the standard deviation. If the
deviation is lower than the standard deviation then we add paraphrases by keeping as much as possible the data balanced after replacement. We invite the interested reader to find more details about the algorithm proposed for balancing data in the paper~\citep{gardent:hal-00905405}.

\paragraph{Substituting Synonyms for Unknown Words} A word is unknown, if it is a well-formed French word\footnote{A word is determined to be a well-formed French word if it occurs in the LEFFF dictionary,  a large-scale morphological and syntactic lexicon for French \citep{sagot00521242}} and if it does not appear in the training corpus.  
When an unknown word $w$ is detected in a player utterance at run-time, we search for a word $w'$ which occurs in the training data and is either a synonym of $w$ or a distributional neighbour. After disambiguation, we substitute the unknown word for the synonym.

\begin{table*}[htbp]
\centering
  \begin{tabular}{|l|l|l|l|l|l|l|l|l|}
 \hline
\multicolumn{2}{|c|}{\scriptsize{H}}&\multicolumn{4}{|c|}{\scriptsize{Lemmatisation}}
\\\hline
 \scriptsize{H-H}&\scriptsize{Orig.}&\scriptsize{Lemmas}&\scriptsize{+EWN}&\scriptsize{+DIC}&\scriptsize{+RI}\\\hline
 \scriptsize{Orig.}&\scriptsize{$65.70\%\pm5.62$}&\scriptsize{$\boldsymbol{66.04\%\pm6.49}$}&\scriptsize{$68.17\%\pm6.98$}&\scriptsize{$67.92\%\pm4.51$}&\scriptsize{$66.83\%\pm5.92$}\\
 \scriptsize{Parap.}&\scriptsize{$70.89\%\pm6.45$}&\scriptsize{$\boldsymbol{74.31\%\pm4.78}$}*&\scriptsize{$\boldsymbol{74.60\%\pm5.99}$}*&\scriptsize{$\boldsymbol{73.07\%\pm7.71}$}*&\scriptsize{$\boldsymbol{72.63\%\pm5.82}$}*\\\hline
 \scriptsize{H-C}&\scriptsize{Orig.}&\scriptsize{Lemmas}&\scriptsize{+EWN}&\scriptsize{+DIC}&\scriptsize{+RI}\\\hline
 \scriptsize{Orig.}&\scriptsize{$59.71\%\pm16.42$}&\scriptsize{$59.88\%\pm7.19$}&\scriptsize{$61.14\%\pm16.65$}&\scriptsize{$61.41\%\pm16.59$}&\scriptsize{$60.75\%\pm17.39$}\\
 \scriptsize{Parap.}&\scriptsize{$59.82\%\pm15.53$}&\scriptsize{$59.48\%\pm14.02$}&\scriptsize{$\boldsymbol{61.70\%\pm14.09}$}*&\scriptsize{$\boldsymbol{62.01\%\pm14.37}$}*&\scriptsize{$61.16\%\pm14.41$}*\\\hline

 \end{tabular}
\caption{Accuracy on the H-H and on the  H-C corpus. The star denotes statistical significance with the Wilcoxon test ($p<0.005$) used for the HH corpus and the McNemar test ($p<0.005$) for the HC corpus.}
 \label{tab:results}
\end{table*}

To identify synonyms, we make use of two lexical resources namely, the
French version of EuroWordNet (EWN)~\citep{Vossen1998}, which includes
$92833$ synonyms, hyperonyms and hyponyms pairs, and a synonym lexicon
for French (DIC)~\footnote{DICOSYN
  (http://elsap1.unicaen.fr/dicosyn.html).} which contains $38505$
lemmas and $254149$ synonym pairs. While words are categorised into Noun, Verbs
and Adjectives in EWN, DIC contains no POS tag information.

To identify distributional neighbours, we constructed semantic word
spaces for each subdialog in the EmoSpeech corpus~\footnote{We also used
  \textit{distributional semantics} from the Gigaword corpus but the results
  were poor probably because of the very different text genre and
  domains between the the Gigaword and the game.} using random
indexing (RI) on the training corpus expanded with paraphrases. Using the cosine measure as similarity metrics, we then retrieve for any unknown word $w$, the word $w'$ which is most similar to $w$ and which appear in the training corpus.
 
For lexical disambiguation, two methods are compared. We use the POS tag
provided by TreeTagger. In this case, disambiguation is syntactic
only. Or we pick the synonym with highest probability based on a trigram language model trained on the H-H corpus.

\subsubsection{Results and Discussion}
\label{sec:evaluation}

Table~\ref{tab:results} summarises the results obtained in four main
configurations: (i) with and without paraphrases; (ii) with and without synonym handling; (iii) with and without lemmatisation; and (iv) when combining lemmatisation with synonym handling. We also compare the results obtained when evaluating using 10-fold cross validation on the training data (H-H dialogs) vs. evaluating the performance of the system on H-C interactions.

\paragraph{Overall Impact} The largest performance gain is obtained by a combination of the three techniques namely, data expansion, synonym handling and lemmatisation (+8.9 points for the cross-validation experiment and +2.3 for the H-C evaluation).

\paragraph{Impact of Lexical Substitution at Run Time}

We found that lexical resources are only useful when combined with
lemmatisation. This is unsurprising since synonym dictionaries and
EuroWordNet only contain lemmas. Indeed when distributional neighbours are used, lemmatisation has little impact (e.g., 65.11\% usingdistributional neighbours without lemmatisation on the H-H corpus without paraphrases vs. 66.41\% when using lemmatisation).

Another important issue when searching for a word synonym concerns lexical
disambiguation: the synonym used to replace an unknown word should
capture the meaning of that word in its given context. We tried using
a language model trained on the training corpus to choose between
synonym candidates (i.e., selecting the synonym yielding the highest
sentence probability when substituting that synonym for the unknown
word) but did not obtain a significant improvement. In contrast, it is
noticeable that synonym handling has a higher impact when using
EuroWordNet as a lexical resource. Since EuroWordNet contain
categorial information while the synonym dictionaries we used do not,
this suggests that the categorial disambiguation provided by
TreeTagger helps identifying an appropriate synonym in EuroWordNet. 

Finally, it is clear that the lexical resources used for this
experiment are limited in coverage and quality. We observed in
particular that some words which are very frequent in the training
data (and thus which could be used to replace unknown words) do not
occur in the synonym dictionaries. For instance when using paraphrases and dictionaries (fourth row and fourth column in Table~\ref{tab:results}) 50\% of the unknown words were solved, 17\% were illformed and
33\% remained unsolved. To compensate this deficiency, we
tried combining the three lexical resources in various ways (taking
the union or combining them in a pipeline using the first resource
that would yield a synonym). However the results did not improve and
even in some cases worsened due probably to the insufficient lexical
disambiguation. Interestingly, the results show that paraphrases
always improves synonym handling presumably because it increases the
size of the known vocabulary thereby increasing the possibility of
finding a known synonym.

In sum, synonym handling helps most when (i) words are lemmatised and
(ii) unknown words can be at least partially (i.e., using POS tag
information) disambiguated. Moreover since data expansion increases
the set of known words available as potential synonyms for unknown
words, combining synonym handling with data expansion further improves
accuracy.

\paragraph{Impact of Lemmatisation}
When evaluating using cross validation on the training corpus, lemmatisation increases accuracy by up to 3.42 points indicating that unseen word forms negatively impact accuracy. Noticeably however, lemmatisation has no significant impact when evaluating on the H-C corpus. This in turn suggests that the lower accuracy obtained on the H-C corpus  results not from unseen word forms but from unseen lemmas. 

\paragraph{Impact of Paraphrases} On the H-H corpus, data expansion has no significant impact when used alone. However it yields an increase of up to 8.27 points and in fact, has a statistically significant impact, for all configurations involving lemmatisation. Thus,  data expansion is best used in combination with lemmatisation and their combination permits creating better, more balanced and more general training data. On the H-C corpus however,
the impact is negative or insignificant suggesting that the decrease
in performance on the H-C corpus is due to content words that are new
with respect to the training data i.e., content words for which neither a synonym
nor a lemma can be found in the expanded training data.

While classifiers are routinely trained on dialog data to model the
dialog management process, the impact of such basic factors as
lemmatisation, automatic data expansion and synonym handling has
remained largely unexplored. The empirical evaluation described here
suggests that each of these factors can help improve performance but
that the impact will vary depending on their combination and on the
evaluation mode. Combining all three techniques yields the best
results. We conjecture that there are two main reasons for
this. First, synonym handling is best used in combination with POS
tagging and lemmatisation because these supports partial lexical
semantic disambiguation. Second, data expansion permits expanding the
set of known words thereby increasing the possibility of finding a
known synonym to replace an unknown word with.

\subsection{Spoken Language Understanding and Few-Shot Learning}
\label{ss:slu}
The following work was published in Coling 2016~\citep{rojasbarahona2016exploiting}. At that time the task of Spoken Language Understanding (SLU), namely semantic decoding, was  seen as a sequence tagging problem with models trained and tested on datasets with word-level annotations \citep{tur2013semantic,mesnil2015using,yao2013recurrent,5947649,deoras2013deep,sarikaya2014application}.  Nevertheless, spoken language understanding from \textit{unaligned data}, in which utterances are annotated with an abstract semantics, faces the additional challenge of not knowing which specific words are relevant for extracting the semantics. This problem was tackled in~\citep{zhou2011learning}, by using conditional random fields (CRFs) driven by finely-tuned hand-crafted features.
Other discriminative approaches that deal with unaligned data use some form of \textit{delexicalisation} or mapping of the input to known ontological concepts \citep{Henderson2012a,Henderson2014b}. The main disadvantage of delexicalisation is the difficulty in  scaling it, not only to larger and more complex dialogue domains but also to handle the many forms of language variation.

We proposed a semantic decoder that learns from unaligned data (Figure~\ref{f:dial}) and that exploits rich semantic distributed word representations instead of delexicalisation. The semantic decoder predicts the dialogue act and the set of slot-value pairs from a set of n-best hypotheses returned by an \ac{ASR}.  The prediction is made in two steps. First, a deep learning architecture is used for the joint prediction of dialogue acts and the presence or absence of slots. Second, the same architecture is reused for predicting the values of the slots that were detected by the first joint-classifier. The deep architecture combines  sentence and context representations. A \ac{CNN}~\citep{collobert2011natural} is used to generate the sentence representation, while a \ac{LSTM} network~\citep{hochreiter1997long} is used to generate the context representation.  A non-linear function then combines the top layers of these neural networks and distinct Softmax layers are used to predict the dialogue act and slots in the first joint model. In the second model, a single Softmax predicts the possible values for each slot.

\begin{figure}[ht]
\noindent
\begin{tabular}{ll}
\scriptsize{\textbf{voip-922209b777-20130325\_155209}}&\\
\scriptsize{\textbf{System: }Hello , welcome to the Cambridge restaurant system? } \\[-1ex]
    \scriptsize{You can ask for restaurants by area, price range or food type. }\\[-1ex]
    \scriptsize{How may I help you?}&
    \scriptsize{$\rightarrow$\textbf{\textit{welcomemsg}}}\\
\colorbox{blue!20}{\scriptsize{\textbf{User: }  i am looking for a moderately priced restaurant}}\\[-1ex]
\colorbox{blue!20}{\scriptsize{in the north part}}&
    \scriptsize{$\rightarrow$ \color{pinegreen}\textbf{\textit{inform(area=north,pricerange=moderate)}}\color{black}}\\
\scriptsize{\textbf{System: }meghna is a nice restaurant}\\[-1ex]
\scriptsize{in the north of town in the moderate price range}&
    \scriptsize{$\rightarrow$ \textbf{\textit{offer(name=meghna),}}} \\[-1ex] &  \hspace{0.3cm}\scriptsize{\textbf{\textit{inform(pricerange=moderate,area=north)}}}\\
\colorbox{blue!20}{\scriptsize{\textbf{User: } is everything else}} &  \scriptsize{$\rightarrow$ \color{pinegreen}\textbf{\textit{reqalts()}}}\color{black}\\
\scriptsize{\textbf{System: }golden wok is a nice restaurant}\\[-1ex]
\scriptsize{in the north of town in the moderate price range}&
   \scriptsize{$\rightarrow$ \textbf{\textit{offer(name=golden...),}}} \\[-1ex] &  \hspace{0.3cm}\scriptsize{\textbf{\textit{inform(pricerange=moderate, area=north)}}}\\
\colorbox{blue!20}{\scriptsize{\textbf{User: } can i get the phone number}} &  \scriptsize{$\rightarrow$ \color{pinegreen}\textbf{\textit{request(slot=phone)}}\color{black}}\\
\scriptsize{\textbf{System: }The phone number of golden wok is ...} &
  \scriptsize{$\rightarrow$  \textbf{\textit{offer(name=golden wok),inform(phone=01..)}}} \\
\colorbox{blue!20}{\scriptsize{ \textbf{User: } \textit{type of food do they serve}}} &  \scriptsize{$\rightarrow$ \color{pinegreen}\textbf{\textit{request(slot=food)}}\color{black}}

\end{tabular}

\caption{Excerpt from a dialogue in the DSTC2 corpus. The top-best ASR hypothesis is shown highlighted on the left, and the corresponding user semantics is shown highlighted on the right.}
\label{f:dial}
\end{figure}

We evaluated our models on two datasets DSTC2~\citep{henderson2014second} and In-car~\citep{tsiakoulis2012statistical} using accuracy, f-measure and the Item Cross Entropy (ICE) score~\citep{thomson2008evaluating}. Our models outperform previous proposed models, without using manually designed features and without any pre-processing of the input (e.g., stop words filtering, delexicalisation). They do this by exploiting distributed word representations and
we claim that this allows semantic decoders to be built that can easily scale to larger and more complex dialogue domains.

\subsubsection{Deep Learning Semantic Decoder}
\label{s:dlsemi}
\begin{figure}[ht]
\centering
\includegraphics[scale=0.4]{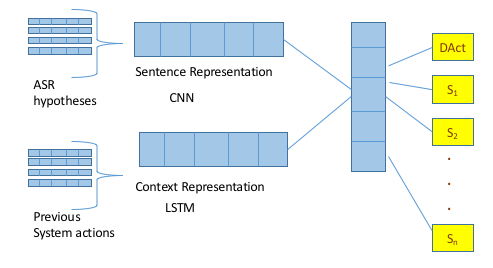}
\caption{Combination of sentence and context representations for the joint prediction of dialogue acts and slots.}
\label{f:dlarchi}
\end{figure}

We split the task of  semantic decoding into two steps: (i) training a joint model for predicting the dialogue act and presence or absence of slots and (ii) predicting the values for the most probable slots detected in (i). 
As shown in Figure~\ref{f:dlarchi}, we use the same deep learning architecture in both steps for combining sentence and context representations to generate the final hidden unit that feeds one or many softmax layers.  In the first step, as shown in the Figure, there are distinct softmax layers for the joint optimisation of the dialogue act and each possible slot.  In the second step there is a single softmax layer that predicts the value of each specific slot. In the following we explain this architecture in more detail.

\subsubsection{Sentence Representation}
\label{ss:srepr}
A CNN is used for generating the hypothesis representation, then these representations are weighted by their confidence scores and then summed up to obtain the sentence representation (Figure~\ref{f:cnn}). 

The CNN is a variant of {~\citep{kim2014convolutional}, in which the inputs are the word vectors in each \textit{ASR hypothesis}. Let $x_i$ be a $k-$dimensional word embedding for the $i$-th word in a hypothesis.
A hypothesis of length $m$ is represented as:
	$x_{1:m}=\mathbf{x}_1 \bigoplus \mathbf{x}_2  \bigoplus ... \bigoplus \mathbf{x}_m$
where $\bigoplus$ is the concatenation operator. A convolutional operation is applied to a window of $l$ words to produce a new feature.
\begin{equation}
 c_i=f(\mathbf{w}\cdot\mathbf{x}_{i:i+l-1}+b)
 \label{eq:conv}
\end{equation} 
where $f$ is the hyperbolic tangent function; $w \in \mathbb{R}^{lk}$ is a filter  applied to a window of $l$ words and $b \in \mathbb{R}$ is a bias term.
The filter is applied to every window of words in the sentence to produce a feature map.
\begin{equation}
	\mathbf{c}=[ c_1, c_2, ... , c_{n-l+1}]
	\label{eq:featmap}
\end{equation}
with  $\mathbf{c} \in \mathbb{R}^{n-l+1}$. A max pooling operation is then applied to give the maximum value $c=max\{\mathbf{c}\}$ as the representative feature for that filter. Multiple filters can be applied by varying the window size to obtain several adjacent features for a given hypothesis.
These features $\hat{f}_j$ for the hypothesis $j \in H$ are then multiplied by the ASR confidence score $p_j$\footnote{The posterior probability of hypothesis $j$ in the N-best list.} 
and summed over all ASR hypotheses to generate a representation for the sentence $s_t$ (Equation~\ref{eq:wsum}), as shown in Figure~\ref{f:cnn}.
\begin{equation}
 s_t= \sum_{j \in H} \hat{f}_j * p_j
 \label{eq:wsum}
\end{equation}

\begin{figure}[h]
\begin{tikzpicture}[scale=.5]
\node[text width=0.2cm] at (-1.3,8.5) {\scriptsize{i}};
\node[text width=0.2cm] at (-1.3,7.5) {\scriptsize{'m}};
\node[text width=0.2cm] at (-1.7,6.5) {\scriptsize{looking}};
\node[text width=0.2cm] at (-1.3,5.5) {\scriptsize{for}};
\node[text width=0.2cm] at (-1.3,4.5) {\scriptsize{uh}};
\node[text width=0.2cm] at (-1.3,3.5) {\scriptsize{a}};
\node[text width=0.2cm] at (-2.1,2.5) {\scriptsize{moderately}};
\node[text width=0.2cm] at (-1.3,1.5) {\scriptsize{priced}};
\node[text width=0.2cm] at (-1.9,0.5) {\scriptsize{restaurant}};
\draw[thick, scale=1] (0, 0) grid (6, 9);
\draw[line width=.8pt,very thick,blue](0,0)--(0,2);
\draw[line width=.8pt,very thick,blue](0,2)--(6,2);
\draw[line width=.8pt,very thick,blue](6,2)--(6,0);
\draw[line width=.8pt,very thick,blue](0,0)--(6,0);

\draw[line width=.8pt,very thick,green](0,3)--(0,6);
\draw[line width=.8pt,very thick,green](0,6)--(6,6);
\draw[line width=.8pt,very thick,green](6,6)--(6,3);
\draw[line width=.8pt,very thick,green](0,3)--(6,3);

\draw[line width=.8pt,very thick,blue](0,7)--(0,9);
\draw[line width=.8pt,very thick,blue](0,9)--(6,9);
\draw[line width=.8pt,very thick,blue](6,9)--(6,7);
\draw[line width=.8pt,very thick,blue](0,7)--(6,7);

\draw[line width=.8pt] (1, 10)--(7, 10);
\draw[thin, scale=1] (7, 1) grid (7, 10);
\draw[line width=.8pt, black](7,10)--(7,10);
\draw[line width=.8pt,black](7,1)--(7,10);
\draw[thin, scale=1] (2, 11) grid (8, 11);
\draw[thin, scale=1] (8, 2) grid (8, 11);
\draw[line width=.8pt,thick,black](2,11)--(8,11);

\draw [ thick,  decoration={  brace,   mirror,  raise=0.5cm  }, decorate ](-1,0) -- (9,0) node [pos=0.5,anchor=north,yshift=-0.55cm] {\scriptsize{ASR hypotheses}};

\draw[thin] (13, 0) grid (14,7);
\draw[thin] (12, 1) grid (13, 8);
\draw[thin] (11, 2) grid (12, 9);
\draw[line width=.8pt,very thick,blue](13,0)--(14,0);\draw[line width=.8pt,very thick,blue](14,0)--(14,1);
\draw[line width=.8pt,very thick,blue](14,1)--(13,1);\draw[line width=.8pt,very thick,blue](13,0)--(13,1);

\draw[line width=.8pt,very thick,green](12,3)--(13,3);\draw[line width=.8pt,very thick,green](13,3)--(13,4);
\draw[line width=.8pt,very thick,green](12,4)--(13,4);\draw[line width=.8pt,very thick,green](12,3)--(12,4);

\draw[line width=.8pt,very thick,blue](13,6)--(14,6);\draw[line width=.8pt,very thick,blue](14,6)--(14,7);
\draw[line width=.8pt,very thick,blue](14,7)--(13,7);\draw[line width=.8pt,very thick,blue](13,6)--(13,7);


\draw [dashed,green] (6,3) -- (12,3);
\draw [dashed,green] (6,6) -- (13,4);
\draw [dashed,blue] (6,0) -- (13,0);
\draw [dashed,blue] (6,2) -- (13,1);

\draw [dashed,blue] (6,7) -- (13,6);
\draw [dashed,blue] (6,9) -- (13,7);
\draw [ thick,  decoration={  brace,   mirror,  raise=0.5cm  }, decorate ](11,0) -- (14,0) node [pos=0.5,anchor=north,yshift=-0.55cm] {\scriptsize{Convolutional layers}};

\draw [decorate,decoration={brace,amplitude=10pt},xshift=-4pt,yshift=0pt,blue]   (-1,9) -- (2,11) node [black,midway,xshift=-0.5cm,yshift=0.5cm] {\scriptsize {N\_best}};

\begin{scope}[
            yshift=-53,every node/.append style={
            yslant=0.4,xslant=-0.5},yslant=0.2,xslant=-1
            ]
        \draw[ black] (19,0) grid (20,4); 
        \draw[ black] (21,1) grid (22,5); 
        \draw[ black] (23,2) grid (24,6); 
    \end{scope}
\draw [dashed,green] (13,3) -- (20.1 ,5.1);
\draw [dashed,green] (13,4) -- (19.4 ,6);
\draw [dashed,blue] (14,7) -- (20.1 ,5.1);
\draw [dashed,blue] (14,0) -- (21,4.3);

\fill[red!40,ultra thick] (24,3) rectangle (25,7);
\draw [ thick,  decoration={  brace,   mirror,  raise=0.5cm  }, decorate ](23,0) -- (26,0) node [pos=0.5,anchor=north,yshift=-0.55cm] {\scriptsize{Sentence Representation:}};
\node at (25,-2) {\scriptsize{weighted sum of hyps}};

\draw [ thick,  decoration={  brace,   mirror,  raise=0.5cm  }, decorate ](16,0) -- (21,0) node [pos=0.5,anchor=north,yshift=-0.55cm] {\scriptsize{hypotheses representations}};
\draw [dashed,red!50] (18,7.8) -- (24 ,7);
\draw [dashed,red!50] (20,2.1) -- (24 ,3);

\end{tikzpicture}
\caption{Sentence Representation: after applying convolution operations on the N-best list of ASR hypotheses,  the resulting hidden layers are weighted by the ASR confidence scores and summed.}
\label{f:cnn}
\end{figure}
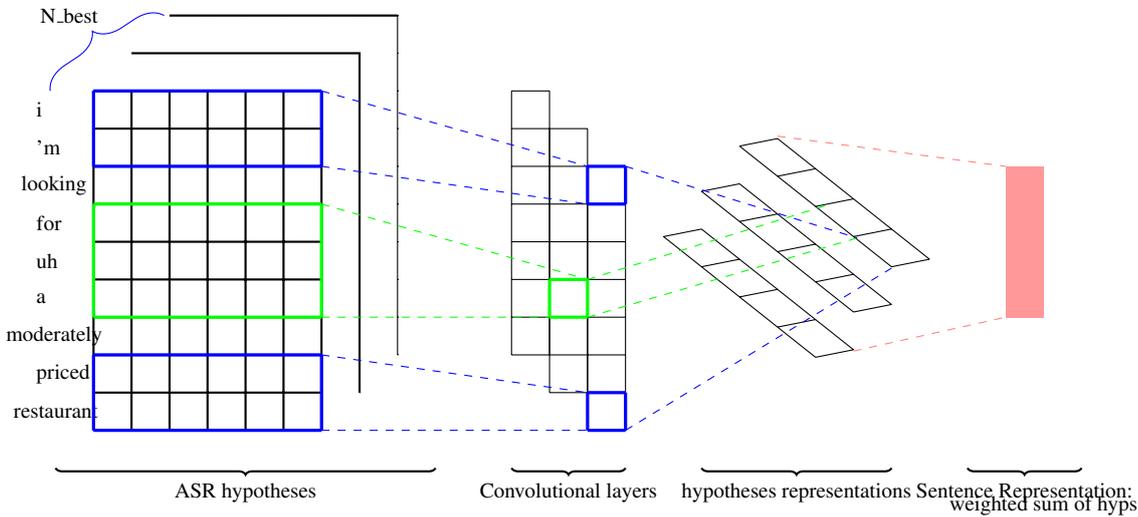

\subsubsection{Context Representation}
\label{ss:ctxtrepr}
An \ac{LSTM}~\citep{hochreiter1997long} is used for tracking the context implied by previous dialogue system actions. The top layer of this \ac{LSTM} network then provides the context representation for decoding the current input utterance. 

An \ac{LSTM} is a sequence model that utilises a memory cell capable of preserving states over long periods of time.   This cell is recurrently connected to itself and it has three multiplication units, an input gate, a forget gate  and an output gate. These gating vectors are in [0,1]. The cell makes selective decisions about what information is preserved, and when to allow access to units, via gates that open and close.

As shown in Figure~\ref{f:dial}, system actions are encoded in the form of a system dialogue act plus one or more slot-value pairs.  To track the history of system actions, 
slots and values are treated as words and the input $x_t$ is formed from its corresponding word vectors. The length of the context can vary.  We consider all the system actions previous to the current user utterance, or a window $l$ of the previous system actions. For instance, if we are currently processing the last user input in Figure~\ref{f:dial},  in which L is the total number of system actions, we can consider all previous system actions (L=4), or the last $l$ system actions, where $l<L$.

\subsubsection{Combining Sentence and Context}
\label{ss:comb}
We study in this paper two ways of combining the sentence $\mathbf{s}_t$ and the context $\mathbf{h}_t$ representations. The first straightforward way is to apply a non linear function to their weighted sum:

\begin{equation}
   \mathbf{\hat{h}_t}=tanh(\mathbf{Ws}\cdot\mathbf{s}_t+\mathbf{Wc}\cdot\mathbf{h}_t)
   \label{eq:comb}
\end{equation}
The second way is to let the sentence representation be the last input to the LSTM network, then $\mathbf{\hat{h}_t} = \mathbf{h}_t$.
For classification a softmax layer is used for each prediction.
The result of the prediction is the most probable class.
The back-propagation optimisation is done by minimising the negative log-likelihood loss function through stochastic gradient descent.

\subsubsection{Experimental Evaluation}
\label{s:exp}
In this section we introduce the corpora, and describe the experiments performed and the evaluation metrics used.
\subsubsection{Corpora}
Experimental evaluation used two similar datasets: DSTC2~\citep{henderson2014second} and In-car~\citep{tsiakoulis2012statistical}. Both corpora were collected using a spoken dialogue system which provides restaurant information system for the city of Cambridge. Users can specify restaurant suggestions by area, price-range and food type and can then query the system for additional restaurant specific information such as phone number, post code and address. The first dialogue corpus was released for the dialogue state tracking challenge and we use here the semantic annotations that were also provided~\footnote{The DSTC2 corpus is publicly available in: \url{http://camdial.org/~mh521/dstc/}}. The trainset has $2118$  dialogues and $15611$ turns in total while the testset has $1117$ dialogues  and $9890$ turns in total. 

The second corpus contains dialogues collected under various noisy in-car conditions. In a stationary car with the air conditioning fan on and off, in a moving car and in a car simulator~\citep{tsiakoulis2012statistical}~\footnote{This corpus has been obtained in an industry funded project and therefore it is not available for public use.}. 
The trainset has $1508$ dialogues and $10532$ turns in total and the testset has $641$ dialogues and $4861$ turns in total.  Because of the noise, the average word error rate (WER = $37\%$) is significantly higher than for DSTC2 (around $29\%$).

\subsubsection{Experiments}
\paragraph{Step I: Joint classification of dialogue-acts and slots:}
We evaluated five different model configurations for the joint classification of dialogue-acts and presence or absence of slots. 
\begin{itemize}
\item{\textbf{CNN}}: the softmax layers for the joint classification of dialogue acts and slots are connected directly to the CNN sentence representation with no context.  
\item{\textbf{CNN+LSTM}}: we study the influence of context by considering the previous system actions (Section~\ref{ss:ctxtrepr}, Eq.~\ref{eq:comb}), here we study the different context length, by using a context window of 1, 4, and all the previous system actions, namely \textbf{CNN+LSTM\_w1},  \textbf{CNN+LSTM\_w4} and  \textbf{CNN+LSTM\_w} respectively.
\item{\textbf{LSTM\_all}}: Finally, we study the impact of long distance dependencies, by using mainly the LSTM model, with the previous system actions as input, but we inject the sentence representation as the last LSTM input.
\end{itemize}
 
\paragraph{Step II: Classification of slot value pairs:}
We select the best model in step I for predicting the presence of slots, then for each slot present we predict the  value, by using again the best architecture from the previous step.

\subsubsection{Evaluation Metrics}
We evaluate the performance of our models by using the conventional metrics for classification, namely accuracy, precision, recall and F-measure (F1-score).  
In addition, we used the ICE score to measure the overall quality of the distribution returned by the models taken into account the hypotheses and the reference semantics (ie. ground-truth)\citep{thomson2008evaluating}. 
\subsubsection{Results and Discussion}
\label{s:eval}
In this section we report the results on DSTC2 and In-car dialogue corpora.

\paragraph{Step I: Joint classification of dialogue-acts and slots:}
For this step, the classifiers must predict jointly 14 dialogue acts and 5 slots for the DSTC2 dataset as well as 14 dialogue acts and 7 slots for the In-car dataset.
We evaluate both (i) using 10 fold cross-validation on the trainsets and (ii) on the corpora' testsets.

Our results on  10 fold cross-validation results on both corpora suggest that for DTSC2, the context representation is not significantly impacting the prediction. Although, the model with a window of 4 ,\textbf{CNN+LSTM\_w4}, improves slightly the accuracy and f1-score. On the In-car dataset, however,  including the context does help to disambiguate the semantic predictions from ill-formed hypotheses. This is expected, since this data set has a much higher error rate and hence higher levels of confusion in the ASR output. Although there is no significant difference on the f1-score when using the immediate previous system act ($w1$) or a longer context, 
\textbf{CNN+LSTM\_w} gives a better accuracy and a lower ICE score on this dataset.

Table \ref{t:tstset} shows the results on the test sets. Consequently, when evaluating on the DSTC2 test set, a window of 4 ($w4$), performs slightly better than other window sizes and better than the simple \textbf{CNN} model.
On the In-car testset, a context window of 4 outperforms all the other settings: \textbf{CNN+LSTM}. However, on this test set using the sentence
representation as the last input to the LSTM context neural network (section \ref{ss:comb})  improves the f1-score and reduces the ICE error. 
 
\begin{table}[htbp]
\centering
  \begin{tabular}{l | ll | lll | l}
 \hline
 \scriptsize{Corpus}&\scriptsize{Metric}&\scriptsize{CNN}&\multicolumn{3}{|c|}{\scriptsize{CNN+LSTM}}&\scriptsize{LSTM\_all}\\
 \hline
 \scriptsize{-}&\scriptsize{-}&\scriptsize{-}&\scriptsize{w1.}&\scriptsize{w4}&\scriptsize{w}&\scriptsize{-}\\\hline
 \scriptsize{DSTC2}&\scriptsize{acc.}&\scriptsize{$96.03\%$}&\scriptsize{$95.79\%$}&\scriptsize{$95.79\%$}&\scriptsize{$95.69\%$}&\scriptsize{$95.59\%$}\\
& \scriptsize{P.}&\scriptsize{$89.73\%$}&\scriptsize{$88.69\%$}&\scriptsize{$88.95\%$}&\scriptsize{$88.38\%$}&\scriptsize{$88.15\%$}\\
 &\scriptsize{R.}&\scriptsize{$84.74\%$}&\scriptsize{$85.09\%$}&\scriptsize{$86.02\%$}&\scriptsize{$85.96\%$}&\scriptsize{$84.76\%$}\\
& \scriptsize{F1}&\scriptsize{$87.14\%$}&\scriptsize{$86.83\%$}&\scriptsize{$\mathbf{87.43}\%$}&\scriptsize{$87.12\%$}&\scriptsize{$86.42\%$}\\
 &\scriptsize{ICE}&\scriptsize{$\mathbf{0.268}$}&\scriptsize{$0.278$}&\scriptsize{$0.292$}&\scriptsize{$0.297$}&\scriptsize{$0.308$}\\
 \hline
 \scriptsize{In-car}&\scriptsize{acc.}&\scriptsize{$\mathbf{87.60}\%$}&\scriptsize{$82.19\%$}&\scriptsize{$82.25\%$}&\scriptsize{$82.14\%$}&\scriptsize{$82.3\%$}\\
 &\scriptsize{P.}&\scriptsize{$69.96\%$}&\scriptsize{$79.52\%$}&\scriptsize{$79.29\%$}&\scriptsize{$80.25\%$}&\scriptsize{$78.12\%$}\\
 &\scriptsize{R.}&\scriptsize{$62.14\%$}&\scriptsize{$71.09\%$}&\scriptsize{$71.59\%$}&\scriptsize{$70.9\%$}&\scriptsize{$74.04\%$}\\
 &\scriptsize{F1}&\scriptsize{$65.53\%$}&\scriptsize{$74.89\%$}&\scriptsize{$75.15\%$}&\scriptsize{$75.02\%$}&\scriptsize{$\mathbf{75.9}\%$}\\
 &\scriptsize{ICE}&\scriptsize{$1.332$}&\scriptsize{$1.344$}&\scriptsize{$1.333$}&\scriptsize{$1.421$}&\scriptsize{$\mathbf{1.106}$}\\
 \hline
 \end{tabular}
 \caption{Evaluation of the Step I on DSTC2 and In-car testsets. We also compare two ways of combining sentence and context representation: \textbf{CNN+LSTM} models (combining sentence and context representation through a non linear function) and \textbf{LSTM\_all} model (embedding the sentence representation into the context model).}\label{t:tstset}
 \end{table}

\paragraph{Step II: Prediction of slot value pairs}
For evaluating Step II, we selected the best model obtained during the 10-fold cross-validation experiments in terms of F1 score. 
For both corpora, this was the \textbf{CNN+LSTM\_w4} configuration. 
For DSTC2, it was the $4^{th}$-fold crossvalidation with $Acc=90.42\%$, $F1=88.69\%$ and $\mbox{ICE}=0.251$. For In-car, it was the $5^{th}$-fold crossvalidation with $Acc=93.13\%$, $F1=81.49\%$ and $\mbox{ICE}=0.393$. We used these models to classify whether a given slot appears in a given hypothesis or not. Then for that slot, we train another \textbf{CNN+LSTM\_w4} classifier for predicting its values. In the In-car corpus the slot "type" has only one possible value "restaurant".  Similarly,  the slot "task" can only be the value "find". For these slots with only one value, we report values using the model of Step I, since it is enough to detect the slot in the utterance. 

\begin{table}[ht]
\centering
  \begin{tabular}{l|lllll|lllll}
 \hline
 \scriptsize{}&\multicolumn{5}{|c|}{\scriptsize{DSTC2}}&\multicolumn{5}{|c}{\scriptsize{In-car}}\\\hline
 \scriptsize{Slot}&\scriptsize{Acc.}&\scriptsize{P.}&\scriptsize{R.}&\scriptsize{F1}&\scriptsize{ICE}&\scriptsize{Acc.}&\scriptsize{P.}&\scriptsize{R.}&\scriptsize{F1}&\scriptsize{ICE}\\
 \hline
 \scriptsize{Slot\footnote{"Slot" is used when no value is given for the slot (e.g., "What kind of food do they serve?"/request(slot=food)).}}&\scriptsize{95.29\%}&\scriptsize{90.89\%}&\scriptsize{95.72\%}&\scriptsize{93.24\%}&\scriptsize{0.478}&\scriptsize{89.92\%}& \scriptsize{74.73\%}& \scriptsize{61.56\%}& \scriptsize{67.51\%}& \scriptsize{ 0.743}\\
\scriptsize{Area}&\scriptsize{91.77\%}&\scriptsize{92.66\%}&\scriptsize{92.83\%}&\scriptsize{92.74\%}&\scriptsize{ 0.563}&\scriptsize{72.03\%}& \scriptsize{72.56\%}& \scriptsize{74.28\%}& \scriptsize{73.41\%}& \scriptsize{1.676}\\
\scriptsize{Food}&\scriptsize{71.37\%}&\scriptsize{73.19\%}&\scriptsize{76.02\%}&\scriptsize{74.58\%}&\scriptsize{1.989}&  \scriptsize{66.46\%}& \scriptsize{64.27\%}& \scriptsize{68.70\%}& \scriptsize{66.41\%}& \scriptsize{2.309}\\
\scriptsize{Price}&\scriptsize{94.62\%}&\scriptsize{91.33\%}&\scriptsize{94.49\%}&\scriptsize{92.89\%}&\scriptsize{ 0.729}&  \scriptsize{93.96\%}& \scriptsize{88.77\%}& \scriptsize{92.03\%}& \scriptsize{90.37\%}& \scriptsize{0.632}\\
\scriptsize{This\footnote{"This" is used for annotating elliptical utterances (e.g., "I dont care"/inform(this='dontcare')).}}&\scriptsize{98.70\%}&\scriptsize{96.79\%}&\scriptsize{93.92\%}&\scriptsize{95.33\%}&\scriptsize{0.113}&  \scriptsize{97.16\%}& \scriptsize{96.14\%}& \scriptsize{84.72\%}& \scriptsize{90.07\%}& \scriptsize{0.214}\\
\scriptsize{Type}&  \scriptsize{-}& \scriptsize{-} &\scriptsize{-} &\scriptsize{-}& \scriptsize{-}&\scriptsize{95.56\%}& \scriptsize{95.09\%} &\scriptsize{86.69\%} &\scriptsize{90.69\%}& \scriptsize{0.290}\\
 \scriptsize{Task}& \scriptsize{-}& \scriptsize{-}& \scriptsize{-}& \scriptsize{-}& \scriptsize{-}& \scriptsize{97.12\%}& \scriptsize{83.24\%}& \scriptsize{64.93\%}& \scriptsize{72.95\%}& \scriptsize{0.175}\\
 \hline
\scriptsize{Mean}&\scriptsize{90.35\%}&\scriptsize{88.97\%}&\scriptsize{90.60\%}&\scriptsize{89.76\%}&\scriptsize{0.774}&\scriptsize{87.47\%}& \scriptsize{82.11\%}& \scriptsize{76.13\%}& \scriptsize{78.77\%}& \scriptsize{0.863}\\
\scriptsize{St.Dev.}&\scriptsize{0.109}&\scriptsize{0.091}&\scriptsize{0.082}&\scriptsize{0.085}&\scriptsize{0.715}& \scriptsize{ 0.128}& \scriptsize{ 0.121}& \scriptsize{0.118}& \scriptsize{0.112}&  \scriptsize{ 0.821} \\\hline\hline
 \end{tabular}
 \caption{Evaluation of the step II: the slot-value pairs classification on DSTC2 and In-car.}
 \label{t:s2dstc2}
 \end{table}

Given that there is no domain specific delexicalisation, the models achieve a good level of performance overall (Table~\ref{t:s2dstc2}).  
Note that the slot "food"  has 74 possible values in DSTC2 and 25 in In-car.  Hence, this slot has much higher cardinality than all the other slots.

 
 \paragraph{Overall performance}
 
A baseline for assessing overall performance is provided by the model presented in~\citep{Henderson2012a}, in which the vector representation is obtained by summing up the frequency of n-grams extracted from the 10-best hypotheses, weighted by their confidence scores. 
Here we compare our performance against Henderson's model with and without context features, namely WNGRAMS+Ctxt and WNGRAMS repectively.
Henderson reported his results on the In-car dataset. A similar model, namely SLU1, was evaluated on DSTC2 in~\citep{williams2014web}. Both  implementations consist of many binary classifiers for dialogue act and slot-value pairs. 

\begin{table}[htbp]
\centering
  \begin{tabular}{llll}
 \hline
 \scriptsize{Corpus}&\scriptsize{Model}&\scriptsize{F1}&\scriptsize{ICE}\\\hline
 \scriptsize{DSTC2}&\scriptsize{SLU1~\citep{williams2014web}}&\scriptsize{$80.2\%$}&\scriptsize{$1.943$}\\
 &\scriptsize{CNN+LSTM\_w4}&\scriptsize{$\mathbf{83.59}\%$}&\scriptsize{$\mathbf{0.758}$}\\
 \scriptsize{In-car}&\scriptsize{WNGRAMS~\citep{Henderson2012a}}&\scriptsize{$70.8\%$}&\scriptsize{$1.76$}\\
&\scriptsize{WNGRAMS+Ctxt~\citep{Henderson2012a}}&\scriptsize{$\mathbf{74.2}\%$}&\scriptsize{$1.497$}\\
 &\scriptsize{CNN+LSTM\_w4}&\scriptsize{$73.06\%$}&\scriptsize{$\mathbf{1.106}$}\\\hline 
\end{tabular}
\caption{Overall performance of the setting CNN+LST\_w4 semantic decoder.}\label{t:overall}
 \end{table}
 In terms of the ICE score, the model CNN+LSTM W4 outperforms all the baselines (Table~\ref{t:overall}). In terms of the F1 score, the model significantly outperforms the SLU1 and WNGRAMS baselines. However it is slightly worse than WNGRAMS+Ctxt, which has been enhanced with context features on In-car. Remember however, that our model uses only word-embeddings for  automatically generating sentence and context representations without having any manually designed features or using explicit application specific semantic dictionaries.


\subsubsection{Few-shot Learning through Risk Minimisation (RM)}
\label{s:rm}
We treat rarely seen slots by following two steps. (i) We optimise jointly in a deep neural network the weights that feed multiple binary Softmax units.  (ii) We further tune the weights learned in the previous step by minimising the theoretical risk of the binary classifiers as proposed in~\citep{riskminim}. In order to apply the second step, we rely on two assumptions:
the rank of the class marginal is assumed to be known and the class-conditional linear scores are assumed to follow a Gaussian distribution. In ~\citep{riskminim}, this approach has been proven to converge towards the true optimal classifier risk.
We conducted experiments on the dialogue corpus released for the third dialogue state tracking challenge, namely DSTC3~\citep{thethirdstc3} and we show positive results for detecting rare slots as well as zero-shot slot-value pairs.

We use the unsupervised approach proposed in~\citep{riskminim} for risk minimisation (RM). 
We assume a binary classifier that associates a score $f_{\mathbf{W}_0}(\mathbf{h})$ to the first class 0 for the hidden unit $\mathbf{h}=(h_1,\cdots,h_{n})$ of dimension $n$:
$$f_{W_0}(\mathbf{h}) = \sum_i^{n} w_i h_i$$
where the parameter $w_i \in {\rm I\!R}$ represents the weight of the feature indexed by $i$ for class 0.

The objective of training is to minimize the classifier risk:
\begin{equation}\label{eq:r}
R(\mathbf{W})=E_{p(\mathbf{h},Y)}[\mathcal{L}(Y,f_\mathbf{W}(\mathbf{h}))]
\end{equation}
where $Y$ is the true label 
and $\mathcal{L}(Y,f_\mathbf{W}(\mathbf{h}))$ is the loss function.
The risk is derived as follows:
{\small{
\begin{equation}\label{eq:rr}
R(\mathbf{W})=\sum_{y\in\{0,1\}} P(y) \int_{-\infty}^{+\infty} P(f_\mathbf{W}(\mathbf{h})=\alpha|y)\mathcal{L}(y,\alpha)d\alpha
\end{equation}
}}

We use the following hinge loss:
\begin{equation}\label{eq:loss}
\mathcal{L}(y,\alpha) = \left( 1+\alpha_{1-y} - \alpha_y\right)_+
\end{equation}
where $(z)_+ = \max (0,z)$, and $\alpha_y=f_{\mathbf{W}_y}(\mathbf{h})$ is the linear score for the correct class $y$.
Similarly, $\alpha_{1-y}=f_{\mathbf{W}_{1-y}}(\mathbf{h})$ is the linear score for the wrong class.

Given $y$ and $\alpha$, the loss value in the integral (Equation~\ref{eq:rr}) can be computed easily. Two terms remain: $P(y)$ and $P(f_\mathbf{W}(\mathbf{h})=\alpha|y)$.
The former is the class marginal and is assumed to be known.
The latter is the class-conditional distribution of the linear scores, which is assumed to be normally distributed.
This implies that
$P(f_\mathbf{W}(\mathbf{h}))$ is distributed as a mixture of two Gaussians (GMM):
$$P(f_\mathbf{W}(\mathbf{h})) = \sum_{y\in\{0,1\}} P(y)\mathcal{N}(f_\mathbf{W}(\mathbf{h});\mu_y,\sigma_y)$$
where
$\mathcal{N}(z;\mu,\sigma)$
is the normal probability density function.
The parameters $(\mu_0,\sigma_0,\mu_1,\sigma_1)$ can be estimated
from an \textbf{unlabeled corpus} $\mathcal{U}$ using a standard Expectation-Maximization (EM) algorithm for GMM training.
Once these parameters are known, it is possible to compute the integral in Eq.~\ref{eq:rr} and thus an estimate $\hat R(\mathbf{W})$ of the risk
without relying on any labeled corpus. In ~\citep{riskminim}, it has been proven that: (i) the Gaussian parameters estimated with EM converge towards their true values, (ii) $\hat R(\mathbf{W})$ converges towards the true risk $R(\mathbf{W})$ and (iii) the estimated optimum converges towards the true optimal parameters, when the size of the unlabeled corpus increases infinitely. This is still true even when the class priors $P(y)$ are unknown.

The unsupervised algorithm is as follows:
\paragraph{\scriptsize{Unsupervised tuning for the binary classifier $c$, where $c=1,...,C$}}
{\small{
\begin{algorithmic}[1]\raggedright
\scriptsize{
\State \textbf{input:}  $\mathbf{h}$  the top hidden layer and the weights $\mathbf{W}^c$, as trained by the  deep learning decoder (Section~\ref{s:dlsemi}). 
\State \textbf{output:} The tuned weights $\hat{\mathbf{W}^c}$
\Repeat

\For{every index $i$ in $\mathbf{h}$} , 
\State Change the weights $\mathbf{W}^c_i=\mathbf{W}^c_i+\delta$, 
\State Estimate the Gaussian parameters using EM 
\State Compute the risk (Eq.~\ref{eq:rr})\footnote{A closed-form is used to compute the risk for binary classifiers.~\citep{rojasbarahona:hal-01184849}} on the unlabeled corpus $\mathcal{U}$ (i.e. the evaluation set). 
\State Compute the gradient using finite differences
\State Update the weights  accordingly $\hat{\mathbf{W}^c_i=\mathbf{W}^c_i}$
\EndFor
\Until{convergence}
}
\end{algorithmic}
}
}


\subsubsection{Experiments}
The supervised and unsupervised models are evaluated on  DSTC3~\citep{thethirdstc3} using the macro F-Measure\footnote{The macro F-score was chosen because we are evaluating the capacity of the classifiers to predict the correct class and both classes positive and negative are equally important for our task. Moreover, being nearly zero-shot classifiers, it would be unfair to evaluate only the capacity of predicting the positive category.}. 
We compare then three distinct models, (i) independent neural models for every binary classifier; (ii) neural models optimised jointly and (iii) further tuning of the weights through RM.

\paragraph{Dataset}
As displayed in Table~\ref{t:dstc3} in DSTC3 new slots were introduced relative to DSTC2. The training set contains only a few examples of these slots while the test set contains a large number of them.  Interestingly, frequent values per slots in the trainset such as \textit{area=north}, are absolutely absent in the testset. 
In DSTC3 the dialogues are related to restaurants, pubs and coffee shops. 
The new slots are: \textit{childrenallowed}, \textit{hastv}, \textit{hasinternet} and \textit{near}. Known slots, such as \textit{food}, can have zero-shot values as shown in Table~\ref{t:dstc3vals}. The corpus contains $3246$ dialogues, $25610$ turns in the trainset and $2264$ dialogues, $18715$ turns in the testset.

\begin{table}[htb]
\begin{subtable}[t]{.5\textwidth}
\caption{Frequency of slots in DSTC3.}
\centering
  \begin{tabular}{l | ll }
 \hline
 \scriptsize{Slot}&\scriptsize{$\#$Train}&\scriptsize{$\#$Test}\\
 \hline
 \scriptsize{\textbf{hastv}}&\scriptsize{1}&\scriptsize{239}\\
 \scriptsize{\textbf{childrenallowed}}&\scriptsize{2}&\scriptsize{119}\\
 \scriptsize{\textbf{near}}&\scriptsize{3}&\scriptsize{74}\\
 \scriptsize{\textbf{hasinternet}}&\scriptsize{4}&\scriptsize{215}\\
 \scriptsize{area}&\scriptsize{3149}&\scriptsize{5384}\\
 \scriptsize{food}&\scriptsize{5744}&\scriptsize{7809}\\
 \hline
 \end{tabular}
 
 \label{t:dstc3}
 \end{subtable}
\begin{subtable}[t]{.5\textwidth}
 \caption{Some zero-shot values per slots in DSTC3.}
\centering
  \begin{tabular}{ll | ll }
 \hline
 \scriptsize{Slot}&\scriptsize{Value}&\scriptsize{$\#$Train}&\scriptsize{$\#$Test}\\
 \hline 
 \scriptsize{near}&\scriptsize{trinity college}&\scriptsize{0}&\scriptsize{5}\\
 \scriptsize{food}&\scriptsize{american}&\scriptsize{0}&\scriptsize{90}\\
  \scriptsize{food}&\scriptsize{chinese takeaway}&\scriptsize{0}&\scriptsize{87}\\
   \scriptsize{area}&\scriptsize{romsey}&\scriptsize{0}&\scriptsize{127}\\
   \scriptsize{area}&\scriptsize{girton}&\scriptsize{0}&\scriptsize{118}\\
   
 \hline
 \end{tabular}

 \label{t:dstc3vals}
 \end{subtable}
 \end{table}



 \begin{table}[htb]
 \begin{subtable}[t]{.5\textwidth}
  \caption{Results for learning rare slots on DSTC3 evaluation set.}
 \centering{
 \begin{tabular}{l|l}
 \hline
 \multicolumn{2}{c}{\scriptsize{Deep Learning Independent Models}}\\\hline
 \scriptsize{Slot}&
 \scriptsize{F-Measure}\\\hline
 \scriptsize{childrenallowed}
 &\scriptsize{$49.84$\%}\\
 \scriptsize{hastv}&
 \scriptsize{$49.68$\%}\\
 \scriptsize{hasinternet}&
\scriptsize{$49.72$\%}\\
 \scriptsize{near}&
 \scriptsize{$49.90$\%}\\\hline
\multicolumn{2}{c}{\scriptsize{Deep Learning Joint Optimisation}}\\\hline
 \scriptsize{childrenallowed}&
 \scriptsize{$58.76$\%}\\
 \scriptsize{hastv}&
 \scriptsize{$59.16$\%}\\
 \scriptsize{hasinternet}&
 \scriptsize{$58.77$\%}\\
 \scriptsize{near}&
 \scriptsize{$56.65$\%}\\
 \hline
\multicolumn{2}{c}{\scriptsize{Risk Minimisation Tuning}}\\\hline
 \scriptsize{childrenallowed}&
 \scriptsize{$\mathbf{61.64}$\%}\\
 \scriptsize{hastv}&
 \scriptsize{$\mathbf{61.35}$\%}\\
 \scriptsize{hasinternet}&
 \scriptsize{$\mathbf{60.87}$\%}\\
 \scriptsize{near}&
 \scriptsize{$\mathbf{58.60}$\%}\\\hline  
 \end{tabular}}

 \label{t:dlsemisres}
 \end{subtable}
 \begin{subtable}[t]{.5\textwidth}
  \caption{Results for learning zero shot slot-value pairs on DSTC3 evaluation set.}
 \centering{
 \begin{tabular}{ll|l}
 \hline
 \multicolumn{3}{c}{\scriptsize{Deep Learning Independent Models}}\\\hline
 \scriptsize{Slot}&\scriptsize{Value}&
 \scriptsize{F-Measure}\\\hline
 \scriptsize{near}&\scriptsize{trinity college}&
 \scriptsize{$49.99$\%}\\
 \scriptsize{food}&\scriptsize{american}&
 \scriptsize{$49.88$\%}\\
 &\scriptsize{chinese take away}&
 \scriptsize{$49.88$\%}\\
 \scriptsize{area}&\scriptsize{romsey}&
 \scriptsize{$49.83$\%}\\
 &\scriptsize{girton}&
 \scriptsize{$49.84$\%}\\
 \hline
\multicolumn{3}{c}{\scriptsize{Deep Learning Joint Optimisation}}\\\hline
\scriptsize{near}&\scriptsize{trinity college}&
\scriptsize{$61.25$\%}\\
 \scriptsize{food}&\scriptsize{american}&
 \scriptsize{$59.93$\%}\\
 &\scriptsize{chinese take away}&
 \scriptsize{$61.02$\%}\\
 \scriptsize{area}&\scriptsize{romsey}&
 \scriptsize{$\mathbf{51.30}$\%}\\
 &\scriptsize{girton}&
 \scriptsize{$\mathbf{55.19}$\%}\\
 \hline
\multicolumn{3}{c}{\scriptsize{Risk Minimisation Tuning}}\\\hline
\scriptsize{near}&\scriptsize{trinity college}&
\scriptsize{$\mathbf{62.08}$\%}\\
 \scriptsize{food}&\scriptsize{american}&
 \scriptsize{$\mathbf{62.52}$\%}\\
 &\scriptsize{chinese take away}&
 \scriptsize{$\mathbf{63.79}$\%}\\
 \scriptsize{area}&\scriptsize{romsey}&
 \scriptsize{$48.76$\%}\\
 &\scriptsize{girton}&
 \scriptsize{$51.45$\%}\\
 \hline
 \end{tabular}}

  \label{t:dlsemisvres}
 \end{subtable}
 \end{table}

 \paragraph{The Gaussianity Assumption}
 As explained in Section~\ref{s:rm}, the risk minimisation tuning assumes the class-conditional linear scores are distributed normally. We verified this assumption empirically on our unlabeled corpus $\mathcal{U}$ (i.e. DSTC3 testset) and we found that for the slots: \textit{childrenallowed}, \textit{hastv} and \textit{hasinternet} this assumption holds. However, the distribution for \textit{near} has a negative skew. When verifying the values per slot, this assumption does not hold for \textit{area}. Therefore, we can not guarantee this method will work correctly for \textit{area} values on this evaluation set. 
 
\subsubsection{Results}
Tables~\ref{t:dlsemisres} and \ref{t:dlsemisvres} display the performance of the models that predict slots and values respectively.  The low F-Measure in the independent models shown their inability to predict positive examples. The models improve significantly the precision and F-Measure after the joint-optimisation. Applying RM tuning results in the best F-Measure for all the rare slots (Table~\ref{t:dlsemisres}) and for the values of the slots \textit{food} and \textit{near} (Table~\ref{t:dlsemisvres}).
For \textit{area}, the joint optimisation improves the F-Measure but the improvement is lower than for other slots. The performance is being affected by its low cardinality (i.e. $20$), the high variability of new places and the fact that frequent values such as  \textit{north} and  \textit{east}, are completely absent in the test set. As suspected,  the RM tuning degraded the precision and F-Measure because the Gaussianity assumption does not hold for \textit{area}. However, RM will work well in larger evaluation sets because the Gaussian assumption will hold when the unlabelled corpus tends to infinite (please refer to ~\citep{riskminim} for the theoretical proofs).


\section{Dialogue Manager}
\label{c:conts:dm}
My work on dialogue management regards learning the reward function. First, I explored inverse reinforcement learning to infer the reward function from human conversations on the EmoSpeech dataset~\citep{rojasbarahona2014bayesian}. Second, I trained a predictor of the interaction quality to infer the reward function in the PyDial dialogue framework~\citep{rojas2020user}. Last but not least, I co-supervised a PhD thesis on imitation learning to solve the problem of the scarce reward signal in dialogue systems. Besides imitation learning~\citep{cordier2020diluted}, we also explored graph neural networks for handling policies in multi-domain and multi-task environments~\citep{cordier-etal-2022-graph}. Furthermore, we use both imitation and graph neural policies for few-shot learning~\citep{cordier2023few}.

\subsection{Bayesian Inverse Reinforcement Learning}
Inverse reinforcement learning (IRL) was defined in~\citep{ng2000algorithms} as the problem of recovering the reward function from experts' demonstrations.
It learns an optimal reward, which leads to a decision policy that follows as closely as
possible the examples provided by experts, while maximising the expected accumulated reward in the long run.

In~\citep{ramachandran2007bayesian} we used Bayesian Inverse Reinforcement Learning (BIRL) to infer human behaviour in the context of the Emospeech serious game (Section~\ref{ss:nluemospeech}), given evidence in the form of stored dialogues provided by experts, who played the role of several conversational agents in the game.
We also reduce the computational complexity in large state spaces by using the approach proposed by~\citep{michini2012improving}. Instead of designing in advance the reward function to ``properly instruct players", which is 
a difficult and subjective task, we rather propose to learn it from humans.

We evaluated BIRL in terms of policy loss~\citep{michini2012improving} and is compared against two baselines. The first one uses random rewards, while the second one exploits corpus-estimated locally optimal rewards (i.e., supervised learning).
The results show that the proposed approach converges relatively quickly and consistently
outperforms both baselines. This suggests that taking into account the dynamic properties of the
environment leads to virtual characters that better reproduce the behaviour of experts.
Qualitatively, our models have thus learned to adequately inform users and provide help when needed.

\subsubsection{States, Actions and Transitions}

As shown in Table~\ref{t:dials}, there are 12 distinct conversations in the game between 7 virtual characters (VC) and 3 players.  Each of these dialogues talks about mandatory and optional goals.
The player either asks for information about these goals or asks for help.
Accordingly, the virtual character either informs about the goals or provides help. It can also handle out of domain topics, misunderstandings or request information (see example of dialogue in Figure~\ref{f:emodial}). 

We designed coarse-grained states containing user and system contributions to the dialogue; either by explicitly asking about the domain specific tasks (i.e. the dialogue goals) or by producing general dialogue acts (e.g., greeting, asking for help, acknowledgments, etc). 
A binary variable that indicates whether the dialogue has finished is also included. With this state representation we have $32$ states for the shortest dialogue (the first dialogue in Table~\ref{t:dials}), 
 and $432$ states for the longest dialogue (i.e., the third dialogue in Table~\ref{t:dials} with 5 goals).

 \noindent
 \paragraph{State variables}
\begin{enumerate}
 \item Has any of the characters ended the dialogue with a farewell action? : 1 for setting a terminal state, 0 otherwise.
 \item The last goal either informed or requested by the system: 0 when the system has not informed/requested about any goal, otherwise the goal id (e.g., from 1 to up to 5 for the longest dialogue).
 \item The last goal either asked or confirmed by the player: 0 when the user has not yet asked/confirmed about any goal, otherwise the id of the goal (e.g., from 1 to up to 5 for the longest dialogue).
 \item The last general dialogue act produced by the system: 0 for absence of general dialog act, 1 when providing help, and 2 when asking the player about the task to be solved (e.g., "How may I help you").
 \item The user has asked for help: 0 if the user has not asked for help, 1 otherwise.
\end{enumerate}

\noindent
\paragraph{Actions}

We are considering only the following actions in our experiments. 

\begin{itemize}
\item \textit{quit}:  farewell greeting.
\item \textit{inform(do($g_i$))}: informing about how to achieve goal $g_i$.
\item \textit{inform(help)}:  providing help
\item \textit{ask(task(X))}: Asking the player about the task, it corresponds to a general welcome sentence (e.g., "How may I help you"). Note that this action neither occurs in dialogue 1 nor in dialogue 7.
\item \textit{WAIT}: the system gives the turn back to the user.
\item \textit{ack}: the system acknowledges understanding.
\item \textit{other}: the system answers to out of context turns. 

\end{itemize}
Virtual characters always greet the player at the beginning; thus we do not need to learn this behaviour.

\noindent
\paragraph{Transition Function}

The transition function is not deterministic when the next state reflects an (unpredictable) user action.
This is typically the case after the WAIT system action.
However, BIRL requires this transition function to be given, and we have thus estimated such non-deterministic transition probabilities
using smoothed counts from the observed corpus as follows:
\[
\begin{array}{l}
P(s'|s,a)=\frac{N(s,a,s')+\alpha}{N(s,a)+N_{\chi}\alpha}
\end{array}
\]
Where $N(s,a,s')$ and $N(s,a)$ are respectively the number of times the transition $(s,a,s')$ and the state-action pair $(s,a)$ have been observed in the corpus, and
$N_\chi$ is the number of observed state-action pairs.
$\alpha$ is a smoothing constant arbitrarily set to 0.1.

The other transitions that reflect a system action are deterministic and have been defined as:
\[
\begin{array}{ll}
\multirow{2}{*}{$P(s'|s,a)=\Bigg\{$} & 1, \hspace{0.1cm} \mbox{if} \hspace{0.2cm} s'=next\_s(s,a)\\ 
&0, \hspace{0.1cm} \mbox{otherwise}
\end{array}
\]
Where $next\_s(s,a)$ is a function that computes the next state given a system action $a$. 
For instance, when the system informs about the first goal, $g_1$, the action $a_t=inform(do(g_1))$ yields
the next state $s'$ to have the state variable 2 set to 1. 

\subsubsection{Bayesian Inverse Reinforcement Learning}
\label{sec:birl}

The IRL problem as defined in~\citep{ng2000algorithms} is described as follows:
given a finite state space S, a set of actions $A = \{a_1 , a_2 , . . . a_k \}$, a transition probability $P^a_{ss'}$ , a discount factor $\gamma$, and a policy $\pi$,
determine a set of possible reward functions $R$ such
that $\pi$ is the optimal policy for the given MDP. The IRL problem is an ill-posed problem~\citep{abbeel2004apprenticeship}, because potentially
an infinite number of rewards may be optimal.
Bayesian IRL approaches model this uncertainty by inferring the posterior distribution of the reward vector $\textbf{R}$,
treating the demonstration sequences as the evidence and relying on a prior on the reward function~\citep{ramachandran2007bayesian}.

The IRL agent receives a sequence of observations of the expert's behaviour:

$O_\chi = \{(s_1,a_1),(s_2,a_2),...,(s_k,a_k)\}$, which means that at time step $i$,
the virtual character $\chi$ that mimics the expert is in state $s_i$ and takes the action $a_i$. After applying Bayes Theorem, the posterior
can be written as:
\begin{equation}
  Pr(\textbf{R}|O_\chi) = \frac{Pr(O_\chi|\textbf{R})Pr(\textbf{R})}{Pr(O_\chi)}
  \label{e:posterior}
\end{equation}

We model next the reward function by a simple $n$-dimensional real vector, where $n$ is the number
of different states.
Then, $Pr(\textbf{R}|O_\chi)$ is the posterior distribution of the reward vector given the observed state-action pairs of the expert. $Pr(O_\chi|\textbf{R})$ is the likelihood of the observed expert state-action pairs given
the reward vector $\textbf{R}$.
This likelihood is modelled in~\citep{ramachandran2007bayesian} with a parameter $\alpha$ representing the degree of confidence we have in the expert's ability
to choose a good action as follows:

\begin{equation}
 Pr(O_\chi|\textbf{R})= \frac{1}{Z}e^{\alpha\sum_iQ^*(s_i,a_i,\textbf{R})}
 \label{e:birlposterior}
\end{equation}

$Pr(\textbf{R})$ is the prior distribution and $Pr(O_\chi)$ is the probability of the evidence over the entire space of reward vectors $\textbf{R}$, which is not needed in the BIRL algorithm.
The original BIRL algorithm, namely PolicyWalk, follows a Markov Chain Monte Carlo (MCMC) technique iterating as follows: Given a reward vector $\textbf{R}$, it performs random walks over the neighbours of $\textbf{R}$ on a grid of length $\delta$,
 finding a new proposal $\bar{\textbf{R}}$, such that: $\bar{\textbf{R}}(s)=\textbf{R}(s) \pm \delta$. The proposal is accepted with probability $\min\{1, \frac{Pr(\bar{\textbf{R}}|O)}{Pr(\textbf{R}|O)}\}$, where the 
 posterior is given by Eq~(\ref{e:posterior}).
 
 The expected value of the reward given this posterior is then computed over all these samples.
 Note that the normalising constants cancel out in the ratio used to accept the proposed $\bar{\textbf{R}}(s)$ and that finding $Q^*$ in Eq~(\ref{e:birlposterior}) requires to solve the MDP at every MCMC iteration.
This can be done for example with the policy iteration (PI) algorithm~\citep{sutton2018reinforcement}.

BIRL converges slowly when applied to large state spaces. One reason for this is that it infers the reward of every state, although many states have little expert evidence.
Second, searching over a reward function space easily increases the number of MCMC iterations needed to approximate the mean of the posterior.
To solve these limitations, ~\citep{michini2012improving} proposed a modified BIRL (MBIRL) that:
\begin{itemize}
 \item infers only those states that are similar to the observed ones according to a \textit{kernel-based relevance function}.
 \item uses simulated annealing to focus the sampled distribution around its maximum, hence reducing the number of samples needed to converge. Therefore,
they use a modified acceptance probability of $\left(\frac{Pr(\bar{\textbf{R}}|O)}{Pr(\textbf{R}|O)}\right)^{\frac{1}{T_i}}$ where $T_i$ is a decreasing \textit{cooling schedule}.
\end{itemize}

\subsubsection{Experiments} 
\label{sec:exps}
We introduce the baselines, the evaluation metrics, and the experiment setup for 12 dialogues in the game.  

\paragraph{Baselines}
We evaluate the performances of the proposed system by comparing it with two baselines:
\begin{itemize}
\item Using random rewards (RR);
\item Exploiting ''locally-estimated`` rewards (LR), i.e., rewards that are trained on the corpus with the additional assumptions that
the reward prior $Pr(\textbf{R})$ is uniform, that the states are conditionally independent given
the reward $P(O_\chi|\textbf{R}) = \prod_i P(s_i|\textbf{R})$ and that the state likelihood is multinomial with parameters representing the reward $P(s=k|\textbf{R})=R_k$,
so that the path that maximizes the cumulated reward also maximizes the likelihood. Then:
\begin{eqnarray}
\nonumber
&& \arg\max_R Pr(\textbf{R}|O_\chi) = \arg\max_R Pr(O_\chi|\textbf{R})\\
&& =\arg\max_R \prod_i P(s_i|\textbf{R})
\end{eqnarray}
Let $n_k$ be the number of times the $k^{th}$ state occurs in the expert observations:
$n_k = |\{(s_i=k,a_i)\}_{i\in O_\chi}|$

Then we want to maximize the likelihood
$\prod_k P(s=k|\textbf{R})^{n_k}$
under the constraint $\sum_k R_k = 1$, which gives the locally optimum reward:
$$\hat R_k = \frac {n_k}{N_\chi}$$
with $N_\chi = |\{(s_i,a_i)\}_{i\in O_\chi}|$ the number of observed state-action pairs.
	\end{itemize}

\paragraph{Evaluation Metrics}
\label{ssec:metrics}
 We consider two evaluation metrics: the policy loss~\citep{michini2012improving} and the system training time.
\begin{itemize}
  \item \textit{Policy loss}: The policy loss is the ratio $\frac{n_{\neq}}{N_\chi}$, where $n_{\neq}=|\{(s_i,a_i \neq \pi(s_i))\}_{i\in O_\chi}|$ is the number of expert state-action pairs that disagree with the learned policy $\pi$ and
  $N_\chi = |\{(s_i,a_i)\}_{i\in O_\chi}|$ is the number of observed state-action pairs.
  \item \textit{Elapsed time}: The time in milliseconds it takes to MBIRL and to the policy iteration algorithms to finish.
\end{itemize}

\label{sec:results}

\begin{figure}[ht!]
    \centering
    \includegraphics[page=1,width=1.0\textwidth]{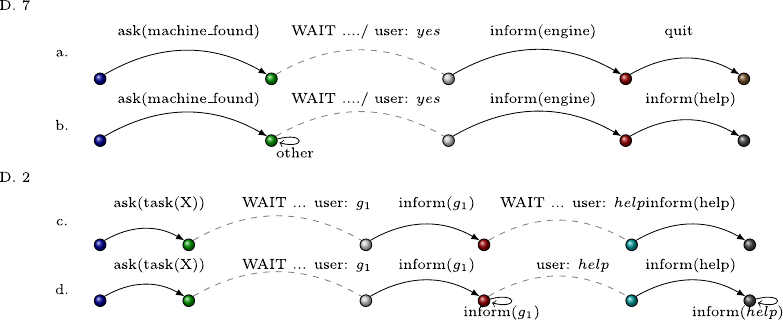}
    \caption{Comparison of expert vs. MBIRL trajectories for dialogues 7 (top) and 2 (bottom). (a) and (c) depict expert trajectories, while (b) and (d) show the 
trajectories of MBIRL optimal policy $\pi$.}
    \label{fig:traj}
    
\end{figure}

\paragraph{Performance:}
Two important issues affect performance: the size of the state-space and the limited number of expert observations.
In general MBIRL outperforms both locally-optimal and random rewards.
However, with a larger state-action space such as in dialogue 3, 4 and 8, the models do not improve over the locally-optimal reward, which suggests that the number of samples that are generated
is not large enough.
Moreover, for state spaces greater than 300 states, MBIRL takes a prohibitively long running time to finish. 
The huge computational expense for large state spaces is an important limitation of MBIRL since it needs to solve one RL problem per iteration. 
A potential solution to this issue might be to use appropriate function approximation both for the $Q$
function and for modelling the reward function \textbf{R}, but this is left for future work.

Unsurprisingly, the MBIRL policy loss is higher for optional dialogues (such as dialogue 10 and 12) showing that the 
scarce number of observations significantly affects performance. 

\paragraph{Trajectories:}
Figure~\ref{fig:traj} shows two dialogue trajectory excerpts with both the gold (or expert) trajectory and
the trajectory inferred by MBIRL.
Interestingly,
in most of the dialogues, both trajectories coincide in the first state and in those states where the system has to inform about mandatory goals just after explicitly requested by the user.
This is also the case of the states where the system properly provides help as requested by the user.
On the other hand, the learned policy usually fails to close the dialogue and it sometimes contains repetitions e.g., once it has informed about a goal, it may inform again later on.

For more details please refer to~\citep{rojasbarahona2014bayesian}.
\subsection{Is the User Enjoying the Conversation?}
The impact of user satisfaction in policy learning for task-oriented dialogue systems has long been a subject of research interest~\citep{walker1997paradise,schmitt2011modeling,ultes-etal-2015-quality,ultes-2019-improving}.
Similarly, sentiment analysis has been widely adopted to analyse massive blogs, recommendations tweets and reviews~\citep{rojas2016deep,do2019deep}. Although sentiment analysis can be used to infer user satisfaction, most of the work that incorporates sentiment analysis in dialogue focused on the generation of empathetic responses in end-to-end chitchat dialogues~\citep{lee2018scalable,ma2020survey}. Moreover, most sentiment analysis solutions focus on the analysis of out-of-context short texts~\citep{rojas2016deep}.  In this work we are interested in the study of user satisfaction for measuring the quality of the interaction in task-oriented dialogues, in which dialogue is modelled as a POMDP~\citep{young2013pomdp}. 
    
First we study distinct neural networks that use distributed representations for predicting satisfaction scores. At this stage we would like to answer the following question: does relying only on distributed representations improve the performance of neural models? Therefore, we evaluate the performance of hierarchical networks and state-of-the-art Transformers. Second, we evaluate the impact of using the best trained network for computing the reward function within a POMDP dialogue framework~\citep{ultes2017pydial}. We would like to determine how realistic it is to incorporate user satisfaction estimators that rely solely on distributional semantics in reinforcement learning (RL) dialogue systems. This approach can be used for instance to train satisfaction predictors from large human-human chats, in which satisfaction has been self-scored by users.

The case study is the English LEGO corpus of human-machine spoken conversations~\citep{schmitt-etal-2012-parameterized}, which has been annotated at each system turn with the \ac{IQ}, ranging from 1 (poor quality) to 5 (good quality). Our results suggest that distributed representations do outperform state-of-the-art models trained on fine-tuned features. We also show that using \ac{IQ} estimators in the reward function greatly improves the task success rate for dialogues in the same domain the networks were trained on, which in this case is the Let's Go domain~\citep{raux2005let}. 

\subsubsection{Networks for Estimating the User Satisfaction}
\label{estimators}
We study three distinct neural networks:  hierarchical bi-directional Gated Recurrent Units(GRUs)~\citep{cho-etal-2014-learning} with attention , Transformers for generating contextual embeddings that feed a GRU layer and solely Transformers~\citep{vaswani2017attention}. We are interested in studying the impact of context-length in transformers, because real dialogues can easily attain a context of thousands of tokens (Table~\ref{t:corp}). Therefore, we explore BERT~\citep{devlin2018bert}, DistilBERT~\citep{sanh2019distilbert} and Transformers eXtra-Large (Transformers-XL)~\citep{dai2019transformer}.

\paragraph{\textbf{(i) Hierarchical GRUs:}}  Figure~\ref{f:bigru}(a) shows the network. It has a Bidirectional GRU layer (BiGRU) at the lower level that returns the turn representation $h_{tk}$ 
Attention is used to weight relevant units in the turn hidden representation. Then, a GRU layer is then used to process dialogues as a sequence of turns. The last layer is a Softmax that predicts the most probable \ac{IQ} class from $0$ up to $5$.

\begin{figure}[ht]
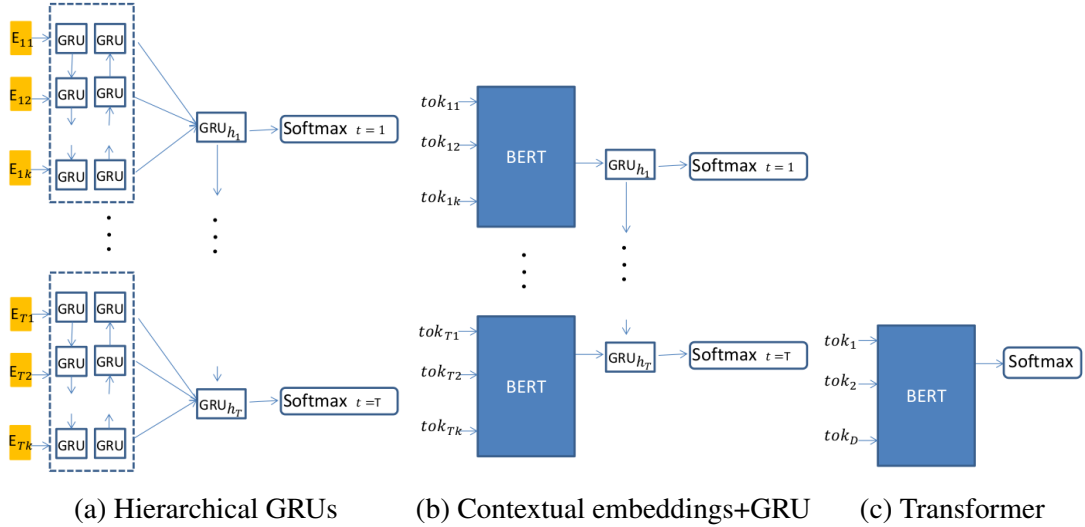

\centering
\subfloat[Hierarchical GRUs]{\includesvg[scale=0.4]{images/BiGRUs.svg}}
\subfloat[Contextual embeddings+GRU]{\includesvg[scale=0.4]{images/BERT_GRUs.svg}}
\subfloat[Transformer]{\includesvg[scale=0.4]{images/Trans.svg}}

\caption{The proposed neural architectures for predicting the user satisfaction. $E_{Tk}$ in (a) is the embedding for the $k$th token of the last turn ($t=T$). In (b) $tok_{T1}=\mbox{[CLS]}$ and  $tok_{Tk}=\mbox{[SEP]}$ after applying the WordPiece tokenization to the last turn ($t=T$). In (c) $tok_1,...,tok_D$ are the tokens of the dialogue after applying the WordPiece tokenization, in which $tok_1=\mbox{[CLS]}$ and the token [SEP] marks turn separation. $tok_{D-1}$ is the last token of the last utterance in the dialogue and $tok_D=\mbox{[SEP]}$.}
\label{f:bigru}
\end{figure}

\paragraph{\textbf{(ii) Contextual embeddings + GRU:}} We investigate the use of transformers as turn representations thus we propose a BERT-like transformer, which inputs are the tokens of the turn. The turn representations then feed a GRU layer.  The output of the GRU then feeds a Softmax layer for predicting the score.

\paragraph{\textbf{(iii) Transformer:}} This network is depicted in Figure~\ref{f:bigru}(c). It consists in a transformer that takes as input the tokens {$tok_1,...,tok_D$} of the previous and current utterances. Then the output [CLS] of the transformer feeds a Softmax layer for predicting the score of the current utterance. 

We evaluate the prediction $y_t$ at each system turn $t$. The back-propagation optimisation is done by minimising the cross-entropy loss function~\cite{tieleman2012lecture} through stochastic gradient descent.

\subsubsection{The Reward Function in a POMDP Dialogue System}
\label{pomdp}

Figure~\ref{f:iqachictecture}, shows the architecture connected to a reward estimator. In the Let's Go bus-scheduled information system, the slots might be \attr{origin} for the bus departure place and \attr{time} for the departure time. The state $s$ might record the current value and confidence level of each slot. From the state, a {\it belief state} $b$ is extracted and an action $a$ is decided based on a dialogue {\it policy}.  Once the appropriate action is determined, it is converted to a textual message and then rendered by a speech synthesiser.
\begin{figure}[ht]
	\centerline{\includesvg[inkscapelatex=false, scale=0.5, keepaspectratio]{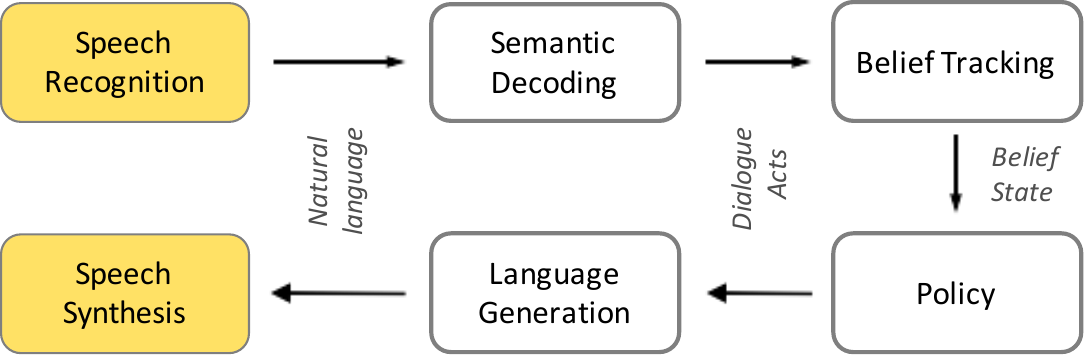}}
	\caption{\small\noindent A POMDP spoken dialogue system with the interaction quality estimator.}
	\label{f:iqachictecture}
\end{figure}

The reward function most commonly adopted for task-oriented dialogues penalises every dialogue turn with $-1$ and sums a reward of $+20$ at the end of the dialogue whenever the system provided the right information to the user or $0$ otherwise (Eq.~\ref{eq:tsrwd})~\citep{gasic2013gaussian}.
\begin{equation}
    R_{TS}=T\cdot(-1)+\mathbbm{1}_{TS}\cdot 20
    \label{eq:tsrwd}
\end{equation}

In this work the reward estimator is based on the IQ as defined in~\citep{ultes-2019-improving}.

\begin{equation}
    R_{IQ}=T\cdot(-1)+(iq-1)\cdot 5
    \label{eq:iqrwd}
\end{equation}
Where $R_{IQ}$ describes the final reward, $iq$ is the \ac{IQ} value predicted by the classifier (Section~\ref{estimators}), which is a number from 1 to 5, where 1 represents poor quality and 5 good quality.

We used PyDial~\citep{ultes2017pydial}, the publicly available POMDP dialogue framework and we implemented an application programming interface (API) that returns the \ac{IQ} estimation predicted by the neural models presented above.
Usually, RL systems first learn the policy on a simulated user until an optimal performance is reached, then they are ready to be tested by humans.
\subsubsection{Experiments}
\label{exps}
In this section we introduce the corpus as well as describe the experiments and the evaluation metrics.

\paragraph{The Dataset}
The LEGO corpus collects \textit{spoken} dialogues between users and the Let's Go dialogue system~\cite{raux2005let}, which provides bus schedule information to the Pittsburgh population during off-peak times. 400 dialogues in the corpus have been manually annotated with the \ac{IQ} score~\cite{schmitt-etal-2012-parameterized}. Since conversations in LEGO are system-initiative they have a lot of system interactions such as misunderstandings, confirmations and repetitions, producing quite long dialogues (i.e. hundreds of turns).

\begin{table}[htbp]

\centering
  \begin{tabular}{c|c|c|c}

  N. Dialogues&Dialogue length
  &Max.turn length & Max. toks p/dial.\\[-1ex]
  &\scriptsize{(max/mean/median)}&\scriptsize{(max/mean/median)}&\scriptsize{(max/mean/median)}\\\hline
  400 & $200/65/53$& $76/26/24$& $6590/1444/1089$\\
 \end{tabular}
   \caption{The LEGO corpus with \ac{IQ} annotations. Its complexity is measured by the maximum dialogue length (number of turns per dialogue), the maximum turn length (number of tokens per turn) and the maximum number of tokens per dialogue.}
 \label{t:corp}
 \end{table}

The complexity of the corpus is presented in Table~\ref{t:corp}, containing long dialogues of up to 200 turns with up to 76 tokens per turn.

\paragraph{User Satisfaction Estimators}
\label{ss:us}


We compared our networks with the networks presented in~\cite{ultes-2019-improving} for predicting the \ac{IQ} with the following evaluation metrics: the unweighted average recall (UAR), which is the arithmetic average of all class-wise recalls, as well as a linearly weighted version of Cohen's $\kappa$ and Spearman's $\rho$. The experiments were conducted in a 10-fold cross-validation, assuring that the same dialogue did not slip into different folds (i.e. dialogue-wise cross validation).  We studied the length of the \textbf{dialogue context}: the turn for which we are predicting the score and the previous turns. We vary the context length and find an optimal context length of up to 100 turns per dialogue for BiGRUs.

Table~\ref{t:iq} shows that the BiGRUs network trained on word embeddings outperformed the state-of-the-art (BiLSTM+att) networks in all performance measures, obtaining an absolute improvement of $+1$ for UAR, $+9$ for $\kappa$ and $+2$ for $\rho$. It is worth noting that the state-of-the-art (BiLSTM+att) networks were trained on fine-tuned \textit{turn features}. 

These results are encouraging and suggest that distributed representations impact positively the performance of satisfaction estimators in hierarchical networks. We would like to study in the next section whether these models can be used to predict task success in dialogue systems.

\begin{table}[htbp]

\centering
  \begin{tabular}{l | lllll  }
 \hline
 \multicolumn{4}{c}{Predicting IQ}\\\hline
 Model&UAR&$\kappa$&$\rho$\\\hline
SVM\_feats~\cite{ultes-2019-improving} &$44\%$&$53\%$&$69\%$\\
BiLSTM+att\_feats~\cite{ultes-2019-improving} &$54\%$&$65\%$&$81\%$\\
BiGRUs&$\mathbf{\mathbf{55}}\%$&$\mathbf{74}\%$&$\mathbf{83}\%$\\
DBert+GRU&$35\%$&$57\%$&$42\%$\\
Trans-XL(ctxt$\approx$1K)&$47\%$&$59\%$&$67\%$\\
Trans-XL(ctxt$\approx$2K)&$44\%$&$58\%$&$67\%$\\
 
 \hline
 \end{tabular}
 \caption{Performance of the proposed models. The BiGRUs with FastText embeddings outperform all the networks trained on fine-tuned \textit{turn features}}.
 \label{t:iq}
 \end{table}

BERT-based Transformers do not perform well for this task on this dataset when using them to get the turn representations. This can be explained by the large number of turns dialogues have (i.e., up to $200$ turns), the large number of parameters a transformer needs and the quite short annotated dataset ($\approx$ 400 dialogues). We first tried (BERT+GRU) and due to its large memory requirement, we could only learn weights of up to $7$ dialogue turns per dialogue in a cluster of 32GB-GPU machines (all the other turns representations were frozen). However, with DistilBERT we could treat up to $15$ turns per dialogue while maintaining the same performance. Fortunately, we could process larger contexts with Transformers-XL, reaching an optimal performance with $\approx 1K$ dialogue tokens. The results of Transformers-XL are comparable with the SVM baseline trained on fine-tuned features, yielding a better UAR ($+3$) and $\kappa$ ($+6$) as well as a slightly lower $\rho$ ($-2$). Having larger contexts $\approx 2K$  do not seem to impact significantly their performance.
These results suggest that in 32GB-GPU nodes the large context length (i.e. up to $6.5$K tokens per dialogue) is affecting transformers performance as they will require a prohibited usage of GPU memory to process the whole context.

\paragraph{The Impact on the Reward Function}

We evaluated the impact of the \ac{IQ} estimators presented in Section~\ref{ss:us} on the reward function for the Let's Go (LetsGo) domain by using PyDial~\citep{ultes2017pydial}.

 The Let's Go dialogue system provides information about bus time-schedule according to the constraints: \attr{origin}, \attr{destination}, \attr{time} and  \attr{route}, which corresponds to LetsGo(4). LestGo(6) also considers \attr{origin neighbourhood} and \attr{destination neighbourhood}. It is important to note that the Let's Go domain is far more complex in terms of the number of database items than other domains available in PyDial.
 
 The experiments run on simulated dialogues as in~\citep{ultes-2019-improving,Casanueva2017}.
 We implemented a template-based generator for the user and the system utterances for the Let's Go domain because our models rely on textual inputs (i.e. distributed representations).  
 We compared our models in an environment without noise because unlike~\citep{Casanueva2017} and~\citep{ultes-2019-improving} the simulator in this work runs at the surface-level and not at the semantic-level and the noise used for User-Simulation in PyDial alters the semantic-dialogue acts regardless the surface form. In addition, we would like to apply these methods to chatbots, thus simulating ASR noise would not be appropriate and studying a more appropriated noise is out of the scope of this paper.
 
 \begin{table}[htbp]
\centering
  \begin{tabular}{l| lll }
 \hline
 Domain&Reward&Task Success Rate($\uparrow$)& Average Turns($\downarrow$)\\\hline
 LetsGo(4)&$R_{TS}$&$\mathbf{99\%\pm1.4}$&$6.5\pm0.96$\\ 
 &$R_{IQ}$&$\mathbf{99\%\pm2.1}$&$\mathbf{6.3\pm0.8}$\\\hline
 LetsGo(6)&$R_{TS}$&$81\%\pm7.78$&$11\pm1.09$\\ 
 &$R_{IQ}$&$\mathbf{97\%\pm3.98}$&$\mathbf{8.9\pm1.26}$\\\hline
 \end{tabular}
  \caption{Task success rate of the simulated experiments for the Let's Go domain over three runs with distinct seeds. Each value is computed after $1000$ training dialogues/$100$ evaluation.}
  \label{t:tsuc}
 \end{table} 
 
 We used a policy model based on the GP-SARSA algorithm~\citep{gasic2013gaussian}, which is a sample efficient Gaussian process approximation to the value function. We used the focus tracker~\citep{henderson2014second} for belief tracking. The policy decides on summary actions of the dialogue state tracker which are based on dialogue acts (e.g., \textit{request}, \textit{inform} or \textit{confirm}). \textit{The task success rate} was the metric used to measure the dialogue performance~\citep{Casanueva2017}.
 
We observe in Table~\ref{t:tsuc} that the reward computed with $R_{IQ}$ outperforms the classical $R_{TS}$ reward when having more constraints, namely LetsGO (6). Moreover, dialogues rewarded by $R_{IQ}$ tend to be significantly shorter. Although there is not significant distinction between $R_{TS}$ and $R_{IQ}$ in terms of the task success for LetsGO (4), dialogues are slightly short with $R_{IQ}$.  
We also conducted preliminary experiments on domain transfer by evaluating $R_{IQ}$ on the Cambridge Restaurants domain, obtaining a success rate of $44\pm25.7$, compared to $99\pm1.83$ for $R_{TS}$. Unsurprisingly, $R_{TS}$ and feature-based $R_IQ$~\citep{ultes-2019-improving} are more robust to unknown domains than embedding-based $R_{IQ}$ because both task-success and quality features are domain-agnostic.
  
For more details, please refer to~\citep{rojas2020user}.

\subsection{Imitation Learning}
\label{ss:ilearning}
This section introduces the work of the PhD candidate Thibault Cordier, who I co-supervised together with Dr.Tanguy Urvoy and Professor Fabrice Lefevre. This work explores imitation learning for learning the policy on single domains. It has been published in the NeurIPS workshop Human in the loop dialogue systems~\citep{cordier2020diluted}.

Deep RL (DRL)~\citep{li_deep_2018} has achieved significant success on many complex decision-making problems and in particular in conversational AI~\citep{gao_neural_2018}.
The ability to learn from few interactions is essential in dialogue applications because human interactions are scarce and costly. Unfortunately, standard RL algorithms usually require a large amount of interactions with the environment to reach good performances.
One solution to speedup the learning process is to guide the agent’s exploration like stochastic learning policies such us soft-kind~\citep{haarnoja_reinforcement_2017,haarnoja_soft_2018,gao_reinforcement_2019}.
The first question we address here is: (i) can dialogue policy learning be improved with stochastic off-policy learning methods?

Several methods search a way to exploit demonstrations to accelerate the learning with the conviction that demonstrations are the solution to \textit{the sparse reward}~\citep{hester_deep_2017}. They can be based on Imitation Learning (IL) to learn an "optimal" policy function or on Inverse Reinforcement Learning (IRL) to learn an "optimal" reward function. 
Demonstrations may be useful to guide efficiently the exploration. We consider that human expertise can be used in different ways. The most classical one is to use demonstrations that humans have already produced. Another way is to use a rule-based agent, namely handcrafted, that has been designed, evaluated and fine-tuned by humans. Indeed,~\citep{casanueva_benchmarking_2017} have shown that handcrafted approaches still perform better than policy learning approaches.

Moreover it is well known that human interactions and manually crafted rules are not only costly but also time consuming; while simulated interactions are cheaper and easier to collect~\citep{su-etal-2016-line,schatzmann2006survey}.
Therefore, the second issue that raised in our work is: (ii) can we use demonstrations without supervision and only as a way to guide exploration in an on-line learning?

\subsubsection{Proposed Approach} \label{sec:approach}

The proposed approach to policy learning is based on Boltzmann sampling, and to which we integrate demonstrations directly into the RL process in order to better guide exploration and make relevant exploitation. RL is usually implemented as either value-based method as Q-learning or policy-based method as actor-critic. Both are our baselines with which we will make our contributions.

\paragraph{Baselines}

Q-learning is the combination of Double and Duelling Deep-Q-Network with Experience Replay (DQN/D3QN)~\citep{van_hasselt_deep_2015,wang_dueling_2016}. In short, DQN is an approximation function that searches to estimate the optimal state-action value function (or Q-value). The Double DQN architecture is an alternative method that mitigates the problem of overoptimistic value estimation and the Duelling DQN architecture is for learning more efficiently by decoupling value and advantage functions. On top of that, experience replay can be added for reducing sample correlation and for improving sample efficiency. 

The actor-critic baseline is the Trust Region Policy Optimisation for Actor-Critic with Experience Replay (A2C/ACER/TRACER)~\citep{wang_sample_2017,weisz_sample_2018}. In brief, an actor $\pi$ tries to maximise the expected reward when in the same time a critic $Q$ learned separately evaluates the actor decisions. 

The importance sampling truncation with bias correction is used to correct the perceived sampling distribution induced by experience replay in order to reduce variance. We apply the Retrace algorithm~\citep{munos_safe_2016} to recursively estimate the advantage function in safe and efficient way with small bias and variance. Finally, we use the trust region policy optimisation (TRPO)~\citep{schulman_trust_2017} for adjusting the policy gradient in order to learn in a safe parameters region limiting the deterioration of the policy performance. 

\subsubsection{Exploration Strategy}
\label{ss:explo}
The exploration strategy learns and plays a stochastic policy related to energy-based model. We propose to learn an energy-based policy in the dialogue environment where the agent samples his actions according to Boltzmann's stochastic sampling~\citep{haarnoja_reinforcement_2017}. We opt for using energy-based policies of the following form:

\begin{gather}
    \pi(a_t|b_t) \propto \exp(-\mathcal{E}(b_t,a_t))
\end{gather}

When using value-based method, the energy function can be represented by the Q-function with parameter $\tau$, the temperature, where we set $ \mathcal{E}(b_t,a_t) = -\frac{1}{\tau}Q^\pi(b_t,a_t)$. When using policy-based method, the energy function is directly represented by the policy network and so is learned implicitly.

A stochastic sampling can be achieved by the Boltzmann sampling. Contrary to the commonly used strategy as $\epsilon$-greedy where actions $a_t$ are sampled from 
$\argmax_a$
$(1-\epsilon) \argmax_a \pi(a|b_t)+ \epsilon U(\mathcal{A})$ 
where $U(\mathcal{A})$ is an uniform distribution over action space, the Boltzmann sampling strategy samples actions from $\exp(-\mathcal{E}(b_t,a_t))$, hence $a_t \sim \pi(a_t|b_t)$. 

One of its advantages is that policy learning is less influenced by the policy function changes. Conversely, the $\epsilon$-greedy sampling faces sudden jumps in action choices due to the argmax operator. So in theory, the Boltzmann strategy can make learning more stable than the $\epsilon$-greedy.

Another advantage is that the temperature parameter can control the exploration-exploitation balance. So, to counter the weakness of its exploitation, it can be interesting to define correctly the temperature. 

In practice, we decide to add random exploration in such a way that the actions are sampled according to $a_t \sim (1-\epsilon)\pi(a_t|b_t) + \epsilon U(\mathcal{A})$, namely $\epsilon$-Boltzmann sampling, with decreasing $\epsilon$ parameter in order to explore enough before following the stochastic policy with a fixed $\tau$ temperature parameter.

\subsubsection{Imitation Learning Strategies} 
\label{sss:il}

During the reinforcement learning process, demonstrations can serve as an efficient way to explore the environment. Indeed, they can lead the agent to receive rewards promptly and so can lead it to exploit confident winning trajectories quickly. A handcrafted agent is used to simulate near-optimal demonstrations and feed-backs. It offers good performance compared with the other deep learning methods. Also, it has been designed, evaluated and fine-tuned by humans. Thus it is a way to mimic human expertise and can be served as a near-optimal expert in our experimentation.

We propose two demonstration sampling strategies corresponding to two different ways of using the knowledge of an expert. 

\paragraph{A) Learning with demonstrations:} Let us assume that we learn with an offline expert i.e. demonstrations are given before learning. We propose that in $\beta$ (in percent) of dialogues, the agent plays for itself. Otherwise in $1-\beta$ of dialogues, the expert gives to the agent one of its expert trajectories as demonstration and the agent replays the dialogue as if it was the one who played it. Therefore, the agent learns as if they are two datasets. The first one contains its trajectories and the second the expert demonstrations.

\paragraph{B) Learning with feed-backs:} Let us assume that we learn with an online expert i.e. demonstrations are given during learning. We propose that in $\beta$ (in percent) of dialogue actions, the agent plays for itself. Otherwise in $1-\beta$ of dialogue actions, the expert gives to the agent its expert action as feed-back and the agent plays the dialogue action as if it was its choice. 

This technique let the agent explore the environment by playing relevant actions in a given state. In other words, it learns about its trajectories in which the expert can redirect it at any moment to more relevant action, as DAgger does~\citep{ross_reduction_2011}.


\subsubsection{Experiments} \label{sec:expres}

In our experiments the Pydial framework~\citep{ultes_pydial:_2017} is used, which implements an agenda-based user simulation~\citep{schatzmann_agenda-based_2007}. As in~\citep{casanueva_benchmarking_2017} we tested our algorithms for policy learning on different domains and in different environments by increasing the inputs' noise. The domains in PyDial differ from each other by the ontology size, impacting the state and action space dimensions.

We evaluate the learned policies according to three levels of noise with respect to the semantic error rate (SER). This corresponds to the noise that comes from the ASR and the NLU channels. In Pydial, this is modelled at the semantic level whereby the true user action is corrupted by noise to generate an N-best-list with associated confidence scores. 

Table~\ref{tab:tab1} shows the compared policy models. HDC corresponds to the handcrafted policy learning, which is a rule-based approach written by experts. DQN and ACER are the baselines enhanced with stochastic (stoc) exploration and either behaviour cloning (BC) or feedbacks (FB).


\begin{table}[hbt]

\centering
\resizebox{0.65\textwidth}{!}{%
\begin{tabular}{llr}
Method name & Abbrev. \\
\midrule \midrule
Handcrafted Policy & {HDC} \\
\midrule
Stochastic Q-learning & {Stoc-DQN} \\
Stochastic Q-learning with Demonstrations & {Stoc-DQN-BC} \\
Stochastic Q-learning with Feed-backs & {Stoc-DQN-FB} \\
\midrule
Stochastic Actor-Critic & {Stoc-ACER} \\
Stochastic Actor-Critic with Demonstrations & {Stoc-ACER-BC} \\ 
Stochastic Actor-Critic with Feed-backs & {Stoc-ACER-FB} \\
\bottomrule
\end{tabular}
}
\caption{Overview of proposed methods}
\label{tab:tab1}
\end{table}

We performed a long training stage over $10\,000$ dialogues. This will evaluate the contribution of stochastic sampling strategy during training and testing stages.  Here we search to answer the question: can we improve dialogue policy learning with stochastic off-policy learning methods in order to compete the handcrafted agent?
All methods are evaluated after training over $1\,000$ dialogues during which the learned policy is fixed. For the second experiments, we decide to compute the average performance over the last five checkpoints from dialogue indices $6\,000$ to $10\,000$ with a step of $1\,000$ dialogues. This calculation is done in order to reduce variance induced by RL when we estimate the performance of the models.

\subsubsection{Results} 

The results are presented in Table~\ref{tab:results_stoch}. In most environments, methods learn very well compared to the benchmarks~\citep{casanueva_benchmarking_2017}. Furthermore, some of them can compete with the handcrafted agent whether it is a stochastic Q-learning approach or a stochastic actor-critic approach. For instance, Stoc-DQN-BC outperforms handcrafted expert with $30\%$ SER for laptops and SFR. Stochastic ACER was more robust to the different environments, showing better performance, particularly for Stoc-ACER-BC with $30\%$ SER for laptops.

\begin{table*}[ht]

\centering
\resizebox{!}{0.8\height}{%
\begin{tabular}{ll||rrrrrr||rr}
 &  & \multicolumn{2}{c}{Stoc-DQN}  & \multicolumn{2}{c}{Stoc-DQN-BC} & \multicolumn{2}{c}{Stoc-DQN-FB}  & \multicolumn{2}{c}{HDC} \\ 
\multicolumn{2}{c}{\textit{Task}}  & \multicolumn{1}{c}{Suc.} & \multicolumn{1}{c}{Rew.} & \multicolumn{1}{c}{Suc.} & \multicolumn{1}{c}{Rew.} & \multicolumn{1}{c}{Suc.} & \multicolumn{1}{c}{Rew.} & \multicolumn{1}{c}{Suc.} & \multicolumn{1}{c}{Rew.} \\
\midrule \midrule
\multirow{3}{*}{\rotatebox[origin=c]{0}{$0\%$ SER}} 
 & CR  & 97.48\% & 12.67 & 98.34\% & 13.30 & 98.88\% & 13.46 & \textbf{100.0\%} & \textbf{14.00} \\
 & SFR & 94.18\% & 10.99 & 87.40\% & 9.58  & 95.62\% & 10.94 & \textbf{98.2\%}  & \textbf{12.40} \\
 & LAP & 95.08\% & 11.10 & \textbf{98.10\%} & \textbf{11.76} & \textbf{98.40\%} & \textbf{11.77} & 97.0\%  & 11.70 \\
\midrule
\multirow{3}{*}{\rotatebox[origin=c]{0}{$15\%$ SER}}
 & CR  & 92.22\% & 10.29 & 94.16\% & 11.26 & 95.76\% & \textbf{11.64} & \textbf{96.7\%} & 11.00 \\
 & SFR & 90.64\% & 8.56  & 88.70\% & 8.06  & 89.22\% & 8.21  & \textbf{90.9\%} & \textbf{9.00}  \\
 & LAP & \textbf{90.34\%} & 8.59  & \textbf{92.56\%} & \textbf{9.40}  & \textbf{91.86\%} & \textbf{9.19} & 89.6\% & 8.70 \\
\midrule
\multirow{3}{*}{\rotatebox[origin=c]{0}{$30\%$ SER}}
 & CR  & 84.32\% & 7.73 & 85.36\% & 8.64 & 85.46\% & 8.65 & \textbf{89.6\%} & \textbf{9.30} \\
 & SFR & \textbf{82.48\%} & 5.11 & \textbf{80.26\%} & 5.05 & \textbf{80.34\%} & 4.63 & 79.0\% & \textbf{6.00} \\
 & LAP & \textbf{82.24\%} & \textbf{5.89} & \textbf{84.62\%} & \textbf{6.25} & \textbf{83.40\%} & \textbf{6.00}  & 76.1\% & 5.30 \\ 
\midrule
\end{tabular}
}
\resizebox{!}{0.8\height}{%
\begin{tabular}{ll||rrrrrr||rr}
 &  & \multicolumn{2}{c}{Stoc-ACER} & \multicolumn{2}{c}{Stoc-ACER-BC} & \multicolumn{2}{c}{Stoc-ACER-FB} & \multicolumn{2}{c}{HDC} \\ 
\multicolumn{2}{c}{\textit{Task}}  & \multicolumn{1}{c}{Suc.} & \multicolumn{1}{c}{Rew.} & \multicolumn{1}{c}{Suc.} & \multicolumn{1}{c}{Rew.} & \multicolumn{1}{c}{Suc.} & \multicolumn{1}{c}{Rew.} & \multicolumn{1}{c}{Suc.} & \multicolumn{1}{c}{Rew.} \\
\midrule \midrule
\multirow{3}{*}{\rotatebox[origin=c]{0}{$0\%$ SER}} 
 & CR  & 99.60\% & \textbf{14.02} & 99.64\% & \textbf{14.03} & 99.30\% & 13.87 & \textbf{100.0\%} & 14.00 \\
 & SFR & 97.66\% & 12.36 & 96.48\% & 11.98 & 96.34\% & 11.98 & \textbf{98.2\%}  & \textbf{12.40} \\
 & LAP & 95.24\% & 11.23 & 95.34\% & 11.28 & 95.40\% & 11.24 & \textbf{97.0\%}  & \textbf{11.70} \\
\midrule
\multirow{3}{*}{\rotatebox[origin=c]{0}{$15\%$ SER}}
 & CR  & \textbf{97.56\%} & \textbf{12.69} & 95.80\%  & \textbf{12.32} & \textbf{96.78\%} & \textbf{12.59} & 96.7\% & 11.00 \\
 & SFR & 88.62\% & \textbf{9.26}  & 87.64\% & 8.91  & 86.98\% & 8.67  & \textbf{90.9\%} & 9.00  \\
 & LAP & 88.74\% & \textbf{8.72}  & 88.00\% & 8.55  & 86.26\% & 8.24  & \textbf{89.6\%} & 8.70 \\
\midrule
\multirow{3}{*}{\rotatebox[origin=c]{0}{$30\%$ SER}}
 & CR  & \textbf{89.82\%} & \textbf{10.19} & 89.38\% & \textbf{9.98} & 89.32\% & \textbf{10.04} & 89.6\% & 9.30 \\
 & SFR & 72.74\% & 4.38  & 78.22\% & 5.45 & 71.02\% & 4.37  & \textbf{79.0\%} & \textbf{6.00} \\
 & LAP & 75.80\% & 4.86  & \textbf{78.66\%} & \textbf{5.40} & \textbf{77.82\%} & 5.23  & 76.1\% & 5.30 \\ 
\midrule
\end{tabular}
}
\caption{Results of Experiment $2$. Long term learning, average from $6\,000$ to $10\,000$ training dialogues, for $1\,000$ testing dialogue. Each bold result represent better models than the handcrafted agent.}
\label{tab:results_stoch}
\end{table*}

These results are encouraging and suggest that stochastic sampling makes it possible to learn a policy that performs as well as the handcrafted agent, which has been designed and fine-tuned by humans, even in hard environments. Also, these results show that demonstrations can significantly contribute to improve the performance at early learning stages.

\subsection{Hierarchical Imitation Learning}
\label{ss:hil}
This section also presents the work of the PhD candidate Thibault Cordier, who I co-supervised together with Dr.Tanguy Urvoy and Professor Fabrice Lefevre.  This work has been published in~\cite{cordier-etal-2022-graph}.

We explore \ac{GNN} for learning the policy in multi-domain and multi-task environments, in which several domains and tasks can be evoked in the same conversation. 

In practice, real applications like personal assistants or chatbots must deal with multiple tasks: the user may first want to \textbf{find} a hotel (first task), then  \textbf{book} it (second task). Moreover, the tasks may cover several domains: the user may want to find a hotel (first task, first domain), book it (second task, first domain), and then find a restaurant nearby (first task, second domain).

One way of handling this complexity is to rely on a \textit{domain hierarchy} which decomposes the decision-making process;another way is to switch easily from one domain to another by scaling up the policy. 
 
Although \textit{structured dialogue policies} can adapt quickly from a domain to another \cite{chen2020structured}, covering multiple domains remains a hard task because it increases the dimensions of the state and action spaces while the reward signal remains sparse. A common technique to circumvent this reward scarcity is to guide the learning by injecting some knowledge through a teacher policy \footnote{\label{hc} For deployment the teacher is expected to be a human expert, however, for experimentation purposes we used the handcrafted policy as a proxy~\citep{casanueva_benchmarking_2017}.}.

We study how structured policies like \textit{graph neural networks} (\ac{GNN}) combined with some degree of \textit{imitation learning} (\ac{IL}) can be effective to handle multi-domain scenarios.

We provide large scale experiments in a dedicated framework~\citep{zhu2020convlab} in which we analyse the performance of different types of policies, from multi-domain policy to generic policy, with different levels of imitation learning.

\subsubsection{Dialogue State / Action Representations}
\label{sss:dip}
One way of standardising the slot representation into a common feature space is to use  \ac{DIP} \citep{wang2015learning} parametrisation. We adopt  \ac{DIP} as state and action representations, which are not reduced to a flat vector but to a set of sub-vectors: one corresponding to the domain parametrisation (or \textit{domain representation}), 
the others to the slots parametrisation (or \textit{slot representations}). 
For any active domain, the input to the \textit{domain representation} is the concatenation of the previous \textit{domain} user and system actions (see examples of the output below, and a formal definition in Section~\ref{ss:graph_neural_network}), the number of entities fulfilling the user's constraints in the database, the booleans indicating if the dialogue is terminated and whether an offer has been found / booked. The output corresponds to action scores such as \textsc{reqmore}, \textsc{offer}, \textsc{book}, \textsc{great}, etc.
Regarding the \textit{slot representation}, its input is composed of the previous \textit{slot-dependent} user and system actions (see output below), the booleans indicating if a value is known and whether the slot is needed for the \textit{find} / \textit{book} tasks. Its output are actions scores such as \textsc{inform}, \textsc{request} and \textsc{select}.
The parameterisation used depends on the representation of the deterministic states of \textsc{ConvLab} which does not consider the uncertainty in the predictions made by the \textit{natural language understanding} (\textsc{NLU}) module.

\subsubsection{Graph Neural Network}
\label{ss:graph_neural_network}

\begin{figure}[ht!]
  \begin{center}
    \subfloat[\textsc{FNN} layer with \textsc{DIP}.]{
        \includegraphics[width=0.45\columnwidth]{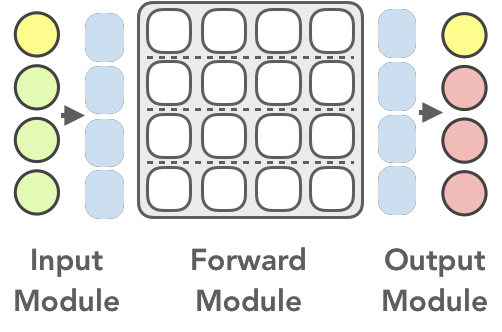}
        \label{subfig:model_stnn}
    }
    \subfloat[\ac{GNN} layer with \textsc{DIP}.]{
        \includegraphics[width=0.45\columnwidth]{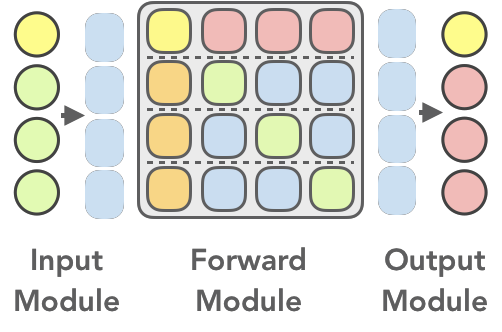}
        \label{subfig:model_gnn}
    }
  \end{center}
  \caption{Structure of the layers with \textsc{DIP}. The central box represents the weight matrix of a layer $\mathbf{W}^l$. It can be decomposed into sub-matrices $\mathbf{W}_{i,j}^l$. 
  The white sub-matrices represent any sub-weight and the coloured sub-matrices represent shared sub-weights.
  The circles represent input and output graph nodes. The \textit{domain representations} are depicted in yellow; the \textit{slot representations} in green and red. 
  }
  \label{fig:models}
\end{figure}

\begin{figure*}[!ht]
    \begin{center}
        \subfloat[\scriptsize{\textsc{FNN}.}]{
        \includegraphics[width=0.24\textwidth]{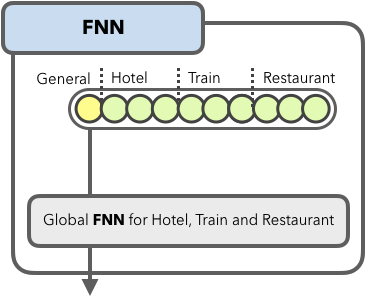}
        \label{subfig:FNN}
        }
        \subfloat[\scriptsize{\textsc{HFNN}.}]{
        \includegraphics[width=0.24\textwidth]{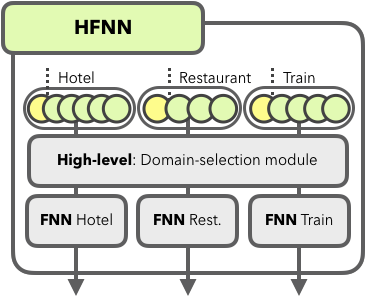}
        \label{subfig:HFNN}
        }
        \subfloat[\scriptsize{\textsc{HGNN}.}]{
        \includegraphics[width=0.24\textwidth]{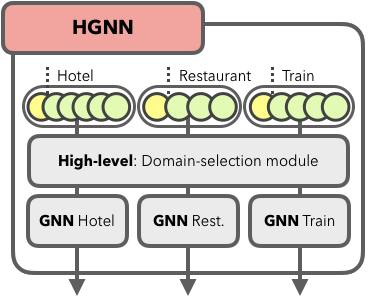}
        \label{subfig:HGNN}
        }
        \subfloat[\scriptsize{\textsc{UHGNN}.}]{
        \includegraphics[width=0.24\textwidth]{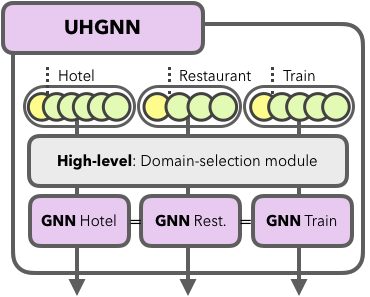}
        \label{subfig:UHGNN}
        }
    \end{center}
    \caption{Policy and input data structures. Different levels of structure are presented from classical \textit{feed-forward neural network} (\textsc{FNN}) to \textit{graph neural network} (\ac{GNN}). The prefix \textsc{H-} corresponds to a hierarchical policy and \textsc{UH-} corresponds to a unique sub-policy for all domains. For a \textsc{FNN} layer, the input data is the concatenation of all \textsc{DIP} slot representations. For a \ac{GNN} layer, the input keeps its structure.}
    \label{fig:proposals}
\end{figure*}

Prior knowledge can be integrated in our models by constraining the layer structure imposing symmetries in the neural policies. Without prior knowledge, the standard structure used is the \textit{feed-forward neural network} layer (\textsc{FNN}) as represented in Figure~\ref{subfig:model_stnn}. This unconstrained structure does not assume any symmetry in the network.

Assuming that sub-policies associated with the slots are the same, a better alternative is to use the \textit{graph neural network} layer (\ac{GNN}) presented in Figure~\ref{subfig:model_gnn}.
This structure assumes that the state and action representations have a graph structure that are identically parameterised by \textsc{DIP}.
The \ac{GNN} structure is a fully connected and directed graph, in which each \textit{node} represents a sub-policy associated with a slot and a directed \textit{edge} between two sub-policies represents a message passing. 
We identify two roles for sub-policies: the general node as \textsc{I-node} associated to the \textit{domain representation} and the slot nodes denoted as \textsc{S-node} associated to the \textit{slot representations}. Both representations were introduced in Section~\ref{sss:dip}. We also identify the relations: \textsc{I2S} for \textsc{I-node} to \textsc{S-node}, \textsc{S2I} and \textsc{S2S} respectively. 

We formally define the GNN structure as follows. Let $n$ be the number of slots and $L$ the number of layers. Let be $x$ the dialogue state, $\mathbf{x}_0=\phi_0(x)$, $\mathbf{h}^l_0\, \forall l\in[0,L-1]$ and $\mathbf{y}_0$ be respectively the input, hidden and output \textsc{I-node} representations. 
Let the input, hidden and output \textsc{S-nodes} representations be respectively $\forall i \in[1,n], \, \mathbf{x}_i=\phi_i(x)$, $\mathbf{h}^l_i\ \forall l\in[0,L-1]$ and $\mathbf{y}_i$.
First, the \ac{GNN} transforms inputs:
\begin{equation}
\begin{split}
    \forall i \in [0, n],\quad \mathbf{h}^0_i = F_i^0(\phi_i(\mathbf{x})) \\
    \mathrm{with}\quad F_i^0(\mathbf{h}) = \sigma^0(\mathbf{W}_i^0\mathbf{h} + \mathbf{b}_i^0)
\end{split}
\end{equation}
Then, at the $l$-th layer, it computes the hidden nodes representations (Eq.~\ref{eq:internal}) by following message sending\footnote{The notation $i \leftarrow j$ denotes a message sending from slot $j$ to slot $i$. It also corresponds to the directed relation between the slots $j$ and $i$. The notation $i \leftarrow *$ denotes all messages sending to slot $i$.} (Eq.~\ref{eq:send}), message aggregation (Eq.~\ref{eq:agg}) and representation update (Eq.~\ref{eq:update}):
\begin{subequations}
\begin{align}
    & \forall i \in [0, n],\quad \mathbf{h}^{l}_{i} = F_i^l(\mathbf{h}^{l-1}) \label{eq:internal} \\
    \mathbf{m}^{l}_{i \leftarrow j} & = M_{i \leftarrow j}^l(\mathbf{h}^{l-1}_{j}) 
    \label{eq:send} \\
    \mathbf{m}^{l}_{i} & = A_i^l(\mathbf{m}^{l}_{i \leftarrow *}) \label{eq:agg} \\
    \mathbf{h}^{l}_{i} & = U_i^{l}(\mathbf{m}^{l}_{i}) =  \sigma^{l}(\mathbf{m}^{l}_{i}  \label{eq:update})
\end{align}
\end{subequations}
The message sending function $M_{i \leftarrow j}^l$ is a linear transformation with bias. 
The message aggregation function $A_i^l$ is the average pooling function.
The representation update function $U_i^{l}$ compute the new hidden representation with \textsc{ReLU} activation function and dropout technique during learning stage.
Finally, the \ac{GNN} concatenates ($\oplus$ symbol) all final nodes representations and computes the policy function with the Softmax activation function.
\begin{gather}
    \mathbf{y} = \sigma^L(\bigoplus_{i=0}^n \mathbf{W}_{i}^L \mathbf{h}_i^{L-1} + \mathbf{b}_i^{L})
\end{gather}

\subsubsection{Imitation Learning}
In addition to the structured architecture, we use some level of \ac{IL} to guide the agent's exploration.
In our experiments, we used \textsc{ConvLab}'s {handcrafted policy} as a \textit{teacher} \footnoteref{hc}, but other policies could be used as well.
\textit{Behaviour cloning} (\textsc{BC}) is a {pure supervised learning} method that tries to mimic the teacher policy. Its loss function is the cross-entropy loss as in a classification problem.
\textit{Imitation Learning From Oracle Demonstrations} (\textsc{ILfOD}) is a \textsc{RL} method which allows the agent to play oracle actions as demonstrations and to inject them in its \textit{replay buffer}. The same presented in Section~\ref{sss:il}. 
In our experiments, we kept half of the agent's own actions in the buffer along with those generated by the oracle.
\textit{Imitation Learning From Oracle Supervision} (\textsc{ILfOS}) is the combination of supervised and reinforcement learning
when the agent learns with a supervised loss, namely the margin loss \cite{hester2018deep}.

\subsubsection{Experiments on GNN and Imitation Learning}
We performed an ablation study: (i) by progressively extending the baseline to our proposed \textsc{GNN}s and (ii) by guiding the exploration with \textsc{IL}.
All the experiments were restarted 10 times with random initialisations and the results evaluated on 500 dialogues were averaged. Each learning trajectory was kept up to 10,000 dialogues with a step of 1,000 dialogues in order to analyse the variability and stability of the methods.

\paragraph{Models}
The baseline is \textsc{\textbf{ACER}} which is a sophisticated actor-critic method~\citep{wang2016sample}. After an ablation study, we progressively added some notion of hierarchy to \textsc{FNN}s to approximate the structure of \textsc{GNN}s. 
\textsc{\textbf{FNN}} is a feed-forward neural network  with \textsc{DIP} parametrisation. Thus, the agent actions are single-actions.
\textsc{\textbf{FNN-REF}} is a FNN with the native parametrisation (no \textsc{DIP}) with multiple-actions of \textsc{ConvLab}\footnote{The native parametrisation manually groups multi-actions based on \textsc{MultiWOZ}~\cite{budzianowski2018multiwoz}.}.
\textsc{\textbf{HFNN}} is a hierarchical policy with domain-selection module and based on \textsc{FNN}s for each domain. 
\textsc{\textbf{HGNN}} is a hierarchical policy with domain-selection module and based on \textsc{GNN}s. 
\textsc{\textbf{UHGNN}} is a \textsc{HGNN} with a unique \textsc{GNN} for all domains. 

\paragraph{Metrics:}
We evaluate the performance of the policies for all tasks. For the {find} task, we use the precision, the recall and the F-score metrics: the \textbf{inform rates}. For the {book} task, we use the accuracy metric namely the \textbf{book rate}.
The dialogue is marked as \textbf{successful} if and only if both inform's recall and book rate are 1.
The dialogue is considered \textbf{completed} if it is successful from the user's point of view (\textit{i.e} a dialogue can be completed without being successful if the information provided is not the one objectively expected by the simulator).

\paragraph{Evaluation of the Dialogue Manager}
We performed an ablation study based on \textsc{ACER} as reported in Figure~\ref{fig:convlab_approach}.
First, all \textsc{RL} variants of \textsc{ACER} (Figure~\ref{subfig:convlab_a2c}) have difficulties to learn without supervision in contrast to \textsc{BC} variants (Figure~\ref{subfig:convlab_bc}).
In particular, we see that hierarchical decision making networks (\textsc{HFNN} in green), graph neural network (\textsc{HGNN} in red) and generic policy (\textsc{UHGNN} in purple) drastically improve the performance compared to \textsc{FNN}s. Similarly, using \textsc{IL} like \textsc{ILfOD} (Figure~\ref{subfig:convlab_a2cilfod}) and \textsc{ILfOS} (Figure~\ref{subfig:convlab_a2cilfos}) notably improves the performance. 
Therefore, learning generic \textsc{GNN}s allows collaborative gradient update and efficient learning on multi-domain dialogues.
\begin{figure*}[ht!]
    \begin{center}
        \subfloat[\scriptsize{Pure \textit{ACER}}]{
        \includegraphics[width=0.43\textwidth]{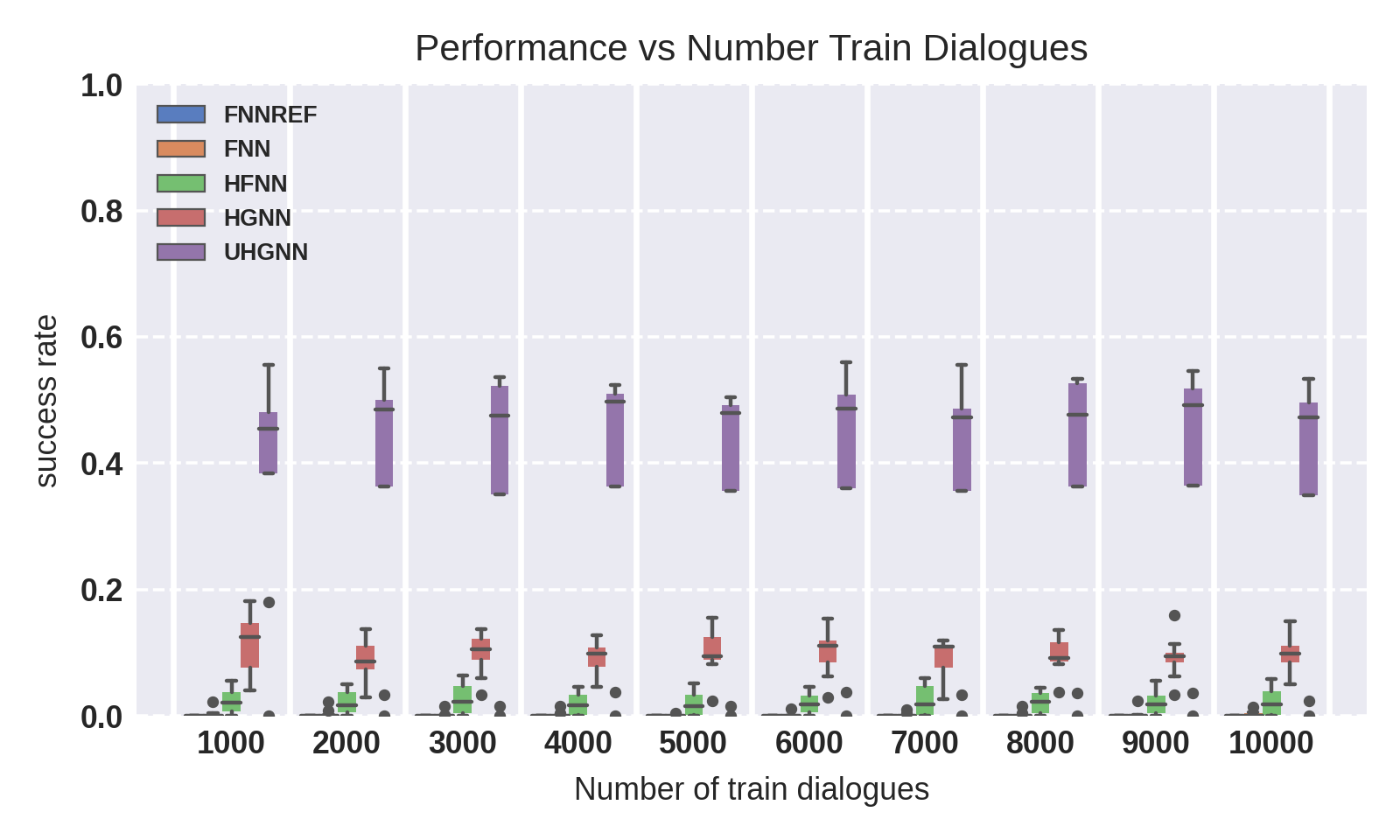}
        \label{subfig:convlab_a2c}
        }
        \subfloat[\scriptsize{Pure \textit{BC}}]{
        \includegraphics[width=0.43\textwidth]{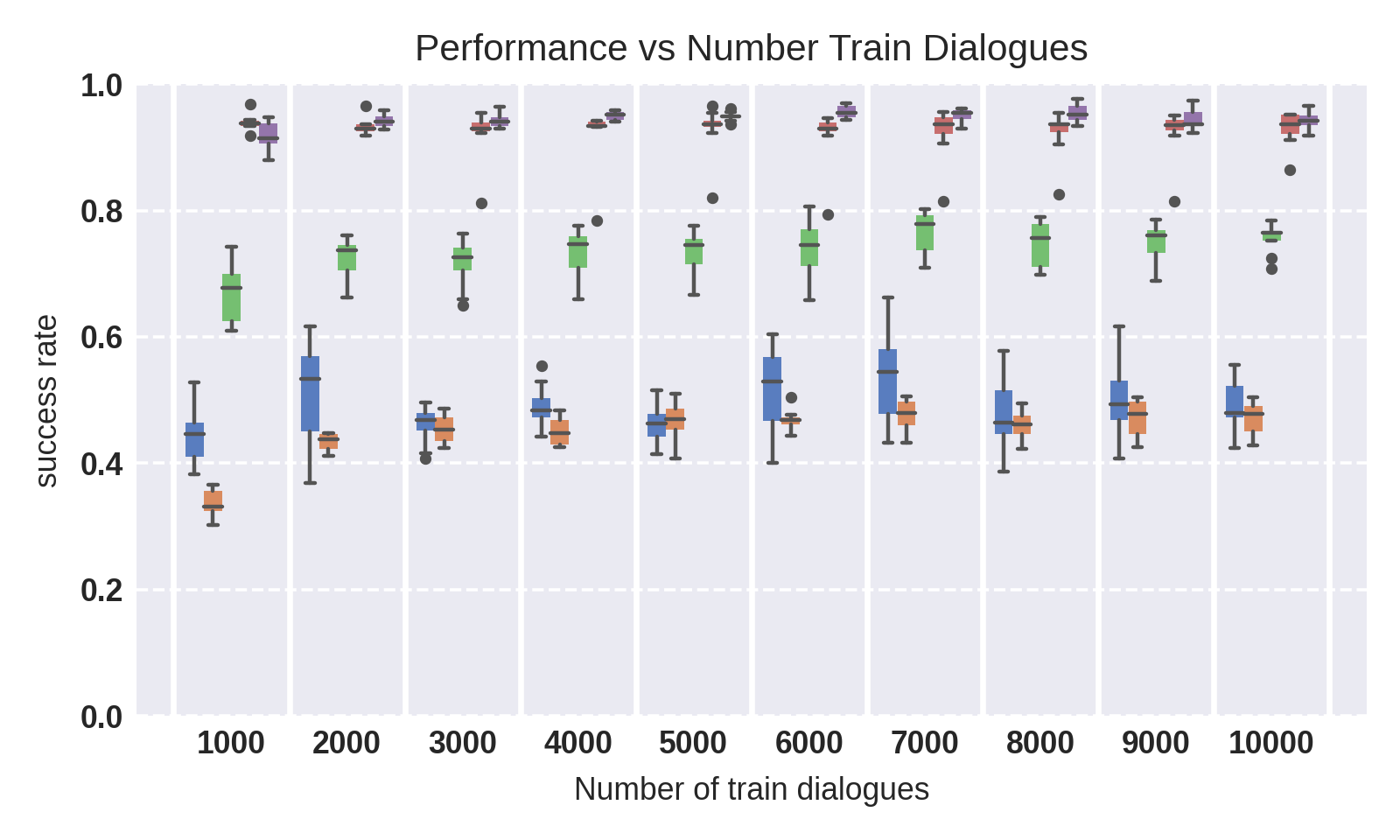}
        \label{subfig:convlab_bc}
        }\\
        \subfloat[\scriptsize{\textit{ACER} with \textit{ILfOD}.}]{
        \includegraphics[width=0.43\textwidth]{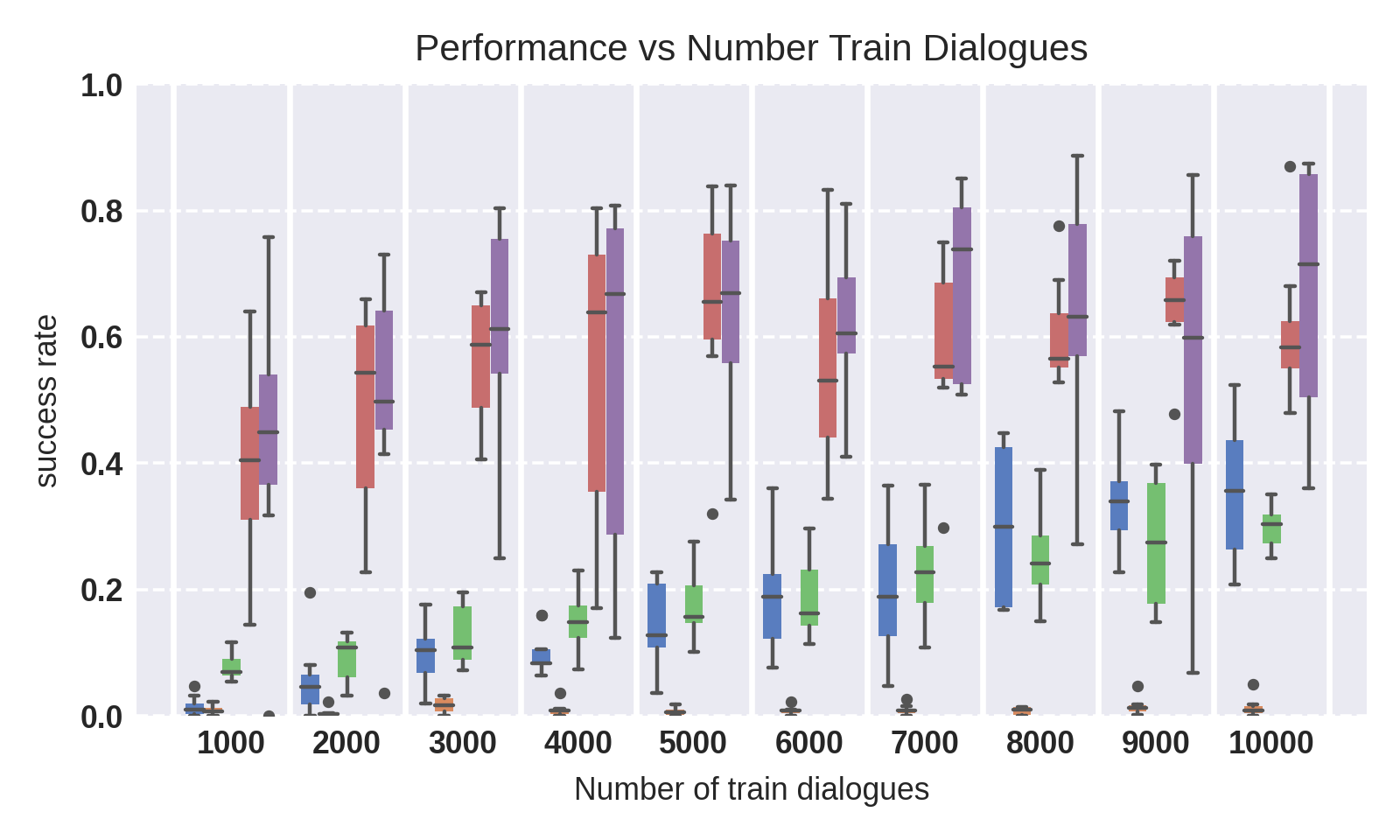}
        \label{subfig:convlab_a2cilfod}
        }
        \subfloat[\scriptsize{\textit{ACER} with \textit{ILfOS}}.]{
        \includegraphics[width=0.43\textwidth]{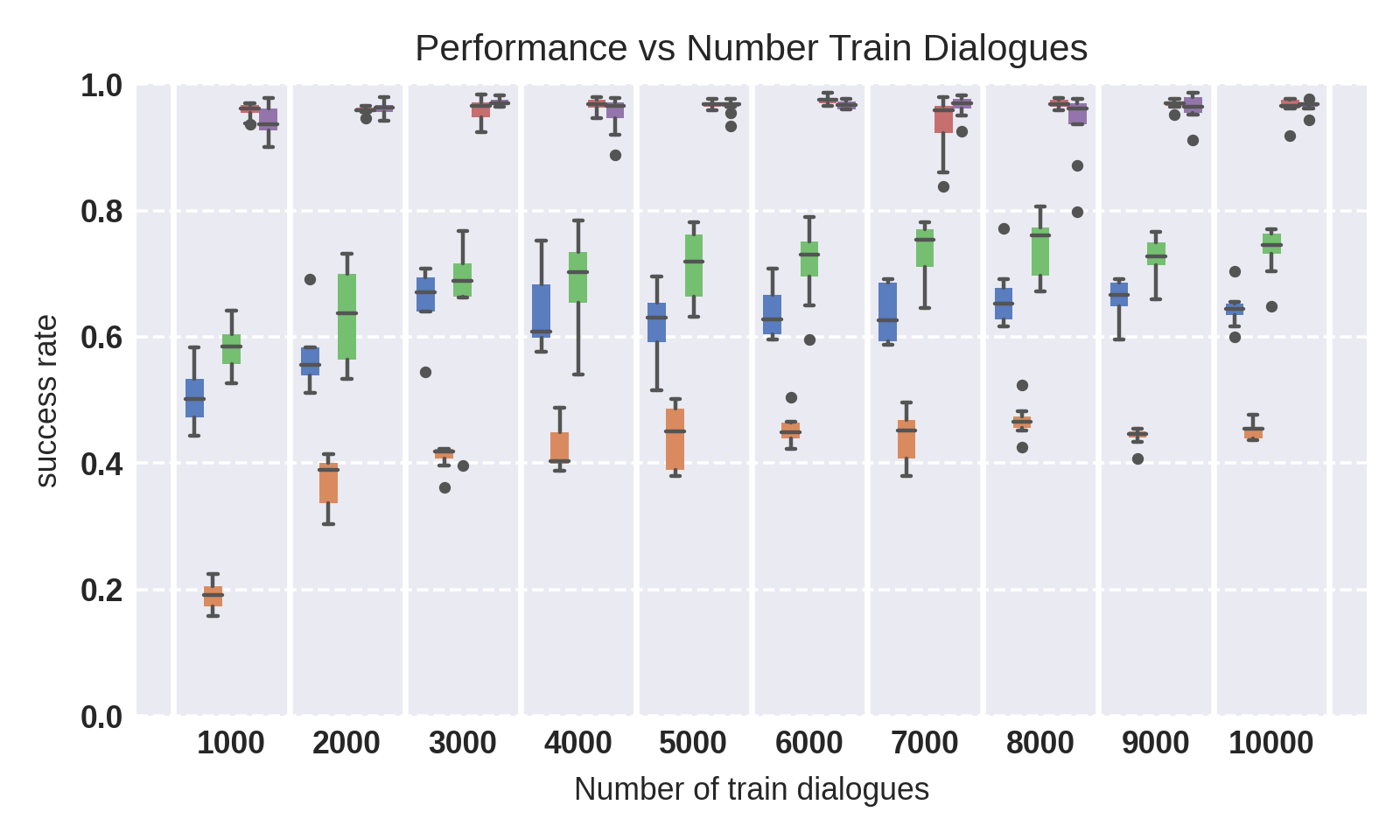}
        \label{subfig:convlab_a2cilfos}
        }\\
    \end{center}
    \caption{Distribution via boxplot of the performance of the proposed approaches on \textsc{ConvLab}, with 10 different initializations and without pre-training. The coloured area represents the interquartile Q1-Q3 of the distribution, the middle line represents its median (Q2) and the points are outliers. 
    }
    \label{fig:convlab_approach}
\end{figure*}

Conversely, we observe that hierarchical decision making 
with \textsc{HFNN}s does not systematically guarantee any improvement. 
These results suggest that \textsc{GNNs} are useful for learning dialogue policies on multi-domain 
which can be transferred during learning across domains on-the-fly to improve performance.
Finally, regarding \textsc{ILfOD} variants (Figure~\ref{subfig:convlab_a2cilfod}), we can observe that all architectures are affected by a large variability. This shows that multi-domain dialogue management is difficult despite the use of demonstrations and that learning with reward is not sufficient to robustly succeed.

\paragraph{Evaluation of the Dialogue System} We evaluate the policy learning algorithms in the entire dialogue pipeline, in particular our best \textsc{DM} policy \textsc{ACER-ILfOS-UHGNN} under a shorter name \textsc{\textbf{ACGOS}}. 

The results of our experimentation are presented in the paper~\citep{cordier-etal-2022-graph}.
We observe that the performance of our approach is closed to the handcrafted policy (the teacher) when directly passing the dialogue acts, when using \textsc{BERT NLU} \cite{devlin2018bert} and template-based \textsc{NLG}. Moreover, the performance of our approach is better than baselines with a significant difference.
These results highlight the benefit of structured policies against standard policies.

A limitation of current policies in~\textsc{ConvLab}, including ours, is that the robustness to noisy inputs is not specifically addressed as it had been done in PyDial~\cite{ultes2017pydial}.
It could be also interesting to study the impact of incorporating real human feed-backs and demonstrations instead of a handcrafted teacher. 

The \textsc{GNN} structured policies combined with imitation learning avoid sparsity, while being data efficient, stable and adaptable. They are relevant for covering multi-domain task dialogue problems.

A continuation of this work for few-shot learning will soon be published in the findings EACL2023, for more details please refer to~\citep{cordier2023few}.

\chapter{Contributions to Conversational \ac{QA} and Other Contributions}
\label{c:contributionsConvQA}

\begin{flushright}

\parbox{.8\textwidth}{`Estaba perdiendo la vista y el oído, parecía confundir a los interlocutores con personas que conoció en épocas remotas de la humanidad, y contestaba a las preguntas con un intrincado batiburrillo de idiomas.'
--- \emph{Gabriel García Marquez, Cien años de soledad.}}
\end{flushright}
\begin{flushright}
\parbox{.8\textwidth}{`He was losing his sight and his hearing, he seemed to confuse the people he was speaking to with others he had known in remote epochs of mankind, and he would answer questions with a complex hodgepodge of languages.'
--- \emph{Gabriel García Marquez, One Hundred Years of Solitude.}}  
\end{flushright}

Conversational \ac{QA} is a relatively recent area of research that groups reading comprehension, \ac{QA} and dialogue. Typically, it consists in a sequence of questions and answers related to a paragraph or to a knowledge graph. My contribution in this field was to enrich existing datasets with (i) information about the ellipsis and coreferences (Section~\ref{s:ellipsiscoref}) and  (ii) question rewriting (Section~\ref{s:qr}).  I also contribute to the creation of a new dataset \ac{KGConv} (Section~\ref{s:KGConv}), that we will soon made public.
This work was part of the industrial research project \ac{DIANA}, of which I was the head. The work described in here was made mainly by the young researcher Quentin Brabant under my supervision.  For the work presented in (Section~\ref{s:qr}), I proposed the idea of enriching the corpus CoQA with question rewriting in 2020. I wrote with Timothy Garwood a document describing the annotations and we manually annotated 10 dialogues. Then, I was in charge of the administrative process to  formalise the collaboration with ELRA to produce the annotations, this took about a year. I  also checked the annotations and conducted experiments on conversational question answering by using RoBERTa. Regarding this work, Gwenole Lecorve worked on question generation, while Quentin Brabant on question rewriting. The work presented in Sections~\ref{s:ellipsiscoref} and~\ref{s:KGConv} were joint work with Claire Gardent as part of the European \ac{ITN} project \ac{NL4XAI}, that involves industrial partners as Orange and academic institutions as the CNRS. I was representing  Orange in this project.

\section{Detection of Ellipsis and Co-reference in conversational corpora }
\label{s:ellipsiscoref}

    This work is joint work with Quentin Brabant and Claire Gardent. It was published in~\citep{brabant:hal-03533906}.
    We made several contributions to the task of ellipsis and coreference detection in conversational corpora. We created labelled data by enriching three existing datasets with annotations indicating whether a turn contains an ellipsis and/or a coreference. As these annotations were incomplete, we drew on inferential relations between incompleteness, pronominalisation, ellipsis and coreference to both extend and complement these annotations.
    We then use these annotated data to train a classifier based on DistilBERT~\citep{sanh_distilbert_2020},
    which assigns to each question in a conversation two labels indicating whether it contains an ellipsis and/or a coreference.
    We also explore how active learning, multilabel approaches and fine-tuning can be used to train this model. 
    
    A \emph{coreference} occurs when an entity is referred via two or more expressions in the same conversation. However, we are only interested in detecting a particular kind of coreference. We say that a coreference happens in a turn if and only if (1) it contains an expression referring to an entity already mentioned in a previous turn and (2) this entity cannot be identified outside of the conversational context. The \emph{resolution} of a coreference consists in replacing the referring expression by an unambiguous reference to the entity.
    
    In linguistics, an \emph{ellipsis} is the omission of one or several words from a clause that preserves the meaning in the context. When a turn is not understandable without its context (i.e. without the conversation history), we call it \emph{incomplete}.
    In this paper, we assume that any conversation turn contains an ellipsis if and only if it is still incomplete after coreferences have been resolved.
    It follows from this definition that an incomplete sentence contains either a coreference, an ellipsis, or both. 

    A conversation is a sequence of alternating questions and answers that starts with a question and ends with an answer: $(q_1, a_1, q_2, a_2 \dots , q_n, a_n)$.
    In many available conversational question answering datasets questions are sentences produced by humans (e.g. \citep{choi_quac_2018,christmann_look_2019,elgohary_can_2019,quan_gecor_2019,reddy_coqa_2019}), while answers are often given by an automated system, and often not in the form of a sentence.
    For this reason, we focus on ellipsis and coreference detection in questions.
    Moreover, we will sometimes use the term question to refer to turns that are not question per-say, but that are produced by a user and not by an automated system (see Section \ref{s:original}, GECOR dataset).
    
    We propose a model to predict whether any given question $q_i$ of a conversation $(q_1, a_1,  \dots, q_n, a_n)$ contains an ellipsis and/or a coreference;
    since any turn can normally be understood based on the context of previous turns, our task can be seen as the classification of $q_i$ with the given context $c = (q_1, a_1, \dots , q_{i-1}, a_{i-1})$.
    We thus formulate our task as a $2$-labels classification:  for a given input question $q_i$ and an input context $c$, output two values $(\coref, \ellipsis) \in \{0,1\}^2$ where $1$ denotes the presence of the phenomenon and $0$ denotes its absence. 
    We call \emph{instance} of our task the couple formed by a question, and its context.
    An instance is \emph{annotated} when it is associated with an annotation of the form $(\coref, \ellipsis)$.

    \subsection{Active learning (AL)}
        We use the following values for annotating the datasets: 1 for the presence of a phenomenon (positive class), 0 for its absence (negative class). Cases where no label is assigned are denoted by the value -1. Note that -1 does not denote a class, but only the absence of information about the actual class.
        We describe how we process each dataset in order to obtain train instances for our task.
        
        \noindent
        {\bf ConvQuestions~\cite{christmann_look_2019}.}
        Many conversations of ConvQuestions~ are centered on the same entity; those conversations tend to be similar to each other, as they often have questions in common.
        In order to maximise the benefits of manual annotations,
        we created subsets of the original data containing exactly one conversation per topic entity.
        This resulted in train/dev/test sets containing respectively 905/330/335 questions in total.
        Based on these new sets,
        we created an instance of our task for each question (except the first one) of each conversation.
        Some of these conversations where manually annotated with $(\coref, \ellipsis)$ values.
        We obtained train/dev/test of 247/329/331 annotated instances.

        \noindent
        {\bf GECOR~\cite{quan_gecor_2019}.}
        We create instances as follows.
        For each conversation $(q_1, a_1, \dots, q_n, a_n)$ in the GECOR dataset, each $i \in \{2, \dots, n\}$, and each variant $q'_i \in \{q_i(e), q_i(r), q_i(c)\}$ of the question $q_i$:
        if $q'_i$ is not empty, then we create the instance $((q_1, a_1, \dots, a_{i-1}), q'_i)$ and annotate it with $(\coref, \ellipsis)$ values.
        Those values can sometimes be deduced by using the following rules:
        \begin{itemize}
            \item $q_i(e)$ contains an ellipsis;
            \item $q_i(r)$ contains a coreference;
            \item $q_i(c)$ contains no ellipsis nor coreference;
            \item if $q_i(e) = q_i(r)$ we infer that both $q_i(e)$ and $q_i(r)$ contain an ellipsis and a coreference;
            \item if $q_i(e)$ is empty, we infer that $q_i$ contains no ellipsis and thus $q_i(r)$ neither;
            \item if $q_i(r)$ is empty, we infer that $q_i$ contains no coreference and thus $q_i(e)$ neither.
        \end{itemize}
        These rules are not sufficient to deduce ellipsis and coreference label values in all cases.
        By default, the value -1 is assigned.

        \noindent
        {\bf CANARD~\cite{elgohary_can_2019}.}
        Instances were extracted similarly as from the GECOR dataset. The two main differences are: for each created conversation, two variants (original and complete) of the last question are used.
        When the complete variant is used, we assign $0$ to both $\coref$ and $\ellipsis$; otherwise, we assign $-1$.
        An example is given in Table \ref{tab:canard-example}.

    \begin{table}
        \setlength{\tabcolsep}{4pt}
        \centering
        Piece of conversation from CANARD:\\
        \begin{tabularx}{\textwidth}{rX}
            \toprule
            $q_1$ & What is On the Sunday of Life? \\
            $q_1(c)$ & What is On the Sunday of Life? \\ \midrule
            $a_1$ & In 1992, Delerium released On the Sunday of Life as an edition of 1,000 copies, complete with a deluxe gatefold sleeve. \\ \midrule
            $q_2$ & Did it do well? \\
            $q_2(c)$ & Did Porcupine Tree, On the Sunday of Life do well? \\ \midrule
            $a_2$ & On the Sunday of Life... had accumulated sales of more than 20,000 copies. \\ \midrule
            $q_3$ & Was it rereleaesd? \\
            $q_3(c)$ & Was Porcupine Tree, On the Sunday of Life rereleaesd? \\
            \bottomrule
        \end{tabularx}
        ~\\
        ~\\
        Corresponding instances of the task:\\
        \setlength{\tabcolsep}{6pt}
        \begin{tabular}{rc||cc||cccc}
            \toprule
            Context & Question & Coref & Ellipsis & \cellcolor{gray!50}Coref & \cellcolor{gray!50}Ellipsis & \cellcolor{gray!50}Incomp. & \cellcolor{gray!50}Pronoun\\ \midrule
            $(q_1, a_1)$ & $q_2$ & -1 & -1 & 1 & -1 & 1 & 1 \\
            $(q_1, a_1)$ & $q_2(c)$ & 0 & 0  & 0 & 0 & 0 & 0 \\
            $(q_1, a_1, q_2, a_2)$ & $q_3$ & -1 & -1  & 1 & -1 & 1 & 1 \\
            $(q_1, a_1, q_2, a_2)$ & $q_3(c)$ & 0 & 0 & 0 & 0 & 0 & 0 \\
            \bottomrule
        \end{tabular}
        \caption{Example of conversation from CANARD and the corresponding instances of the task. Columns with gray headers show the result of label filling.}
        \label{tab:canard-example}
    \end{table}
    
         At this point many labels are missing in the instances of the task.
        In particular, instances from CANARD do not contain any positive label.
        We addressed this issue via two approaches: multilabel learning and label filling.

        Multilabel classification can be seen as a particular case of multitask learning, since a single model is trained on several binary classification tasks. One justification for using this approach (instead of one model per classification) is that the parameters are shared during training which has been shown in the literature to beneficial to all the classifiers.

        The $4$-labels classification task considers the following labels: coreference, ellipsis, incompleteness, and pronoun detection.
        Formally, it means that annotations of the form $(\coref, \ellipsis)$ are replaced by annotations of the form $(\coref, \ellipsis, \inc, \pronoun)$.
        We used automatic pronoun detection to provide a $0$ or $1$ value to $\pronoun$ in all questions.
        By default, the value of $\inc$ is set to -1, except for instances from CANARD where the value is known.
        
        We then replace some of the $-1$ values by taking advantage of the logical dependencies between labels: a pronoun always indicates a coreference; incompleteness is either due to a coreference or an ellipsis; coreferences and ellipses always cause incompleteness. We therefore applied the following rules to each instance, in order:
        \begin{enumerate}
            \item if $\pronoun = 1$ then $\coref \gets 1$,
            \item if $\coref = 1$ or $\ellipsis = 1$ then $\inc \gets 1$,
            \item if $\coref = 0$ and $\ellipsis = 0$ then $\inc \gets 0$,
            \item if $\inc = 0$ then $\coref  \gets 0$ and $\ellipsis  \gets 0$.
        \end{enumerate}
        Remark that in some cases these rules are not sufficient to get rid of all unknown values.
        Such cases can be found in the examples of Table~\ref{tab:canard-example}.        

        Active learning (AL) is a human-in-the-loop method that aims at maximizing the performance gains relatively to the number of manual annotations. It is especially interesting when few labeled data are available and only a small fraction of unlabeled data can be manually annotated in reasonable time. We apply several rounds of AL for labeling (separately) ellipses and coreferences. Each round consists in the following steps:
        \begin{enumerate}
            \item {\it Train and evaluate a model.} We use CANARD/GECOR as a training set. All CANARD instances that have already been manually annotated during previous rounds are included.
            The evaluation is done on ConvQuestions test set.
            \item {\it Run the model on unlabeled data.} The model trained in step 1 associates a prediction $(\coref^*, \ellipsis^*)$ to each instance.
            \item {\it Select a subset of unlabeled data.}
            We select the 50 CANARD conversations on which the model displays the least certainty. Since one conversation is the source of several instances, 
            we define the certainty of a conversation as the average certainty of the corresponding instances.
            The certainty of the model (for a given label, on a given instance) is defined as the distance from $0.5$ of the output corresponding to the predicted label value, i.e.: $|\coref^* - 0.5|$ for coreference and $|\ellipsis^* - 0.5|$ for ellipsis.
            \item {\it Manually label the selected subset.} We label the selected conversations (either for ellipsis or coreference).
            Labeled conversations are used during training in the next loop.
        \end{enumerate}
        We stop repeating these steps when the evaluation score stops increasing.

\subsection{Experiments}\label{s:experiments}

    
    
        We evaluate the following model variants.
        \begin{itemize}
            \item {\it Baseline}.
            The baseline is a DistilBert\citep{sanh2019distilbert}  model trained on the $4$-label classification task on CANARD/ GECOR.
            \item {\it Fine tuning only}.
            The model is fine-tuned on the $4$-label classification task on the training set of ConvQuestions.
            \item {\it Baseline + AL.}
            The model is fine-tuned on the $4$-label classification task on CANARD/GECOR, but labelled instances of CANARD are added via AL. Each round of AL adds 50 instances that are labelled for either coreference or ellipsis. We evaluate several versions of this variant: three versions use instances that were annotated for coreference via, respectively, 1, 2, and 3 rounds of AL. Three others versions use instances that were annotated for ellipsis via 1, 2, and 3 rounds.
            \item {\it Baseline + all AL.}
            Identical to baseline + AL, but using all annotations produced for coreference and ellipsis (3 rounds for each).
            \item {\it Baseline + all AL + fine tuning}.
            Identical to {\it Baseline + all AL.}, but training on CANARD/GECOR is followed by a fine-tuning step on the training set of ConvQuestions.
            \item {\it 2-label variants}.
            We evaluate three of them. They are respectively identical to { \it baseline}, to {\it baseline + all AL}, and to {\it baseline + all AL + fine tuning}, with the difference that the model is trained on the 2-labels classification task.
        \end{itemize}
        
    
    We use GECOR and CANARD for training our models, while ConvQuestions is used for evaluation and fine tuning. In this way we can better assess how well the classifier behaves on unseen data, data that is different from the data the model was trained on.
    During training, labels with -1 value are simply ignored (no error is retro-propagated).
    During evaluation, we measure the recall, precision, and F-measure on ellipsis and coreference detection.    
        
    \subsection{Results}
    
        The results are displayed in Table \ref{tab:resultsel}. Each line corresponds to a variant of the model.
        
        \begin{table}
            \centering
            \setlength{\tabcolsep}{9pt}
            \begin{tabular}{rlcccccc}
            	\toprule
            	& & \multicolumn{3}{c}{Coreference} & \multicolumn{3}{c}{Ellipsis} \\
            	 & & P & R & F1 & P & R & F1 \\ \midrule
            	1 &  fine tuning only & 81 & 65 & 72 & 51 & 67 & 57 \\ \midrule
            	2 & baseline & \bf 97 & 64 & 77 & 64 & 36 & 46 \\
            	3 & ~~~~~ + AL for ellipsis (1 round) & 92 & 63 & 75 & 71 & 48 & 56 \\
            	4 & ~~~~~ + AL for ellipsis (2 rounds) & 89 & 72 & 80 & 83 & 41 & 55 \\
            	5 & ~~~~~ + AL for ellipsis (3 rounds) & 85 & 79 & 82 & 74 & 48 & 57 \\
            	6 & ~~~~~ + AL for coref. (1 round) & 87 & 84 & 85 & 72 & 46 & 56 \\
            	7 & ~~~~~ + AL for coref. (2 rounds) & 92 & 81 & 86 & 71 & 39 & 50 \\
            	8 & ~~~~~ + AL for coref. (3 rounds) & 95 & 79 & 86 & 67 & 31 & 43 \\
            	9 & ~~~~~ + all AL labels & 94 & 81 & 87 & 84 & 46 & 59 \\
            	10 & ~~~~~ + fine tuning & 94 & \bf 93 & \bf 94 & 83 & \bf 71 & \bf 77 \\ \midrule
            	11 & baseline, 2-labels variant & 89 & 68 & 77 & \bf 100 & 10 & 19 \\
            	12 & ~~~~~ + all AL labels & 91 & 86 & 89 & 88 & 35 & 50 \\
            	13 & ~~~~~ + all AL labels + fine-tuning & 94 & \bf 93 & 93 & 84 & 70 & 76 \\
            	\bottomrule
            \end{tabular}
            \caption{Results of the experiments. Scores are given as percentages.}
            \label{tab:resultsel}
        \end{table}
        
        Generally, the results show that coreference detection performs better than ellipsis detection.
        Moreover, by looking at lines 2 to 9 in the table, we see that AL is clearly beneficial; the {\it all AL labels} variant improves F1 scores for coreference and ellipsis detection by $10$ and $13$ points compared to the baseline.
        The same conclusion is drawn when comparing lines 11 and 12.
        The effects of training on 4 labels versus 2 are less clear:
        by comparing lines 2, 9, 10 to lines 11, 12, 13, we see that 4-labels variants perform roughly as well as their 2-labels counterparts on coreference detection.
        For ellipsis detection, they score significantly higher on F1 score when no fine tuning is applied, but the scores are too low to propose a meaningful interpretation.
        Fine tuning increases scores for both ellipsis and coreference detection; however the increase is way larger in the case of ellipsis. In fact, coreference detection arguably performs reasonably well without fine-tuning, contrary to ellipsis detection. A possible explanation is that the kinds of ellipses occurring in one dataset can be different from those occurring in another. In contrast, coreferences cover a narrower set of phenomena.

        In addition to measuring performances, we looked at the output of the model on the test set: we noticed that coreferences due to pronouns use are well recognized, while many false negatives correspond to cases where an entity is referred to via its type or function, as in: ``To which continent does Germany belong? What size is the country?''.

\section{Question Rewriting}
\label{s:qr}

    As mentioned before, \ac{CQA} \citep{reddy_coqa_2019,choi_quac_2018,saha2018complex} is a task in which a system  interacts with a user. The interaction takes the form of a conversation, where the user always asks questions that the system answers.
    In this work, we focus on the case where the system searches for answers in a passage,
    although settings relying on structured data (e.g. knowledge bases) also exist~\citep{saha2018complex}, as the one presented in the following section (Section~\ref{s:KGConv}).
    Compared to \ac{QA}, the system faces an additional difficulty: each question is asked in a \textit{conversational context} that consists in previous turns. Therefore, implicit references to the context may happen in the form of ellipses and coreferences, making the understanding of questions more difficult for the system.
    One way to overcome this difficulty is Question Rewriting (QR), which consists in rewriting each original (\textit{in-context}) question into an \textit{out-of-context} question that is understandable by itself, i.e., that can be answered without knowing the conversational context.

    We present the corpus \ac{CoQAR}, which is an annotated subset of the CQA corpus CoQA~\citep{reddy_coqa_2019}. \ac{CoQAR} was obtained by asking specialised native speakers to annotate
    original questions with at least two and at most three distinct \textit{out-of-context} rewritings. This work was published in~\citep{brabant-etal-2022-coqar}. Our contribution is two-fold.

    Firstly, we provide \ac{CoQAR}, which contains high-quality questions rewritings. The corpus is publicly available\footnote{\label{coqar}The COQAR dataset is publicly available at \url{https://github.com/Orange-OpenSource/COQAR}}; moreover, its annotations were conducted in accordance to ethical concerns: every annotator involved was properly hired.
    
   Secondly, we assess the quality of the annotations of \ac{CoQAR} through several experiments. We train Question Rewriting (QR) models. We then rate these models' outputs via human evaluation. We also evaluate these models as preprocessing steps of (conversational and non-conversational) QA models. To this end, we compare the performance of a stat-of-the-art QA model with and without QR.
    
   Our results support the claim of \citep{vakulenko2021question} that QR models can be successfully used in combination with existing QA models. Indeed, we found that adding QR as a preprocessing step boosts the performances of QA models and allows reusing non-conversational state-of-the-art QA systems while reducing performance degradation on CQA.

\subsection{Annotations}
\label{sec:CoQAR}
        We decided to hire two specialised native-speakers' annotators.
        Their task was to annotate original (\textit{in-context}) questions from CoQA with at least two and at most three distinct \textit{out-of-context} rewritings.
        To make sure that they understand what was expected, we ourselves annotated a conversation and provided it as an example.
        An example of conversation annotated by the annotators is provided in Table \ref{tab:example}.

        While annotators were told to preserve the meaning of the original sentences, they were also asked to paraphrase in their rewritings. As a results, these annotations contrast with those of CANARD, where the structure of the original question is usually preserved in the rewriting.
        In total, $4.1k$ conversations of CoQA train set were annotated as well as all $500$ conversations of the dev set. Since the test set of CoQA is not available, no conversation was annotated from it. The train and dev sets of \ac{CoQAR} respectively contain 45k and 8k questions. Table \ref{tab:nb-rewritings} summarises the number of questions that have 0,1,2 or 3 rewritings.

        \begin{table}
            \centering
            \begin{tabular}{cccccc}
                \hline
                & \multicolumn{4}{c}{Number of rewritings} &\\
            	& $0$ & $1$ & $2$ & $3$ & total \\
            	\hline
            	train & 365 & 108 & 31,378 & 13,210 & 45,061 \\
            	dev & 9 & 0 & 37 & 7,937 & 7,983 \\
            	\hline
            \end{tabular}
            \caption{Number of questions depending on the number of rewritings.}
            \label{tab:nb-rewritings}
        \end{table}
        
        Overall, passages contain from 75 to 1079 words, with an average of 275. Conversation length distribution is displayed in Figure \ref{fig:dialogue-length}.

        On average, out-of-context rewritings are longer (8.8 words) than the original questions (5.5 words); Figure \ref{fig:question-length} shows the question length distribution.

        Most conversations were annotated by only one annotator, but $50$ conversations were annotated by both.
        We relied on these conversations to analyse the annotations.
        We extracted two rewritings per question and per annotator and,
        using a pair of rewritings as references and the other as hypothesis, we computed the SacreBLEU score~\citep{post-2018-call} and the BERT-score~\citep{bert-score}. SacreBLEU gives us an insight on the similarity of the surface form of rewritings, while BERT-score gives us an insight on the semantic similarity.
        We obtained a SacreBLEU score of $32.67$ and a BERT-score of $90.22$: this suggests that the rewritings have diverse surface form while being close in terms of meaning.

        \begin{figure}[t]
            \centering
            \begin{tikzpicture}\centering
            \begin{axis}[
                ybar,
                xmajorgrids = false,
                height = 4cm,
                width = \columnwidth,
                major tick length = 0pt,
                minor tick length = 0pt,
                ymajorgrids = true,
                ymin = 0, ymax = 1300,
                xtick={0,5,...,30},
                ytick={500,1000},
                xmin = 0, xmax = 26,
                xlabel={\# questions},
                ylabel={\# conversations}
            ]
            \addplot+ [ybar interval, color=black!40, draw=black] coordinates { (1, 134) (2, 105) (3, 75) (4, 53) (5, 54) (6, 80) (7, 61) (8, 51) (9, 57) (10, 1233) (11, 563) (12, 488) (13, 391) (14, 365) (15, 379) (16, 32) (17, 38) (18, 45) (19, 42) (20, 323) (21, 23) (22, 5) (23, 3) (24, 0) (25, 1) };
            \end{axis}
            \end{tikzpicture}
            \caption{Distribution of conversations' length.}
            \label{fig:dialogue-length}
        \end{figure}
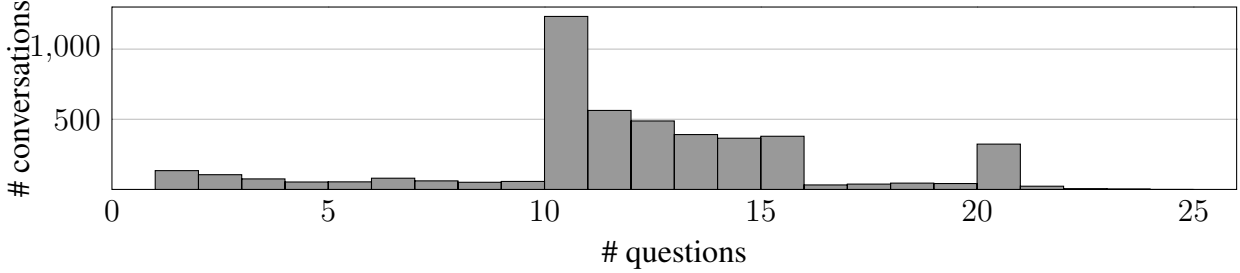
        
        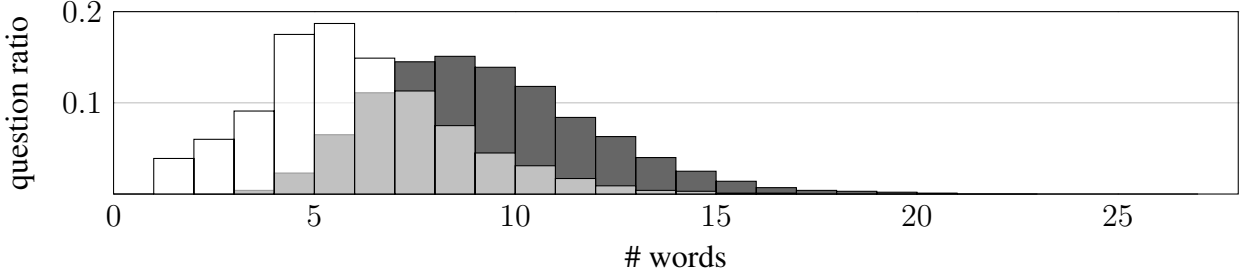
\begin{figure}[t]
            \centering
            \begin{tikzpicture}\centering
            \begin{axis}[
                ybar,
                xmajorgrids = false,
                height = 4cm,
                width = \columnwidth,
                major tick length = 0pt,
                minor tick length = 0pt,
                ymajorgrids = true,
                ymin = 0, ymax = 0.2,
                ytick={0.1,0.2},
                xtick={0,5,...,30},
                xmin = 0, xmax = 28,
                xlabel={\# words},
                ylabel={question ratio}
            ]
            \addplot+ [ybar interval, color=black!60, draw=black] coordinates{ (2, 0.0) (3, 0.004) (4, 0.023) (5, 0.065) (6, 0.111) (7, 0.145) (8, 0.151) (9, 0.139) (10, 0.118) (11, 0.084) (12, 0.063) (13, 0.04) (14, 0.025) (15, 0.014) (16, 0.007) (17, 0.004) (18, 0.003) (19, 0.002) (20, 0.001) (21, 0.0) (22, 0.0) (23, 0.0) (24, 0.0) (25, 0.0) (26, 0.0) (27, 0.0) };
            \addplot+ [ybar interval, fill=white, fill opacity=0.6, draw=black] coordinates { (0,0) (1, 0.039) (2, 0.06) (3, 0.091) (4, 0.175) (5, 0.187) (6, 0.149) (7, 0.113) (8, 0.075) (9, 0.045) (10, 0.031) (11, 0.017) (12, 0.009) (13, 0.004) (14, 0.003) (15, 0.001) (16, 0.001) (17, 0.0) (18, 0.0) (19, 0.0) (20, 0.0) (21, 0.0) (22, 0.0) (23, 0.0) };
            \end{axis}
            \end{tikzpicture}
            \caption{Distribution of length
            for original questions (white) and out-of-context rewritings (dark grey). Overlap of the distribution is light grey.}
            \label{fig:question-length}
        \end{figure}  

    \subsection{Evaluation on Question Rewriting (QR)}\label{sec:nlp}

        In QR, the model receives as input an in-context question, its conversational context, and the associated passage. Its task is to generate an out-of-context rewriting of the question.
        We conducted the following experiment: (1) training QR models on \ac{CoQAR} and CANARD; (2) evaluating these models, via standard metrics and human evaluation. Furthermore, we evaluate these QR models on downstream conversational question answering as presented in the next section. 
        
\label{subs:qrew}
           
        \paragraph{Datasets.}
        For training and evaluation, we rely on CANARD and \ac{CoQAR}.
        For CANARD, we use the original train/dev/test splits.
        For CoQAR, we use the original dev set as test set, and split the original train set into a train set and dev set, in such manner that CANARD and CoQAR dev sets have the same size.
        For training, we also make use of a mixture of CANARD and \ac{CoQAR}, that we refer to as \ac{CoQAR}+CANARD, whose train and dev sets are, respectively, the union of both corpora's train and dev sets.
        We train three variants of the QR model: one variant is trained on CANARD, one is trained on \ac{CoQAR}, and the third one is trained on a mixture of both datasets.
           \begin{table*}[h!]
            \centering
            \begin{tabular}{llcccc}
                \hline
                \multirow{2}{*}{Test set} & \multirow{2}{*}{Model} & \multicolumn{2}{c}{Meaning preservation} & \multicolumn{2}{c}{Linguistic correctness} \\
                
                & & {MOS} & {(Std dev.)} & {MOS} & {(Std dev.)}\\
                \hline
                \multirow{2}{1cm}{\ac{CoQAR}} & Human rewriting      & \textbf{4.5} & (0.86) & \textbf{4.86} & (0.45) \\
                                             & T5(\ac{CoQAR})      & 3.82 & (1.42) & 4.66 & (0.82) \\
                \hline
                                             
                                             & Human rewriting      & \textbf{4.60} & (0.96) & 4.7 & (0.89) \\
                CANARD                          & T5(CANARD)           & 3.92 & (1.34) & 4.43 & (1.08) \\
                                             & T5(\ac{CoQAR}+CANARD)  & 3.96 & (1.47) & \textbf{4.76}  & (0.77)  \\
                \hline
            \end{tabular}
            
            \caption{Results of the human evaluation of QR.}
            \label{tab:human_eval_rewriting}
        \end{table*}      
        
        \paragraph{Model:} We train a QR model based on T5 on three datasets: \ac{CoQAR}, CANARD, and \ac{CoQAR}+CANARD.
        For each dataset, we fine-tune the small 1.1 version of T5\footnote{\url{https://huggingface.co/google/t5-v1_1-small}}.
        The model is evaluated on the dev set using METEOR.
          
        Two Mean Opinion Score (MOS) evaluations were carried out on 8 human testers who were asked to judge the quality of rewritten questions.
        We sampled 50 original questions from \ac{CoQAR} and 50 original questions from CANARD.
        Each original question was then paired with several rewritings:
        \begin{itemize}
            \item one rewriting from the corpus, to which we refer as the \textit{reference};
            \item one or several rewritings generated by different T5 models: each source question from \ac{CoQAR} is paired with a rewriting generated by T5(\ac{CoQAR}), while each source question from CANARD is paired with one rewriting generated by T5(CANARD) and one rewriting generated by T5(\ac{CoQAR}+CANARD).
        \end{itemize}
        The pairs were then used in two evaluations.
        
        In the first evaluation, rewritten questions were presented to human testers, together with the original question and its context (preceding turns and the corresponding text passage). Testers assessed the semantic similarity of the rewritten and original questions.
        In the second evaluation, rewritten questions were presented alone to the testers for them to assess linguistic correctness.
        Both semantic similarity and linguistic correctness were evaluated on the 5-points scale.
        In the end, each rewritten question received one rating for semantic similarity and one for linguistic correctness.
        The results are reported in Table~\ref{tab:human_eval_rewriting}.

        We see that QR models obtain scores that are clearly below human performance in terms of meaning preservation. We also observe that the T5 model that was trained on \ac{CoQAR} and CANARD obtains higher linguistic correctness scores than the model that was only trained on CANARD, and this result does not seem due to chance (a Mann-Whitney U test gives a $p$-value of $0.026$). It is plausible that, although adding data from \ac{CoQAR} to the training set does not improve meaning preservation, it improves linguistic correctness because of its greater diversity in term of rewritings' surface forms.

    \subsection{Evaluation on Conversational Question Answering}
        
        Typically, the inputs to a \ac{CQA} neural model are:
        a question,
        its conversational context (i.e. the sequence of previous questions and answers),
        and the associated passage. The output is the answer, either in the form of a span from the passage or in the form of valid tokens such as ``yes", ``no" or ``unknown".
    
        A challenge for conversational question answering was also released with CoQA\footnote{\url{https://stanfordnlp.github.io/coqa/}}.  The models are evaluated with the F1 score~\citep{reddy_coqa_2019}. Transformers have been successfully used in this task: to the time this paper was written, the best model (a RoBERTa-based model~\citep{ju2019technical}) got $90.7$ of overall F1 measure, overcoming human performance $88.8$. 
        
        Our goal is to indirectly assess the quality of QR by comparing the performance of a model taking original questions and their context as inputs with a model using out-of-context rewritings instead. In other words, we would like to know whether replacing the original question with its conversational context by the out-of-context rewriting has a positive impact on answer extraction. First, we evaluate the impact of rewritten questions in the performance of a RoBERTa baseline~\citep{liu2019roberta}.
        Second, in order to assess the reusability of QR models trained on \ac{CoQAR}, we further evaluate a state-of-the-art non-conversational QA model trained on SQuAD~\citep{rajpurkar_know_2018} by testing it with the rewritten questions.  

     We would like to assess the impact of QR on state-of-the art models for CQA by answering the following question: would the models be able to extract the correct answer from the passage without dealing with the conversational context? To this aim we propose three experiments in which we train and evaluate a transformer on several variations of QR: no rewriting, human rewriting, and model rewriting.
     
        \begin{table}[t]
            \centering
            \setlength{\tabcolsep}{10pt}
            \begin{tabular}{lcc}\hline
                 QR mechanism&F1&EM  \\\hline
                 None (question+context)&\bf{68.13}&\bf{49.63} \\
                 Human rewriting&63.26&45.10 \\ 
                 T5(\ac{CoQAR}+CANARD)&63.30&44.97\\
                 
                 \hline
            \end{tabular}
            \caption{Results of the CQA evaluation.}
            \label{tab:convqares}
        \end{table}   
     \paragraph{Datasets and Variants.} We use \ac{CoQAR}, with distinct rewriting.
     \begin{enumerate}[i]
         \item No rewriting: the orginal dataset, taking into account the conversational context.
         \item Human rewriting:  the dataset containing only the question rewritten by human annotators, ignoring completely the conversational context.
         \item QR model: instead of using human annotations we use questions that were generated automatically by the T5(\ac{CoQAR}+CANARD) model presented in Section~\ref{subs:qrew}.
     \end{enumerate}
   
     \paragraph{Model.} For the CQA experiments, we train and evaluate a RoBERTa\footnote{\url{https://huggingface.co/}} transformer on \ac{CoQAR} with the distinct rewriting mechanisms described above. 
        
    \paragraph{Evaluation.}
    Results are presented in Table~\ref{tab:convqares}.
    Surprisingly, resolving the context with human question rewriting does not seem to help RoBERTa to better identify the answer in terms of F1 and exact match (EM) as defined in~\cite{rajpurkar2016squad}. We obtained an  F1 and EM gain of $4.87$ and  $4.53$ respectively of the original in-context questions over the out-of-context human rewritings.
   
   Unlike ~\citep{vakulenko2021question}, where results of the same task are reported on CANARD, the setting relying on original questions (referred to as CANARD\_O)
    and the one relying on human-written questions (CANARD\_H)
    respectively obtain $53.65$ and $57.12$ F1 scores,
    which correspond to a gain of $3.47$ points for human rewriting.
    We suspect that the self-attention mechanism of RoBERTa solves the coreferences and ellipsis present in short in-context questions limited by the separation token from the context and the passage. While processing a long self-contained rewriting might be more difficult.  These results confirm the good performance of RoBERTa  on the original task of CQA \citep{ju2019technical}.
  
    Interestingly, automatically rewritten questions trained on both \ac{CoQAR} and CANARD obtained similar performance than human rewritings, although human rewriting, got a slightly better EM.  These results are comparable with the ones reported on CANARD in~\citep{vakulenko2021question}.

We also conducted experiments on the re-usability of \ac{QA} systems by solving the context through question rewriting. The results are promising, rewriting out-of-context question will let us reuse existing \ac{QA} systems.  We invite the reader to look for details in the paper~\citep{brabant-etal-2022-coqar}.
    
\section{A conversational QA corpus grounded in Wikidata}
\label{s:KGConv}

   After the great success of ChatGPT that spread out nonfactual generative neural models to the big audience, guiding semantically these models to enable explainability and to reduce their typical errors: hallucinations, distortions, omissions and repetitions\citep{faille-etal-2021-entity-based, narayan-etal-2022-well,nie2019simple} is an urgent  need.   We propose \ac{KGConv}\footnote{\url{https://github.com/Orange-OpenSource/KGConv}}, a corpus of Conversational Question Answering (CQA) grounded on Wikidata\footnote{\url{https://www.wikidata.org/}} to constraint the generation with a \ac{KG}.   
    
    \ac{KGConv} is composed of conversations between two participants, one that always asks questions and another one that always answers based on facts.  Thus, it contains sequences of question-answer pairs. The grounded sequences are composed of Wikidata triples of the form: $(s,p,o)$, in which $s$ is the subject, $p$ is the property and $o$ corresponds to the object of a fact belonging to a \ac{KG}.
    
        \begin{table*}[t!]
        \newcommand{\interQspace}{0.15cm}
        \center
        \footnotesize
        \begin{tabular}{rrrl}
        \toprule
        \#1 & \multicolumn{2}{r}{\textbf{Triple}} & \triple{NGC 4833}{part of}{Milky Way}\\[\interQspace]
        & \multirow{6}{*}{\rotatebox[origin=l]{90}{\bf Question variants}}
        &    original & NGC 4833 is part of what astronomical object? \\
        & &  subject & NGC 4833 \\
        & &  rewritten & NGC 4833 is part of what astronomical object? \\[\interQspace]
        & &  original & Where is NGC 4833 located? \\
        & &  subject & NGC 4833 \\
        & &  rewritten & Where is NGC 4833 located? \\[\interQspace]
        & \multicolumn{2}{r}{\textbf{Answer}} & Milky Way \\[0.10cm]
        \toprule
        \#2 & \multicolumn{2}{r}{\textbf{Triple}} & \triple{NGC 4833}{discoverer or inventor}{Nicolas Louis de Lacaille}\\[\interQspace]
        & \multirow{10}{*}{\rotatebox[origin=l]{90}{\bf Question variants}}
        &    original & Who was behind the discovery of NGC 4833? \\
        & &  subject & NGC 4833 \\
        & &  rewritten & Who was behind the discovery? \\[\interQspace]
        & &  original & What was the name of the discoverer of NGC 4833? \\
        & &  subject & NGC 4833 \\
        & &  rewritten & Who discovered this object? \\[\interQspace]
        & &  original & Who found NGC 4833? \\
        & &  subject & NGC 4833 \\
        & &  rewritten & Who found this object? \\[\interQspace]
        & \multicolumn{2}{r}{\textbf{Answer}} & Nicolas Louis de Lacaille \\[0.10cm]
        \toprule
        \#3 & \multicolumn{2}{r}{\textbf{Triple}} & \triple{Nicolas Louis de Lacaille}{religion or worldview}{Catholic Church}\\[\interQspace]
        & \multirow{6}{*}{\rotatebox[origin=l]{90}{\bf Question variants}}
        &   original & What was his religion? \\
        & & subject & his \\
        & & rewritten & What was his religion? \\[\interQspace]
        & & original & What faith did he follow? \\
        & & subject & he \\
        & & rewritten & What faith did he follow? \\[\interQspace]
        & \multicolumn{2}{r}{\textbf{Answer}} & Catholic Church \\
        \bottomrule
        \end{tabular}
        \caption{Excerpt of a question-answer conversation along with the related triples. The root entity is NGC 4833, from the theme ``space object''. The rewritten corresponds to the in-context question that has been automatically generated by a T5 model.}
        \label{tab:conversation_example}
    \end{table*}

    In total \ac{KGConv} gathers $71$K~conversations ($604$K question-answer pairs in total), where each pair relates to an underlying fact from the public KG Wikidata\footnote{\url{https://www.wikidata.org/}}.  Each conversation is focused on a given \emph{root} entity. As illustrated by Table~\ref{tab:conversation_example}, the first question is directly about this root entity, while the next ones explore new facts about any entity discovered during the conversation (including the root entity itself). This corpus can be used for distinct tasks such as factual question generation, question rewriting as well as generation of sequence of questions and answers from a given Knowledge-graph or vice-versa.

    \begin{table*}[ht!]
        \centering
        \small
        \begin{tabular}{r|cccc|cccc}
        \toprule
        \rotatebox[origin=l]{90}{~~~~}
        & \multirow{2}{*}[-0.5em]{\rotatebox[origin=l]{90}{entities}} & \multirow{2}{*}[0.5em]{\rotatebox[origin=l]{90}{properties}} & \multirow{2}{*}[-1em]{\rotatebox[origin=l]{90}{triples}} & \multirow{2}{*}[-1.4em]{\rotatebox[origin=l]{90}{conv.}} & \multicolumn{4}{c}{questions} \\
       & & & & & \rotatebox[origin=l]{90}{~~train~~} & \rotatebox[origin=l]{90}{~~dev~~} & \rotatebox[origin=l]{90}{~~test~~} & \rotatebox[origin=l]{90}{~~total~~} \\[-2ex] \hline
       person                 &     31671 &         327 &    71915 &      25918 &  184939 &  29352 &   11386 &  225677  \\
country                &      2171 &         171 &     3475 &        703 &    5085 &    817 &     214 &    6116 \\
ideology               &      1220 &         169 &     1677 &        450 &    3112 &    581 &     228 &    3921\\
space object           &      2586 &         116 &     6360 &       5961 &       0 &      0 &   50158 &   50158 \\
molecular entity       &     17798 &         151 &    38314 &      23033 &  154511 &  24587 &    9531 &  188629  \\
historical event       &      4695 &         189 &     7770 &       4972 &   35270 &   5684 &    2247 &   43201  \\
food                   &      2532 &         166 &     4012 &       2099 &   15050 &   2230 &    1011 &   18291  \\
taxon                  &      3190 &         215 &     5408 &       1902 &       0 &      0 &   16099 &   16099  \\
with\_unseen\_properties &     13651 &         404 &    24123 &       5558 &       0 &      0 &   51813 &   51813 \\
whole dataset                  &     63345 &         458 &   142691 &      70596 &  397967 &  63251 &  142687 &  603905  \\

        \bottomrule
        \end{tabular}
        \caption{For each theme, the table gives: the number of different entities and properties appearing in conversations, the number of conversations, and the number of questions for each split. Note that in the entities and properties columns, the ``total'' values are not the sum of the cells above; this is because some entities and properties appear in several themes.}
        \label{tab:general_stats}
    \end{table*}

    This is ongoing work. The corpus has been released publicly and a paper presenting the corpus will be submitted for publication soon.
    This dataset was also used by the PhD student Juliette Faille supervised by Claire Gardent as part of the collaboration with Orange in the \ac{ITN} European Project \ac{NL4XAI}. Particularly, it was her subject of study during her secondment at Orange. Her work focused on studying explainability and factual question generation.

\section{Other Contributions: Graph Embeddings}
\label{ss:graphemb}
This work was made by the PhD candidate Sebastien Montella in co-supervision with Dr. Johannes Heinecke at Orange and it was published in~\citep{montella-etal-2021-hyperbolic}.
While most of dialogue systems store their knowledge using simple structures, namely a set of slot-value pairs, world knowledge is usually stored in Knowledge Graphs. 
A \ac{KG} is a collection of triples $\langle$\textit{s, p, o}$\rangle$; where \textit{s}, \textit{p} and \textit{o} stand for the \textit{subject}, \textit{predicate} and \textit{object} respectively. An \textit{entity} denotes whether a subject or an object and a \textit{relation} denotes a predicate that links two entities.
The main contributions of this work are: (i) it explores graph embeddings in the hyperbolic space instead of the Euclidian space; (ii) it considers the time parameter; (iii) it propose a hyperbolic model aware of time and (iv) it compares the performance of state-of-the-art models that takes into consideration time versus models that use negative sampling.

Since KGs are sometimes incomplete, one important task of \ac{NLP} is Link Prediction (LP), which consists in predicting the missing connections between entities. This task can help for instance to build knowledge graphs on the fly. Each entity and relation are map into a vector space to learn low-dimensional embeddings such that, valid triples maximise a defined scoring function and that fallacious triples minimise it. An approach is efficient if it can model multiple relational patterns. Some predicates are symmetric (\textit{e.g. marriedTo}), asymmetric (\textit{e.g. fatherOf}), an inversion of another relation (\textit{e.g. fatherOf} and \textit{childOf}) or a composition (\textit{e.g. grandfatherOf}). Hierarchical relations have remained challenging to model in Euclidean space, while hyperbolic geometry reveals to be a strong asset to capture hierarchical patterns. 
Nevertheless, the aforementioned approaches represent embeddings as invariant to time. For example, the triple $\langle$\textit{Donald Trump, presidentOf, U.S.}$\rangle$ is not longer correct in $2022$. 

This work shows that an optimised number of negative samples enables the state-of-the-art model \textsc{AttH} \citep{chami-etal-2020-low} to reach competitive or even better performance on temporal link prediction while being unaware of the temporal aspect. It also introduces an extension of \textsc{AttH}, namely \textsc{Hercules}\footnote{\underline{\textbf{H}}yperbolic Representation with Tim\underline{\textbf{E}} and \underline{\textbf{R}}elational \underline{\textbf{CU}}rvatures for Tempora\underline{\textbf{L}} Knowledg\underline{\textbf{E}} Graph\underline{\textbf{S}}}. 

This was the first attempt to leverage the curvature of a manifold to coerce time-aware representation. An ablation study of distinct curvature definitions has been done to investigate the compelling results of \textsc{AttH} over time-aware models.

\subsubsection{Problem Definition}
Lets consider a valid quadruplet $\langle$\textit{s, p, o, t}$\rangle$ $\in \mathcal{S} \subset \mathcal{E}\times\mathcal{R}\times\mathcal{E}\times\mathcal{T}$, with $\mathcal{E}$, $\mathcal{R}$ and $\mathcal{T}$ the sets of entities, relations and timestamps respectively and $\mathcal{S}$ the set of correct facts. A scoring function $f:\mathcal{E}\times\mathcal{R}\times\mathcal{E}\times\mathcal{T} \rightarrow \mathbb{R}$ is defined such that $f(s,p,o,t)$ is maximised for any quadruplet $\in$ $\mathcal{S}$, and minimised for corrupted quadruplet ($\notin$ $\mathcal{S}$). Throughout the optimisation of the foregoing constraint, representations of entities, relations and times are learned accordingly. The resulting embeddings should then capture the multi-relational graph structure. Thus, $f$ is measuring the probability that an entity \textit{s} is connected to an entity \textit{o} by the relation \textit{p} at time \textit{t}. 

\subsubsection{Hyperbolic Geometry}
Hyperbolic geometry belongs to non-Euclidean geometry. In contrast to Euclidean geometry relying on Euclid's axioms \citep{euclid-et-al-1956-element}, non-Euclidean geometry rejects the fifth axiom known as the parallel postulate. It states that given a point $x$ and a line $l_{1}$, there exists a unique line $l_{2}$ parallel to $l_{1}$ passing through $x$. This is only possible due to a (constant) zero curvature of the space. The curvature defines how much the geometry differs from being flat. The higher the absolute curvature, the curvier. Euclidean space has a zero curvature hence called \textit{flat space}. When represented in an Euclidean space, straight lines become curved, termed as \textit{geodesics} (Fig. \ref{fig:poincare_ball}).

\begin{figure}[h]
    \centering
    \begin{tikzpicture}[
  point/.style = {draw, circle, fill=black, inner sep=0.7pt},
]
\def\rad{2cm}
\definecolor{manifold_color}{rgb}{0.80,0.98,0.98}
\coordinate (O) at (0,0); 
\coordinate (N) at (0,\rad); 

\filldraw[ball color=manifold_color] (O) circle [radius=\rad];
\draw[dashed] 
  (\rad,0) arc [start angle=0,end angle=180,x radius=\rad,y radius=5mm];
\draw
  (\rad,0) arc [start angle=0,end angle=-180,x radius=\rad,y radius=5mm];
\begin{scope}[xslant=0.5,yshift=\rad,xshift=-2]
\filldraw[fill=gray!50,opacity=0.3]
  (-3,1) -- (3,1) -- (3,-1) -- (-3,-1) -- cycle;

  (-4,1) -- (3,1) -- (3,-1) -- (-4,-1) -- cycle;
\node at (2,0.6) {$\mathcal{T}_{x}^{c}\mathbb{B}^{n, c}$};  
\node at (1.2 * \rad,-2 * \rad) {$\mathbb{B}^{n, c}$};
\end{scope}
\draw[color=red, ->, thick] (0, \rad) -- (2.3, 1.4);
\node at (1.25,1.9) {$u$};

\draw[dotted, color=blue, thick, ->] (2.3, 1.4) arc (5:-50:1.5) coordinate (c);
\node[text=blue] at ($(c) + (-0.75,-0.06)$) {$\exp_{x}^{c}(\textcolor{black}{u})$} ;

\draw[dotted, color=red, thick, ->] ($ (c) + (0.3,0) $) arc (-50:8:1.35) coordinate (d);
\node[text=red] at ($(d) + (0.6,-0.8)$) {$\log_{x}^{c}(\textcolor{black}{v})$} ;

\draw[dashed, color=blue, thick] (0, \rad) arc (250:215:-2.8) coordinate (e) ;
\draw[solid, color=blue, thick, ->] (e) arc (215:198:-2.8) ;
\node at ($(e) + (-0.25, -0.2)$) {$v$};

\draw[dashed]
  (N) node[above] {$x$} -- (O) node[below] {$O$};
\node[point] at (N) {};
\end{tikzpicture}
    \caption{Illustration of the exponential and logarithmic maps between the Poincar\'e ball $\mathbb{B}^{n, c}$ and the tangent space  $\mathcal{T}_{x}^{c}\mathbb{B}^{n, c}$.}
    \label{fig:poincare_ball}
\end{figure}
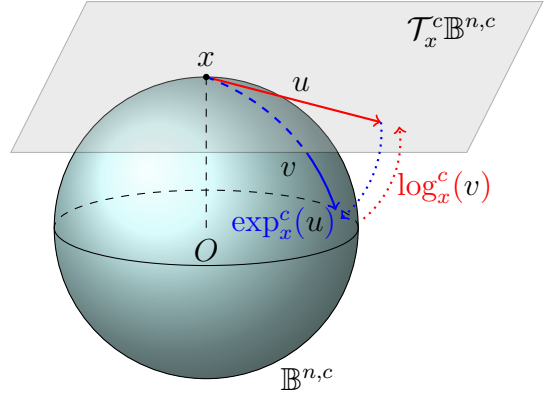

Hyperbolic geometry comes with a constant negative curvature. The interested reader is referred to formal definitions in~\citep{montella-etal-2021-hyperbolic}.
\subsubsection{From \textsc{AttH} to \textsc{Hercules}}
Given a quadruplet, $\langle$\textit{s, p, o, t}$\rangle$, we note $e_{s}^{H}$, $r_{p}^{H}$ and $e_{o}^{H}$ the hyperbolic embeddings of the subject, predicate and object respectively.\footnote{Since \textsc{AttH} is not considering time, the parameter $t$ is not used.} \textsc{AttH} uses relation-specific embeddings, rotations, reflections and curvatures. The curvature is defined as depending on the corresponding relation \textit{p} involved. Precisely, a relation \textit{p} is attributed with an individual parametric curvature $c_{p}$. The curvature $c_{p}$ is defined in Eq. \ref{eq:curvature_definition} as:
\begin{equation}
\label{eq:curvature_definition}
    c_p = \sigma(\mu_p)
\end{equation}
where $\mu_p$ is a trainable parameter $ \in \mathbb{R}$ and $\sigma$ is a smooth approximation of the ReLU activation function defined in $[0, +\infty]$.
With such approach, the geometry of the manifold is learned, thus modified for a particular predicate. The curvature dictates how the manifold is shaped. Changing the curvature of the manifold implies changing the positions of projected points. 
This means that for distinct relations, the same entity will have different positions because of the different resulting geometries for each relation. For example, lets consider the triples $t_1:=\langle$\textit{Barack Obama, visit, France}$\rangle$ and $t_2 := \langle$\textit{Barack Obama, cooperate, France}$\rangle$. The Euclidean representations of entities \textit{Barack Obama} and \textit{France} from both facts will be projected onto the riemannian manifold. However, the structure (i.e. curvature) of the manifold changes as a function of the relation of each fact (\textit{i.e.} '\textit{visit}' and '\textit{cooperate}'). Therefore, the resulting hyperbolic embbeding of \textit{Barack Obama} of $t_1$ will not be the same resulting hyperbolic embedding of \textit{Barack Obama} in $t_2$. By analogy, the same holds for entity \textit{France}. 

In order to learn rotations and reflections, \textsc{AttH} uses 2 $\times$ 2 Givens transformations matrices~\citep{chami-etal-2020-low}. Those transformations conserve relative distances in hyperbolic space and can therefore directly be applied to hyperbolic embeddings (isometries).
Furthermore, \textsc{AttH} utilizes an hyperbolic attention mechanism to represent complex relations that can be a mixture of rotation and reflection. The attention scores are computed in the tangent space by projecting the hyperbolic rotation embedding and hyperbolic reflection embedding with the logarithmic map into the euclidian space, as shown in Figure~\ref{fig:poincare_ball}. Then, the attention vector is mapped back to manifold using the exponential map.
We propose \textsc{Hercules}, a time-aware extension of \textsc{AttH}. \textsc{Hercules} redefines the curvature of the manifold as being the product of both relation and time.
The main intuition of \textsc{Hercules} is that both relation and time directly adjust the geometry of the manifold such that the positions of projected entities are relation-and-time-dependent. This is advantageous in that no additional temporal parameters per entity are needed. Since the whole geometry has changed for specific relation and time, all future projections onto that manifold will be aligned to the corresponding relation and timestamp. We investigate different curvature definitions and time translation in our experiments (see the next Section). The scoring function of \textsc{Hercules} remains same as \textsc{AttH}. 

When learning hyperbolic parameters, the optimisation requires to utilise a Riemannian gradient \citep{bonnabel-et-al-2011-rsgd}. However, proven to be challenging, we instead learn all embeddings in the Euclidean space. The embeddings can then be mapped to the manifold using the exponential map. This allows the use of standard Euclidean optimisation strategies.

\subsubsection{Experiments}
\label{sss:experiments}
\paragraph{Datasets}
For fair comparisons, we test our model on same benchamark datasets used in previous works, \textit{i.e.} ICEWS14 and ICEWS05-15. Both datasets were constructed by \citep{garcia-duran-etal-2018-learning} using the Integrated Crisis Early Warning System (ICEWS) dataset \citep{boschee-et-al-2018-icews}. ICEWS provides geopolitical information with their corresponding (event) date, \textit{e.g.} $\langle$\textit{Barack Obama, visits, France, 2009-03-11}$\rangle$. More specifically, ICEWS14 includes events that happened in 2014 whereas ICEWS05-15 encompasses facts that appeared between 2005 and 2015. We give the original datasets statistics in Table \ref{table:datasets_stats}. To increase the number of samples, for each quadruplet $\langle$\textit{s, p, o, t}$\rangle$ we add $\langle$\textit{s, $p^{-1}$, o, t}$\rangle$, where $p^{-1}$ is the inverse relation of \textit{p}. This is a standard data augmentation technique usually used in LP \citep{balazevic-et-al-2019-murp, goel-et-al-2020-diachronic, han-etal-2020-dyernie}.

\begin{table*}
    \centering
    \begin{tabular}{cccccccc}
    \hline
         \textbf{Datasets} & $\boldsymbol{\lvert \mathcal{E} \rvert}$ & $\boldsymbol{\lvert \mathcal{R} \rvert}$ & $\boldsymbol{\lvert \mathcal{T} \rvert}$ & \textbf{Training} & \textbf{Validation} & \textbf{Test}  \\ \hline
         ICEWS14 & 7,128 & 230 & 365 & 72,128 & 8,941 &
         8,963 \\ \hline
         ICEWS05-15 & 10,488 & 251 & 4017 & 368,962 & 46,275 & 46,092 \\
         \hline
 
 \end{tabular}
    \caption{ICEWS14 and ICEWS05-15 Datasets Statistics}
    \label{table:datasets_stats}
\end{table*}
\paragraph{Evaluation Protocol \& Metrics}
Given a (golden) test triple $\langle$\textit{s, p, o, t}$\rangle$, for each entity $s' \in \mathcal{E}$, we interchange the subject \textit{s} with $s'$ and apply the scoring function $f$ on the resulting query $\langle s'$\textit{, p, o, t}$\rangle$. Since replacing \textit{s} by all possible entity $s'$ may end up with a correct facts, we filter out those valid quadruplets and give them extremely low scores to avoid correct quadruplets to be scored higher than the tested quadruplet in final ranking \citep{bordes-et-al-2013-transe}. We then rank the entities based on their scores in descending order. We store the rank of the correct entity \textit{s} noted $z_{s}$. Thus, the model should maximize the returned score for the entity \textit{s} such that $z_{s}=1$. The same process is done using the object $o$.

To evaluate our models, we make use of the Mean Reciprocal Rank (MRR). We also provide the Hits@1 (H@1), Hits@3 (H@3) and Hits@10 (H@10) which assess on the frequency that the valid entity is in the top-1, top-3 and top-10 position, respectively.

\paragraph{Results}
We provide link prediction results on ICEWS14 and ICEWS05-15 for \textsc{AttH}, \textsc{Hercules} and different models from the literature. As \citep{han-etal-2020-dyernie}, we adopted a dimension analysis to investigate behaviors and robustness of approaches. When possible, we re-run official implementation of models. Otherwise, official or best results in literature are reported. Results are shown in Table \ref{table:results_link_prediction}.

As expected, hyperbolic-based strategies (\textit{i.e.} \textsc{DyERNIE}, \textsc{AttH} and \textsc{Hercules}) perform much better at lower dimensions, outperforming most of other approaches with ten times less dimensions. We report an average absolute gain of 11.6\% points in MRR with only 10 dimensions over the median performance of other approaches with 100 dimensions. This strengthens the effectiveness of hyperbolic geometry to induce high-quality embeddings with few parameters.

Astonishingly, we notice that \textsc{AttH} model is highly competitive despite the absence of time parameter. \textsc{AttH} exhibits new state-of-the-art or statistically equivalent performances compared to \textsc{DyERNIE} and \textsc{Hercules}. We remark no statistically significant differences in performances between hyperbolic models.\footnote{\label{stests}We performed the Mixed-Factorial Analysis of Variance (ANOVA), in which the independent variables are \textit{the dimension} and \textit{the model} and the dependent variable is \textit{the metric}. We consider two groups one for each dataset. We report \textit{p}-values of 0.842, 0.872, 0.926 and 0.229 for MRR, H\char64{}1, H\char64{}3 and {H\char64{}10} respectively.} Importantly, unlike other research carried out in this area, time information here does not lead to any notable gain. This seems to indicate that other parameters should be considered. We examine this phenomenon in section
\ref{section:time_is_all_you_need_negative}.

On ICEWS14, for \textit{dim} $\in \{20, 40, 100\}$, both \textsc{AttH} and \textsc{Hercules} outperform \textsc{DyERNIE} by a large margin. We witness an improvement of 2.5\% and 5\% points in MRR and Hits@1 with 100-dimensional embeddings. On ICEWS05-15, \textsc{Atth} and \textsc{Hercules} yield comparable achievements with the state-of-the-art. In contrast to \textsc{DyERNIE}, it is noteworthy that \textsc{AttH} and \textsc{Hercules} utilize a single manifold while reaching top performances. 

We also distinguish tempered results on Hits@10 metric for \textsc{AttH} and \textsc{Hercules} models. This suggests that during optimization, \textsc{AttH} and \textsc{Hercules} favor ranking some entities on top while harming the representation of others.



\begin{table*}[th!]
\centering
\begin{tabular}{c|c|cccc|cccc}
\hline
\multicolumn{2}{c}{Datasets} & \multicolumn{4}{|c}{{ICEWS14 (filtered)}} & \multicolumn{4}{|c}{{ICEWS05-15 (filtered)}} \\
\hline
{\textit{dim}} & {Model} & {MRR} & {H@1} & {H@3} & {H@10} & {MRR} & {H@1} & {H@3} & {H@10} \\ \hline 


{} & \textsc{ATISE\textsuperscript{\ding{61}}} & {18.0} & {3.03} & {23.9} & {48.7} & {15.9} & {4.35} & {19.22} & {41.0} \\

{} & \textsc{TeRo\textsuperscript{\ding{61}}} & {7.25} & {2.39} & {6.40} & {16.6} & {10.3} & {3.54} & {10.1} & {23.2} \\

{10} & \textsc{DyERNIE\textsuperscript{\ding{83}}} & \textbf{46.2} & \textbf{36.0} & {51.1} & \underline{66.3} & \textbf{58.9} & \textbf{50.5} & \textbf{63.2} & \textbf{75.1} \\
\cline{2-10}

{} & \textsc{Hercules} & \underline{46.0} & \underline{34.9} & \textbf{52.4} & {66.0} & \underline{54.7} & \underline{43.8} & \underline{61.8} & {73.2} \\

{} & \textsc{AttH} & {45.6} & {34.2} & \underline{52.0} & \textbf{66.4} & {49.9} & {34.4} & {61.6} & \underline{73.6} \\

\hline 

{} & \textsc{ATISE\textsuperscript{\ding{61}}} & {19.1} & {1.28} & {28.2} & {54.7} & {24.5} & {7.67} & {32.3} & {59.2} \\

{} & \textsc{TeRo\textsuperscript{\ding{61}}} & {24.5} & {13.8} & {28.01} & {46.3} & {27.1} & {13.5} & {33.3} & {54.1} \\

{20} & \textsc{DyERNIE\textsuperscript{\ding{83}}} & {53.9} & {44.2} & {58.9} & \textbf{72.7} & \textbf{64.2} & \textbf{56.5} & \textbf{68.2} & \textbf{79.0} \\

\cline{2-10}

{} & \textsc{Hercules} & \textbf{55.5} & \textbf{47.2} & \underline{59.4} & \underline{71.4} & {63.2} & {55.2} & \underline{67.7} & \underline{77.6} \\

{} & \textsc{AttH} & \underline{55.2} & \underline{46.7} & \textbf{59.7} & \underline{71.4} & \underline{63.5} & \underline{55.8} & \underline{67.7} & {77.5} \\

\hline 

{} & \textsc{ATISE\textsuperscript{\ding{61}}} & {38.4} & {23.3} & {47.6} & {67.3} & {35.7} & {19.2} & {44.3} & {69.1} \\

{} & \textsc{TeRo\textsuperscript{\ding{61}}} & {35.1} & {22.7} & {40.5} & {60.8} & {28.3} & {12.7} & {35.3} & {60.5} \\

{40} & \textsc{DyERNIE\textsuperscript{\ding{83}}} & {58.8} & {49.8} & {63.8} & \textbf{76.1} & \textbf{68.9} & {61.8} & \textbf{72.8} & \textbf{82.5} \\

\cline{2-10}

{} & \textsc{Hercules} & \underline{61.2} & \underline{54.3} & \underline{64.7} & {74.1} & \underline{68.5} & \textbf{62.1} & \underline{72.0} & \underline{80.9} \\

{} & \textsc{AttH} & \textbf{61.7} & \textbf{54.5} & \textbf{65.4} & \underline{75.4} & \underline{68.5} & \underline{62.0} & {71.9} & {80.6} \\

\hline 

{} & \textsc{TransE\textsuperscript{\ding{83}}} & {30.0} & {14.8} & {42.7} & {60.1} & {30.4} & {13.3} & {42.4} & {61.1} \\ 

{} & \textsc{DistMult\textsuperscript{\ding{83}}} & {57.5} & {46.9} & {64.2} & {77.9} & {47.1} & {33.6} & {55.1} & {72.5} \\ 

{} & \textsc{ComplEx\textsuperscript{\ding{83}}} & {49.3} & {36.6} & {56.2} & {74.2} & {39.0} & {22.9} & {49.2} & {68.4} \\ 

{} & \textsc{TTransE\textsuperscript{\ding{83}}} & {34.4} & {25.7} & {38.3} & {51.3} & {35.6} & {15.4} & {51.1} & {67.6} \\ 

{} & \textsc{TComplEx\textsuperscript{\ding{83}}} & {31.8} & {12.9} & {45.7} & {63.0} & {45.1} & {36.3} & {49.2} & {62.0} \\

{100} & \textsc{HyTE\textsuperscript{\ding{83}}} & {33.1} & {6.8} & {54.5} & {73.6} & {38.1} & {7.6} & {65.0} & {80.4} \\

{} & \textsc{ATISE\textsuperscript{\ding{61}}} & {52.2} & {41.0} & {60.0} & {72.7} & {47.0} & {32.4} & {55.5} & {76.4} \\

{} & \textsc{TeRo\textsuperscript{\ding{61}}} & {45.4} & {34.0} & {52.2} & {67.0} & {41.1} & {26.3} & {48.9} & {71.7} \\

{} & \textsc{DyERNIE\textsuperscript{\ding{83}}} & {66.9} & \underline{59.9} & \underline{71.4} & \textbf{79.7} & \textbf{73.9} & \underline{67.9} & \textbf{77.3} & \textbf{85.5} \\

\cline{2-10}

{} & \textsc{Hercules} & \underline{69.4} & \textbf{65.0} & \underline{71.4} & {77.9} & {73.5} & \textbf{68.6} & \underline{76.1} & \underline{82.9} \\

{} & \textsc{AttH} & \textbf{69.5} & \textbf{65.0} & \textbf{71.5} & \underline{78.2} & \underline{73.6} & \textbf{68.6} & {76.0} & \underline{82.9} \\

\hline 

\end{tabular}
\caption{Link prediction results on ICEWS14 and ICEWS05-15 datasets: (\ding{61}) results are obtained using the official implementation of \citep{xu-etal-2020-tero}, (\ding{83}) results are taken from \citep{han-etal-2020-dyernie}. For each dimension (\textit{i.e. dim}), best results are in bold and second-to-best underlined. No statistically significant differences in performance are observed between \textsc{DyERNIE}, \textsc{Hercules} and \textsc{AttH}.}
\label{table:results_link_prediction}
\end{table*}
\subsubsection{Time Awareness vs Negative Sampling.}
\label{section:time_is_all_you_need_negative}
First, besides time translation, we probe different curvature definitions to identify fluctuation in performances. We analyse how time information alters the LP results by adding time as part of the curvature (\textit{i.e.} \textsc{Hercules}) and as a translation. We also explore if incorporating the Euclidean dot product of the subject and object embeddings (noted $\langle e_{s}^{\mathbb{E}},\:e_{o}^{\mathbb{E}}\rangle$) into the curvature helps to learn a better geometry. An ablation study is given in Table \ref{tab:ablation}.
 
\begin{table*}[ht!]
    \centering
    \begin{tabular}{cccc||c|c|c|c}
    \hline
    \begin{tabular}[c]{@{}c@{}}Relation\\ Curvature\end{tabular} & \begin{tabular}[c]{@{}c@{}}Time\\ Curvature\end{tabular} & \begin{tabular}[c]{@{}c@{}}Time\\ Translation\end{tabular} & \begin{tabular}[c]{@{}c@{}}$\langle e_{s}^{\mathbb{E}},\:e_{o}^{\mathbb{E}}\rangle$\\ Curvature\end{tabular} & {MRR} & {H@1} & {H@3} & {H@10} \\ \hline
    {\ding{51}} &\ding{55}&\ding{55}&\ding{55}& {61.7} & {54.5} & {65.4} & {75.4} \\ 
    {\ding{51}} &{\ding{51}}&\ding{55}&\ding{55}& {61.2} & {54.3} & {64.7} & {74.1} \\ 
    {\ding{51}} &{\ding{51}}&{\ding{51}}&\ding{55}& {60.1} & {52.1} & {64.5} & {75.0} \\ 
    {\ding{51}} &{\ding{51}}&{\ding{51}}&{\ding{51}}& {49.5} & {38.9} & {55.4} & {69.2} \\ \hline
    \end{tabular}
    \caption{Ablation study: Link prediction results on ICEWS14 using \textsc{AttH} (\textit{dim} = 40) with different curvature definitions and time translation applied.}
    \label{tab:ablation}
\end{table*}

Albeit counter-intuitive, we observe that our results corroborate with our initial finding: time information is not the culprit of our high performances. More strikingly, a simple relational curvature 
(\textit{i.e.}\ \textsc{AttH}) is sufficient to perform best on ICEWS14 (\textit{dim} = 40). Neither the inclusion of a time translation, similarly to \textsc{TTransE}, nor the Euclidean dot product provide interesting outcomes.

We then probe the sensitivity of \textsc{Hercules} towards temporal feature by performing LP with incorrect timestamps. Our intuition is to inspect whether feeding invalid timestamps during evaluation exhibits significant variation or not compared to the reference performances, \textit{i.e.} LP results with initial (non-corrupted) testing samples. To do so, for each testing quadruplet, we replace the (correct) time parameter with each possible timestamp from $\mathcal{T}$. We therefore collect multiple LP performances of \textsc{Hercules} corresponding to each distinct timestamp. Our finding is that despite erroneous timestamps, LP results show insignificant discrepancies with the initial \textsc{Hercules} performance (dashed red line). This indicates that \textsc{Hercules} gives little importance to the time parameter and thus only relies on the entity and the predicate to perform knowledge graph completion. This further highlights our finding that timestamp is not responsible for our attracting performances.

We therefore assume that the optimisation procedure may be involved. We consequently question the effect of negative sampling. Precisely, we train \textsc{Hercules} with \textit{dim} = 40 by tuning the number of negative samples between 50 to 500. 
For both, ICEWS14 and ICEWS05-15, negative sampling shows considerable gain as the number of samples increases. We record an absolute gain of 5\% points in MRR from 50 to 500 samples. We can see a rapid growth in MRR when the number of samples is inferior to 200. Adding 50 samples is equivalent to about 2\% points gain in MRR. Then, performances reach a plateau around 300 negative samples. We conjecture that a diversity in negative samples is enough to learn good representations. 
Notwithstanding that a large number of negative samples heavily constraints the location of entities in space, the resulting embeddings might benefit from it to be better positioned relatively to others.

We conclude that despite the present time parameter, an optimal negative sampling enables to reach new state-of-the-art outcome. Therefore, we argue that time is not the only parameter that should be considered when performing LP. We highlight that one should be raising awareness when training TKG representations to identify if time is truly helping to boost performances. 

\chapter{Data Collection, Annotation and Frameworks}
\label{c:datafmwks}
This chapter summarises the corpora, annotations, and frameworks in which I worked on.  It also describes briefly the national and international research projects in which I have been involved.

\section{Corpora}
This section describes briefly the work I have done in data collection and annotation.
\subsection{The French Portmedia Corpus}
\label{ss:pm}
The French MEDIA corpus collects about 70
hours of spontaneous speech (1258 dialogues,
46k utterances, 494.048 words and 4068 distinct
words) for the task of hotel reservation
and tourist information\citep{bonneau2009media}. Calls from 250 speakers to a simulated
reservation system (i.e. the Wizard-of-Oz) were
recorded and transcribed. Dialogues are full of
disfluencies, hesitations, false starts, truncations or
fillers words (e.g., euh or ben).
I worked on the semantic annotations of this corpus
as I was involved in the French ANR project PORTMEDIA~\citep{rojas-barahona2011using,rojas-barahona2011incremental}. 330 utterances were manually annotated
with semantic relations (i.e. High-Level Semantics).
This gold corpus gathers 653 head segments
and 1555 argument segments, from which
around 20 are both arguments and heads, such
as une chambre in Figure 4. 
This work focuses on annotating the semantic
structure, by segmentating  users’ utterances into  concepts. 

\begin{figure}[h!b]
\begin{center}
\scalebox{0.85}{

\begin{tikzpicture}[node distance = 1.5cm, bend angle=45, auto]
  \tikzstyle{concept} = [draw , ellipse, text width=2em, text centered, minimum height=1em]
  
  \node [concept] (at) {\tiny Attributes };
  \node [concept,below of=at] (price) {\tiny Price};
  \draw [-] (at) -- (price);
  \node [concept,right of=price] (pl) {\tiny General};
  \node [concept,below of=pl] (park) {\tiny Park};
  \draw [-] (pl) -- (park);
  \node [concept,right of=pl] (rl) {\tiny Relative};
  \node [concept,below of=rl] (near) {\tiny Near};
  \draw [-] (rl) -- (near);
  \node [concept,right of=rl] (nl) {\tiny Restaurant};
  \node [concept,right of=at] (l) {\tiny Location}
    edge (pl)edge (rl);
  \node [concept,right of=l] (per) {\tiny Person};  
  \node [concept,right of=per] (ti) {\tiny Time};  
  \node [concept,right of=nl] (h) {\tiny Hotel};  
  \node [concept,right of=h] (r) {\tiny Room};
  \node [concept,right of=ti] (lobj) {\tiny Object}
    edge (nl) edge (h)edge (r);
  \node [concept,above of=per] (t) {\tiny Thing}
    edge (at)
    edge (l)
    edge (per)
    edge (ti)
    edge (lobj);
    
\end{tikzpicture}}

\caption[]{Excerpt of MEDIA ontology}
\label{fig:onto}
\end{center}
\end{figure}
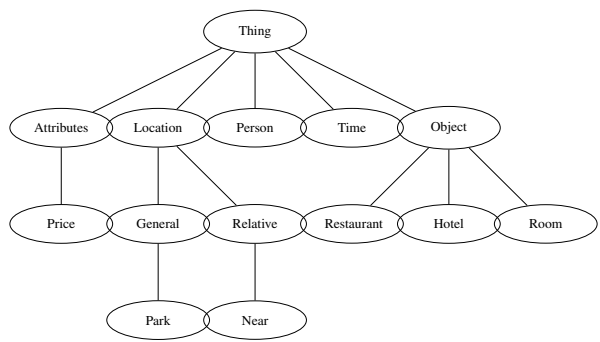

This ontology identifies the concepts that can have arguments, and we thus use this information to further
distinguish between {\it head} segments that can have arguments.

\begin{figure*}[ht]
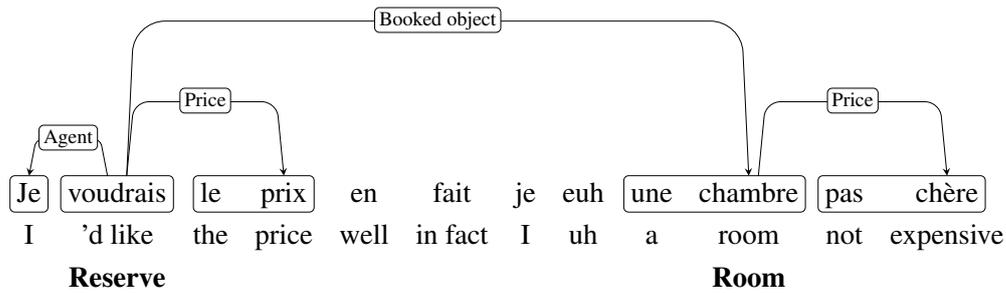

\begin{center}
\scalebox{0.9}{
\begin{dependency}
   \begin{deptext}[column sep=0.2cm, row sep=0.1cm]
  Je\&voudrais\&le\&prix\&en\&fait\&je\&euh\&une\&chambre\&pas\&ch\`ere\\
  I\& 'd like\&the\&price\&well\&in fact\&I\&uh\&a\&room\&not\&expensive\\
  \& \textbf{Reserve} \& \& \& \& \& \& \& \& \textbf{Room} \& \& \\
   \end{deptext}
   \wordgroup{1}{1}{1}{x}
   \wordgroup{1}{2}{2}{x}
   \wordgroup{1}{3}{4}{x}
   \wordgroup{1}{9}{10}{x}
   \wordgroup{1}{11}{12}{x}
   \depedge{2}{1}{Agent}
   \depedge{2}{4}{Price}
   \depedge{10}{12}{Price}
   \depedge[edge unit distance=1.5ex]{2}{10}{Booked object}
\end{dependency}}

\caption[]{Excerpt of the semantic structure for a sentence in the PORTMEDIA corpus.
Traditional dependency notations are used: the head segment points to the argument segment, where segments are shown with boxes (arrows link segments, not words~!). The semantic class assigned to each head segment is shown in bold below the translated text. }
\label{fig:exSimple}
\end{center}
\end{figure*}

Besides the gold annotation, a silver annotation of the whole MEDIA dataset was also provided. It was generated automatically after a pipeline of syntactic analysis, semantic role labelling and extraction of semantic frames. I implemented the whole pipeline as well as the annotation tool used by the annotators. I also collaborate to the creation of a Bayesian model for inferring these annotations~\citep{lorenzo-etal-2013-unsupervised}.

\subsection{The French Emospeech Corpus}
The French Emospeech corpus was already introduced in Section~\ref{ss:nluemospeech}.  As described in~\citep{rojas-barahona2012building}, to collect Human-Game dialog data, we developed a Wizard-of-OZ (WOZ)
interface using the MITRE Dialog Toolkit Midiki~\citep{Burke2003}. 
Midiki, is a portable toolkit for building
dialogue managers in Java. It implements the
information-state model of dialogue
\citep{TraumLarsson2003} where in essence,
the information state models the progression of
dialog while update rules formalise the way that
information state is changed as the dialog
progresses. 

We first extended Midiki to support a multi-agent architecture and the
configuration from a relational database. We then used this extended
Midiki (i) to develop a rule-based dialog system for the MP game and
(ii) to implement two Wizard-of-OZ interfaces for data collection: the
free- and the semi-automatic WOZ interface.

The {\it free WOZ interface} aims to simulate mixed-initiative dialogs
by allowing the wizard to chat with the player as she moves around the
game while simultaneously storing all interactions in a database. A
virtual dialog manager ensures that the wizard respects the game
logic, starting the appropriate subdialogs at the appropriate place in
the virtual world. In this setup, the interactions between the wizard
and the player simulate a direct Human-Human dialog in the context of
the MP game.

In contrast, {\it the semi-automatic wizard} favours system-driven
dialogs by connecting the Wizard not only with the player and the game
but also with the rule-based dialog manager (Figure~\ref{fig:midikiWOZ}). This dialog manager supports the Wizard by automatically
interpreting the player's input and selecting a possible response. As
the Wizard interacts with a player, she can then either accept the
response suggested by the rule-based dialog manager (if this response
is appropriate) or enter a different response (whenever the response
suggested is incorrect or inappropriate).

Figure~\ref{f:chatarchi} shows the architecture of the WOZ interface:
the dialog manager (either Midiki or a virtual DM), the MP game, the
Wizard of Oz interface and an automatic speech recognition module
(ASR) \footnote{Although the Wizard Framework supported both speech
  and written input, we did not record speech in our first
  experiments. All data is therefore written data.}  communicate
together within the Open Agent Architecture
(OAA)~\citep{oaaCheyer}. The WOZ interface is implemented as a web
service and all interactions are logged into a relational database.

\begin{figure}[htbp!]
\center
\includegraphics[scale=0.43]{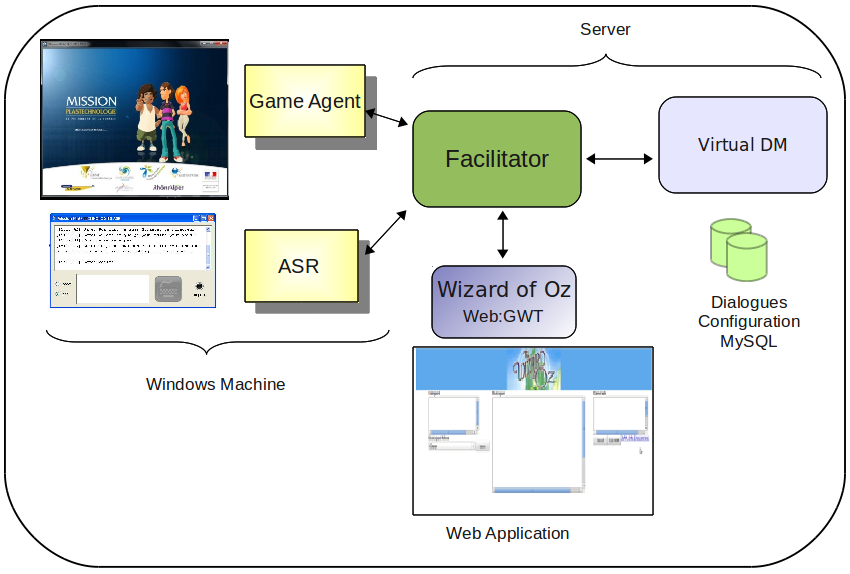}
\vspace{-2mm}
\caption{General Architecture for the Wizard of OZ experiments: modules are implemented as agents within the Open Agent Architecture.} 
\label{f:chatarchi}
\end{figure}

To support data collection for different game scenarios, we also
developed a Dialogue Configuration Tool that permits defining for each new game the
information that is relevant for the dialog, namely, which
characters are present in the game; which goals are being pursued at
each step in the game; and which subdialogs are being conducted in
which order during the game, between which characters and to achieve
which goal.

\begin{figure*}[htbp]
\begin{center}
\begin{tabular}{|l|r|r|r|r|r|r|r|}
\hline
&Subjects & Dialogs & Uttces & Tokens & Player U. & Player Tokens & Player Token Types \\\hline
Semi-Aut. &  40 & 591 &4874 &77854&1321&12901&1427\\
Free &  50 & 658  &5580 &90655&2288&18712&1542\\\hline
Total & 90 & 1249 & 10454 & 168509 & 3609 & 31613 & 2969\\\hline
\end{tabular}
\end{center}
\caption{Data collected}\label{tab:data}
\end{figure*}

Data collection for the MP game proceeded in two steps. First, a
native French speaker played the wizard using the semi-automatic WOZ
with 40 subjects. Next, three groups of students from the Language and
Communication Erasmus Mundus Master in Nancy collected dialogs using
the free WOZ. The results are shown in Table \ref{tab:data}.

Dialog length varies between 78 and 142 turns with an average length of
106 turns per dialog. Expert players completed the game in around 50
minutes in average while novice players took between 1 and 1.5 hour.

\begin{figure}[ht!]
\center
\includegraphics[scale=0.30]{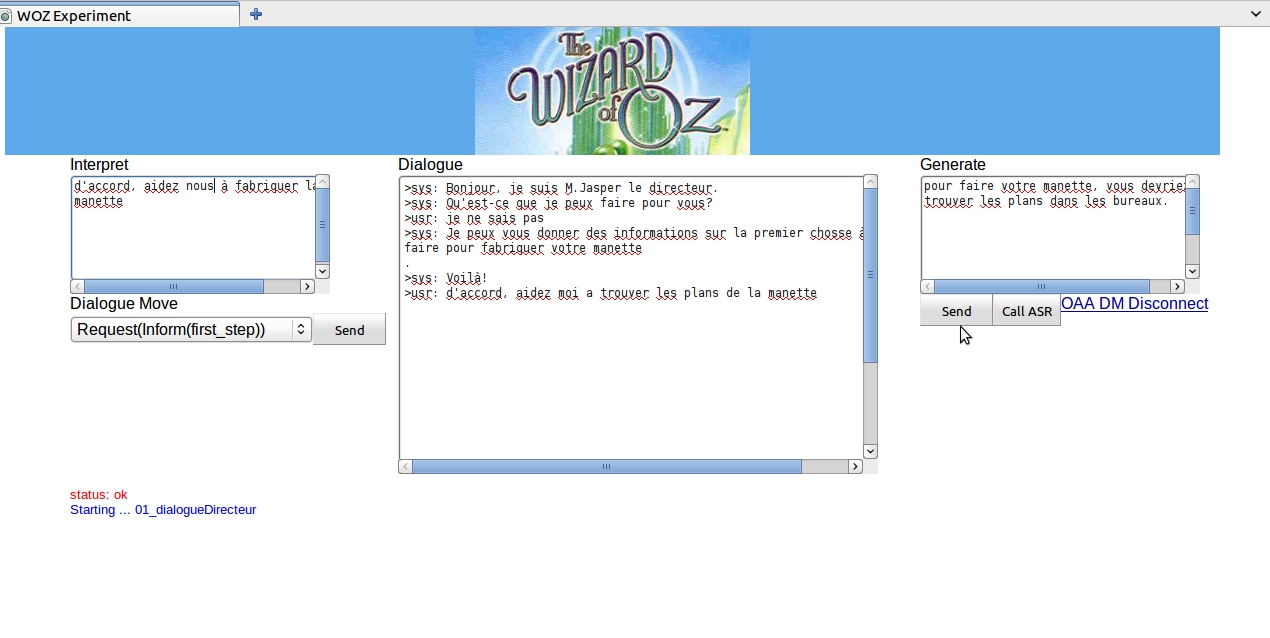}
\caption{Wizard of OZ Graphical User Interface (GUI).It is split in three parts: the interpretation, the dialog and the generation. 
The wizard receives the input sentence at the left-side, she can see the whole dialog in the centre, and she edits the generated utterance at the right side. She can introduce also the dialog move associated
to the input and output sentences. } 
\label{f:wozgui}
\end{figure}

After configuration of the WOZ tool using the Dialog Configuration Tool mentioned in the preceding section, the free-WOZ was also used by Master students from Rennes University for collecting dialogs in a game simulating the visit of an exhibition on Alice in Worderland.  In
this way, they collected 25 dialogs using Lewis Carroll's subrealist.

\subsection{Annotating Posts Following Cognitive Behavioural Therapy Principles}
\label{ss:cbt}
This work was published in~\citep{rojas-barahona-etal-2018-deep}. The main goal of annotations based on \ac{CBT} was to develop the understanding component of a health assistant for preventive intervention of mental health. The corpus consists of $500$K written posts that users anonymously posted on the Koko platform\footnote{\url{https://itskoko.com/}}. This platform was based on the peer-to-peer therapy proposed by~\citep{mosp15}. In this set-up, a user anonymously posts their problem (referred to this as the \textit{problem}) and is prompted to consider their most negative take on the problem (referred to this as the \textit{negative take}). Subsequently, peers post responses that attempt to offer a re-think and give a more positive angle on the problem. Initially, any first-time Koko user would be given a short introductory tutorial in the art of 're-thinking'/'re-framing' problems (based on CBT principles), before being able to use the platform; this however changed over time, as the age group of the users decreased, and a different introduction, emphasizing empathy and optimism, was used in relation to suggesting helpful responses (less CBT-based than the 're-thinking'). Some of the data annotated in this study was drawn from this later phase.  When first developed, this framework was shown to be more efficacious than expressive writing, an intervention that has been shown to improve physical and emotional well-being~\citep{mosp15}. Since then, the company has developed an app that has collected a very large number of posts and associated responses. In this work we only focus on analysing the posts. Figure~\ref{fig:koko} gives an example of an annotated post. 

We draw from principles of Cognitive Behavioural Therapy (CBT) to define the ontology.  CBT is derived originally from Beck's Cognitive Therapy model theory \citep{beck76,brse79} which says that our emotions and behaviour are influenced by the way we think and by how we make sense of the world.  Thus, if the patient changes the way he or she thinks about their problem that will in turn change the way he or she feels and behaves.
A major underlying principle of CBT is the idea of cognitive distortion, and the value in challenging this.  In CBT, patients are helped to test their assumptions and views of the world in order to check if they fit with reality. When patients learn that their perceptions and interpretations are distorted or unhelpful, they then work at correcting them.  Within the realm of cognitive distortion, CBT identifies a number of specific self-defeating thought processes, or thinking errors. There is a core of around 10 to 15 thinking errors, with their exact titles having some fluidity.  A strong component of CBT is teaching the client to be able to recognise and identify the thinking errors themselves, and ultimately discard the negative thought process, and 're-think' their problem.  

We consider the first step that a machine should be able to perform is to adequately decode these 'thinking error' concepts, along with identifying the key emotion(s) expressed, and situational context, within a particular presented problem.  Therefore, our ontology consists of \emph{thinking errors}, \emph{emotions}, and \emph{situations}. 
\begin{figure}
\centering
\includegraphics[width=80mm,trim={0 0 0 0},clip]{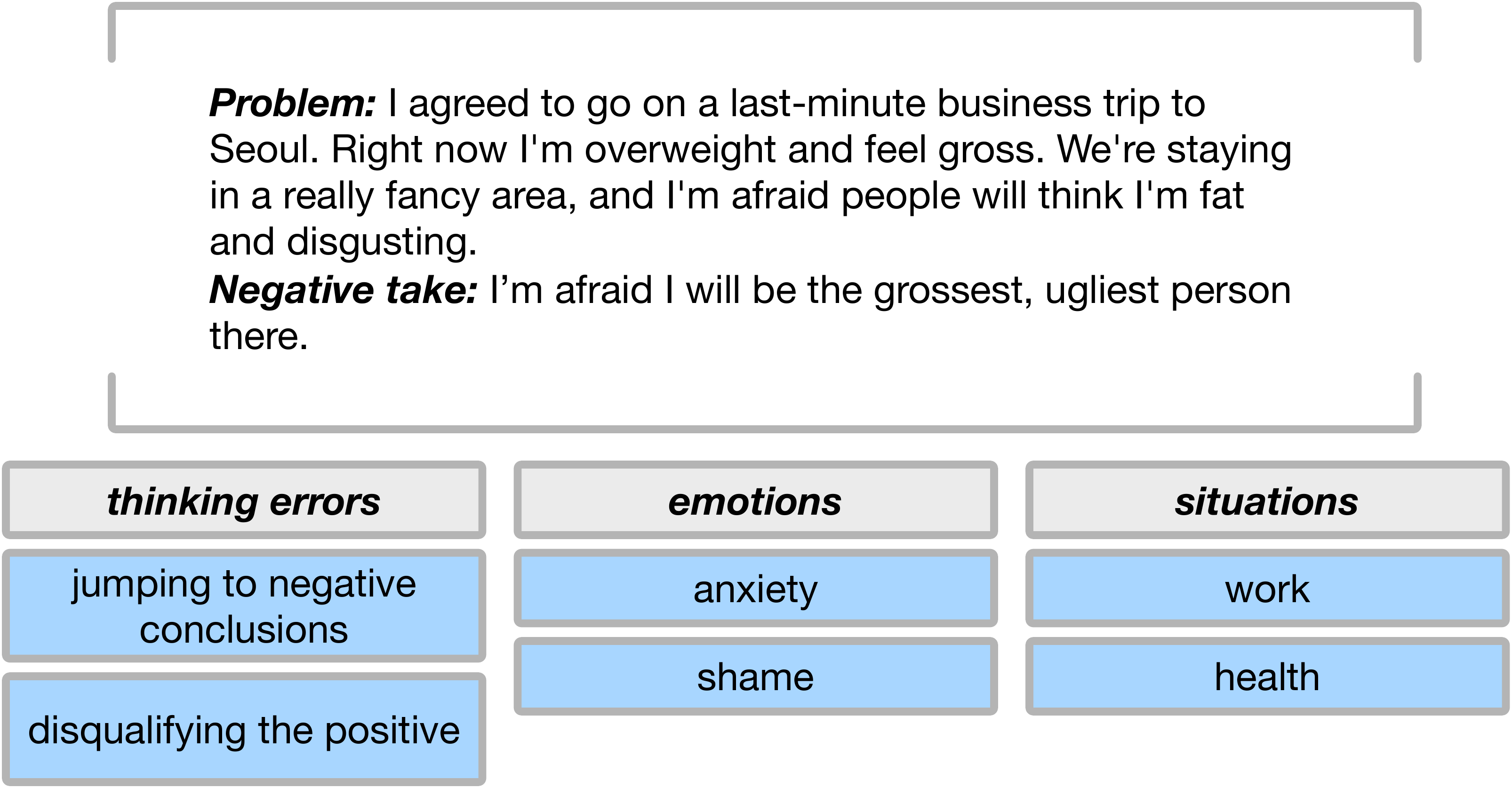}
\caption{An example of an annotated Koko post.}
\label{fig:koko}
\end{figure}

\subsection{Conversational Question Answering with Rewriting}
I lead the extension of the corpus CoQA with question rewriting as explained in Section~\ref{s:qr}.
\subsection{Conversational Question Answering in French}

I proposed to extend the French corpus Calor with sequence of questions and answers. The annotations were done by the University of Aix-Marseille.  This is a joint work with Geraldine Damnati and Frederic Bechet. This work has been published in~\citep{bechet2022calor}.

Calor-Dial is an enriched version of the Calor corpus~\citep{MarzinottoL18-1159}, collected from French encyclopedic data in order to study Information Extraction on domain specific data. The corpus was initially annotated in semantic Frames (Calor-Frame~\citep{bechet:hal-01780348}) and enriched with a first set of questions for Machine Reading Question Answering (Calor-Quest~\citep{bechet2019calor}). Calor-Dial addresses the scope of conversational Question Answering. The main originality is that different types of questions are annotated, including more challenging configurations than in classical QA corpora. 

\subsection{Conversational QA grounded in Wikidata}
I contribute to the creation of the corpus \ac{KGConv} and I lead the work from the Orange side, as part of the research project \ac{DIANA}. This work is in colaboration with Claire Gardent (CNRS) as part of the partnership of the ITN-European project NL4XAI.  This corpus is presented in Section~\ref{s:KGConv}.

\section{Dialogue Frameworks}
During the course of my research career, I was directly involved in the development of dialogue frameworks. Starting from AdaRTE~\citep{rojas2009adaptable} during my PhD studies to the most widespread one, PyDial~\citep{ultes_pydial_2017}. In this Section I will briefly present PyDial.

\subsection{PyDial}

PyDial is an open-source end-to-end statistical spoken dialogue toolkit developed by the University of Cambridge~\citep{ultes2017pydial}. It provides implementations of statistical approaches for all dialogue modules: NLU. DST, Policy, NLG. The term statistical means that: (i) the framework preserves the confidence probability of each module, thus each module outputs are the N-Best list of hypotheses; (ii) Deep learning models can be easily integrated in it; (iii) The framework implements POMDP dialogue systems. Thus, reinforcement learning dialogue management is fully supported, with the user simulator and reward estimators. Moreover, it has been extended to support multiple domains.  It offers easily extensible to other domains and or specialised module implementations. It also offers domain-independent implementations of the dialogue modules (see Figure~\ref{fig:sds}). The toolkit is available for download under the Apache 2.0 license.

\begin{figure}[ht]
\begin{center}
  \includegraphics[width=\linewidth]{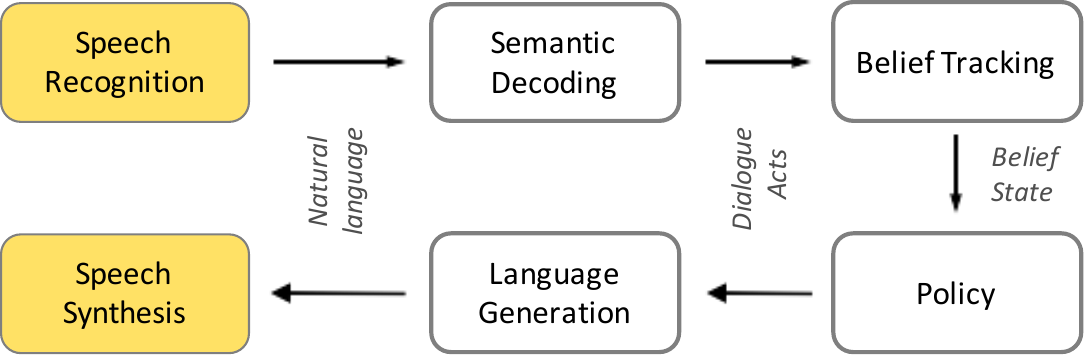}
  \caption[Spoken Dialogue System]{Architecture of a modular Spoken Dialoug System.}
  \label{fig:sds}
  \vspace{-0.5cm}
\end{center}
\end{figure}

\subsection{Dialport}
Dialport~\citep{lee-etal-2017-dialport} is an academic dialogue portal to collect large amounts of real user data for spoken
dialog systems (SDS). Sophisticated statistical
representations in state-of-the-art SDS, require
large amounts of data, which is difficult to obtain by academic teams.
With one central portal, connected to many different
systems, the task of advertising and affording
user access can be done in one centralised place
that all systems can connect to. DialPort provides
a steady stream of data, allowing system creators
to focus on developing their systems. The portal
decides what service the user wants and connects
them to the appropriate system which carries on a
dialog with the user, returning control to the portal
at the end.
Dialport was connected to the Cambridge restaurant information system, which helps users find a restaurant in Cambridge,UK based on the area, the price range
or the food type. The connection is done through an \ac{API} that connects Dialport to PyDial.
An assesment of DialPort is presented in~\citep{lee2019assessment}, to summarise, $28.8\%$ of chatbot utterances were non-understanding recovery turns, such as ``can you please rephrase that?". $62.85\%$ of the times DialPort successfully recommended users and $78.40\%$ it correctly directed users to the appropriate system.

\subsection{Conversational Search for General Knowledge}

\ac{CS4GK} is  a spoken conversational question answering proof of concept that is able to answer questions about general knowledge from Wikidata\footnote{\label{wiki}\url{https://www.wikidata.org}}~\citep{rojasbarahona2019spoken}. The dialogue component does not only orchestrate various components but also solve coreferences and ellipsis.

The architecture of the proposed system consists of a speech-processing front-end, an understanding component, a context manager, a generation component, and a synthesis component. The context manager provides contextualised mediation between the dialogue components and several question answering back-ends, which rely on data provided by Wikidata. Interaction with a human user is achieved through a graphical user interface (GUI). Figure~\ref{f:cs4gk-achictecture} depicts the components together with their interactions.

\begin{figure}[h!]
	\centerline{\includegraphics[width=\linewidth]{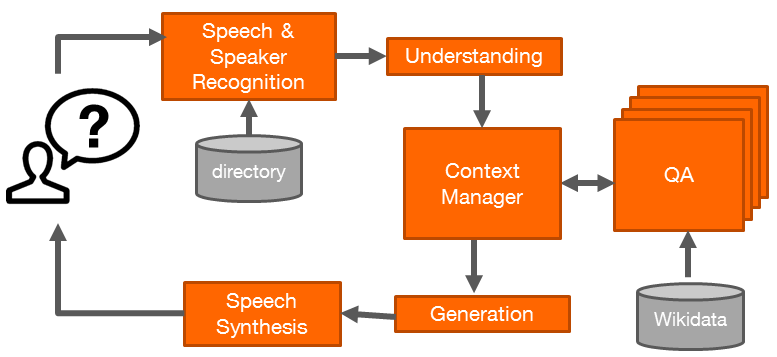}}
	\caption{\small\noindent High-level depiction of the proposed spoken conversation question answering system. Arrows indicate data flow and direction.}
	\label{f:cs4gk-achictecture}
\end{figure}

\chapter{Scientific Project}
\label{c:project}

After the breakthrough of ChatGPT~\citep{ouyang2022training}, Large Language Models have been widely used for distinct daily activities such as summarisation, translation, sentiment analysis, question answering, redaction, code-generation, etc. Nevertheless, it is not clear how these models can be used to solve complex decision-making tasks, such as task-oriented dialogue.  
Although promising approaches have recently emerged~\citep{wei2022chain,yao2023tree, yao2022react}, utilizing \ac{LLMs} to solve complex tasks is not straightforward. In the case of task-oriented dialogue, one important point concerns the lack of control (Section~\ref{ss:control}). Although the conversation is fluid and pleasant is it following the necessary steps to solve the task? Are these steps validated by experts? Are these steps correctly grounded in the World knowledge?
These questions bring us to an important issue the evaluation (Section~\ref{ss:eval}). Can we estimate whether the task was accomplished, and whether the sequence of selected actions was indeed optimal?  
\ac{RAG}~\citep{mao-etal-2021-reader,asai-etal-2022-evidentiality}, can be used to retrieve useful information that can be injected to the LLM as a prompt. This is a way of grounding the LLM in "factual" information. However, is there a way to be sure the retrieved information is indeed factual?(Section~\ref{ss:rag}) Moreover, the effort of prompt-engineering can not be neglected. Are we coming back to hand-crafted solutions by hand-crafting prompts? Would it be difficult to maintain and to keep these prompts up-to-date? 

This scientific project proposes first to study recent LLM-based reasoning approaches for task-oriented dialogue, providing a rigorous evaluation in terms of task-completion and success rate. We can also, inspired by the seminal evaluation framework Paradise~\citep{walker-etal-1997-paradise}, think in ways to find the correlation between task-completion and user satisfaction. For instance, unlike the study presented in Paradise, in which  at that time users were not enjoying long conversations with repetitive systems, maybe users now really enjoy talking to \ac{LLMs}. However, a strong indicator of poor performance might be to confirm that users usually need to call again because despite having a fluid and natural interaction, their problems were not solved at all. 

Multimodality is another interesting research topic we will discuss in this proposal. Concerning dialogue, emergent approaches are \ac{VQA} and speech-aware emotion detection. The first one can be used in \ac{TOD} wherein heterogeneous sources of knowledge are identified. The second, can be used to personalise dialogue according to users' mood detected by speech analysis.


\section{LLMs for Task-Oriented Dialogue}

Recently synergising reasoning and acting in large language models (ReAct)~\citep{yao2022react} has shown promising results employing few-shot prompting in a LLM with a sequence of thoughts, actions, and observations. \ac{LLMs} are indeed capable of performing complex tasks. Thoughts refer to the internal reasoning that decompose a problem into sub-problems. For example, if the model is asked the age of Barack Obama's wife power of 3, the thoughts might be as follows.  First, I need to find out who Barack Obama's wife is (by calling a question-answering API such as Google Search). Second, I need to find the age of Barack Obama’s wife (by calling a question-answering API). Finally, I will calculate the age power of 3 (by calling a calculator API). Examples have shown promising results for tasks such as multi-hop question-answering, WebShop, and a textual version of a butler in a virtual environment, ALFWorld: the butler can report on actions he has taken to solve a given task in a kitchen~\citep{yao2022react}. 


We need to study the recent state-of-the art in task-decomposition utilizing \ac{LLMs}\citep{wei2022chain,yao2023tree, yao2022react}. 
We can also explore Algorithm Distillation~\citep{laskin2022context}, wherein the logs of cross-episodic events generated during the learning process of a \ac{RL} algorithm, are used to feed in a LLM, which as consequence can learn the optimal strategy. Since the state-of-the-art is moving impressively fast, this proposal is open to upcoming approaches.

\subsection{Benchmarcks and Evaluation}
\label{ss:eval}
 We will propose an evaluation framework that takes into account long-term memory, beyond one dialogue session~\citep{xu-etal-2022-beyond}. For evaluating LLMs an evaluation framework that is both model-agnostic and domain-independent is suitable. Therefore, we should think in ways to detect the conversation goal together with performance indices, with or without humans in the loop.
 In an initial state we can compare our previous results presented in Section ~\ref{ss:hil} on hierarchical reinforcement learning for dialogue~\citep{cordier-etal-2022-graph, cordier2023few} with \ac{LLMs} that follows ReAct and Algorithm Distillation by using the metrics introduced in ConvLab~\citep{zhu2020convlab}. Therefore, we can start this study by using the dataset MultiWoz, in which the user goal is formally defined and provided. This corpus also includes the instructions given to annotators during data collection. Later, we can move to complex tasks in which the user goal is not provided. We can also study more realistic cases such as commercial or technical-support systems.  We will perform human evaluation to perceive their satisfaction with the system. As in Paradise~\citep{walker-etal-1997-paradise} and in~\citep{rojas-barahona2012should}, we would like to make a correlation study between objective metrics (indices of performance that can be computed automatically) and subjective metrics (e.g. user satisfaction). 
 This framework can be extended to even more complex interactions such as in multi-modal dialogue systems (Section~\ref{s:multimod}).
 
\subsection{Interpreting LLMs}
\label{ss:interpretation}
After evaluating \ac{LLMs} performance for task-oriented dialogue, we can use model agnostic black-box interpretability methods~\citep{10.3389/frai.2023.1220476} to understand how these models are able to solve complex tasks.  Where in the LLM architecture this behaviour is being produced? It is worth noting that these methods can be applied only to Open-Source \ac{LLMs}, in which we can have access to the model (e.g., Llama~\citep{touvron2023llama}, Falcon~\citep{penedo2023refinedweb}).
Interpretability might give us an insight of how we can correct \ac{LLMs} to avoid undesirable behaviour (e.g. forgetting an important instruction, biases). Methods to correct \ac{LLMs} are introduced in the following Sections.

\subsection{Retrieval Augmented Generation}
\label{ss:rag}
One way to find grounded knowledge to feed into  \ac{LLMs} is through information retrieval~\citep{mao-etal-2021-reader,asai-etal-2022-evidentiality}. Despite these techniques rank documents according to their relevance, an important aspect is factuality~\citep{thorne-etal-2018-fever}. Another aspect is that the sources of knowledge are heterogeneous: they concern not only documents, but also images, recorded interactions, knowledge graphs, tabular data, among others. Once we have retrieved crucial information there are several ways to inject this information into \ac{LLMs}. The most adopted one is prompt engineering, but controlling decoding might be more interesting in terms of scientific research.
 
\subsection{Controlling Decoding} 
\label{ss:control}
There are distinct sampling mechanisms used for decoding such as greedy ($\mathcal{T}=0$ in Equation~\ref{eq:sam}), beam search, top-k, nucleus or penalised sampling~\citep{fan-etal-2018-hierarchical,holtzman2019curious,keskar2019ctrl}. An emergent method to control decoding is by adjusting sampling weights in beam search~\citep{ghazvininejad-etal-2017-hafez}.

Let $p_\theta$ be a pre-trained generative language model which has learned the distribution over token sequences by optimising:
\begin{equation}
	\mathcal{L} = -\sum_{t}{\log{p_\theta(x_t|x_{<t})}}
\end{equation}
The next token can be sampled by applying Softmax with temperature $\mathcal{T}$ because the final decoder layer predicts logits $o$, over the vocabulary space:
\begin{equation}
	\label{eq:sam}
	p_i \varpropto \frac{\exp{(o_i/\mathcal{T})}}{\sum_j{\exp{(o_j/\mathcal{T})}}}
\end{equation}
Beam search is a breadth-first search algorithm which explores the $\beta$ tokens which best scores at each level.
Then, the likelihood of sampling for the next token $x_{t+1}$ at step $t$ can be augmented by a scoring function:
\begin{equation}
	score(x_{t+1},b_t)= score(b_t) + \log{p(x_{t+1})+ \sum_i{\alpha_i f_i(x_{t+1})}}
\end{equation}
Where the log-likelihood predicted by the pre-trained language model is defined by $\log{p(x_{t+1})}$. The accumulated score of the already-generated words in the current beam state $b_t$ is $score(b_t)$. A set of feature functions that define the preferences is $f(.)$, which can be a binary classifier that predicts whether a sample is from the true data distribution~\citep{grover2019bias}. Finally, $\alpha_i$ are the associated weights that work like "control knobs". Therefore, the classifier can be used to constrain factuality by predicting whether the token is grounded in knowledge or not. This is an interesting path that can be further explored.

Another interesting approach is Chain of hindsight~\citep{liu2023chain}, in which pairs of (answer, feedback) are provided as input to the model during fine-tuning. Thus, the model learns to condition the generation on feedbacks during training. 

\section{Multimodal task-oriented dialogues}
\label{s:multimod}
 Generative models were also trained for process multiple modalities~\citep{Antol_2015_ICCV,zhu2016visual7w,srivastava2021visual}, supporting image captioning and \ac{VQA}. After the release of multi-modal \ac{LLMs}, the borders between domains, tasks and modalities have been vanished. 
One can think in \ac{VQA} applications that interact with elders, visual impaired~\citep{Chen_2022_CVPR} or more broadly patients or practitioners in the medical domain.
Efforts for ethical \ac{AI} can not be diminished before putting \ac{VQA} to interact with sensitive population. This is particularly true in Europe, after the AI regulation~\citep{hacker2023regulating}. Therefore, important research areas concern interpretability and evaluation of multi-modal \ac{LLMs} to correct model biases that might produce any harm. 
Generally, the datasets used for training these large models were built without including minorities or people at risk. Therefore, these models can not respond to their needs. A rigorous study to evaluate LLMs, involving these individuals must be carried out. Interpretability methods (Section~\ref{ss:interpretation}) can help us to understand biases deep inside the model and to provide insights to correct them. This can be done in public models for image captioning or \ac{VQA} such as OFA~\citep{wang2022ofa}. 

Mulitmodality might also be highly related to \ac{RAG} in both text-to-image and image-to-text models~\citep{yasunaga2023retrieval}, wherein the information is available as images, audio or video (Section~\ref{ss:rag}).
Finally,  multi-modality  also involves speech, in this proposal I talk about ways to adapt task-oriented dialogues that use either \ac{RL} or \ac{LLMs} for policy learning and that take into account the speech signal for emotion detection .

\subsection{Exploiting weak speech signals in the reward function}
Treating speech and text together is a research topic widely studied in \ac{SLU}. An interesting research path would be to go beyond intention recognition up to response generation.  We worked with the PhD candidates Leo Jacqmin, Lucas Druart and other researchers in a cascade approach that integrated the ASR (i.e. Whisper)\citep{radford2022robust} with a generative model for \ac{DST} and we participate to the challenge DSTC-11~\citep{jacqmin2023olisia}\footnote{\url{https://storage.googleapis.com/gresearch/dstc11/dstc11_20221102a.html}}. \ac{RL} can be used for learning the strategy (Section~\ref{c:intross:ssds}). However, it assumes there is a reward function. In dialogue systems the reward signal is scarce, because it is unbearable for a user to send a satisfaction signal at each dialogue turn. For fluidity, the reward function is asked at the end of the conversation, which produces a scarce signal, complicating the task of policy learning. Thus, the policy spectrum will have a large variance from very poor to very good.  

We explored already two ways of solving this problem: (i)predicting the user satisfaction at each dialogue turn by using the interaction quality as reward signal (Section~\ref{pomdp}) and (ii) using imitation learning to guide the policy learning (Section~\ref{c:conts:dm}). 
  
One can think in improving the reward scarcity by exploiting the information in the speech signal. Studying weak signals in the speech, such as the emotion, could be a way to improve the reward signal. Emotion detection is important for developing conversational systems that could adapt better to the users' needs, improving as consequence the user satisfaction.
For instance, an early detection of distress would entail changing the dialogue strategy to quickly solve the misunderstanding.
Emotion detection involves an active research community producing datasets build from actors playing the emotion~\citep{burkhardt2005database,schroder2007should}, from wizard of oz or more recently from more natural content~\citep{zadeh2018multimodal,zadeh2020cmu,dhall2012collecting,scheidwasser2022serab}. Neural methods for emotion detection have been proposed in~\citep{trigeorgis2016adieu}, since there are not many dataset for training deep models from scratch transfer learning or distillation from other modalities have been also proposed~\citep{pepino2021emotion, albanie2018emotion}.

The main research questions behind this study will be:
\begin{itemize}
    \item Concerning the emotion detection: we need to study domain adaptation for emotion recognition to better exploit the available cross-domain datasets to dialogue.
    
    \item Combine multi-modality to produce the reward, one can think in using speech signal and other domain agnostic metrics, such as counting the number of repetitions, counting the number of dialogue turns, measuring the misunderstandings, detecting the sentiment from text.
    \item Evaluate whether the emotion-based reward is suitable for policy learning
    \item Compare RL with LLM-based policies (e.g. algorithm distillation~\citep{laskin2022context}, Chain of Hindsight~\citep{liu2023chain}).
\end{itemize}


\chapter{Conclusion}
\label{c:conclusion}
I presented in this dissertation a selected number of contributions I made to the areas of task-oriented dialogue systems, conversational question answering and graph embeddings. 

The contributions to task-oriented dialogue were in the fields of \ac{NLU}, \ac{SLU} and \ac{DM}. Particularly, I explored classical machine learning techniques for \ac{NLU}, convolutional and recurrent neural networks to deal with noisy inputs for \ac{SLU}, as well as data-augmentation techniques. Although not mentioned in this work, I am currently supervising with Benoit Favre from the University of Aix-Marseille a PhD thesis on \ac{DST}, recently we competed in the challenge DSTC-11 and we were awarded the first and second place~\citep{jacqmin2023olisia}\footnote{\url{https://storage.googleapis.com/gresearch/dstc11/dstc11_20221102a.html}}. Regarding the \ac{DM}, I explored Inverse Reinforcement Learning, Deep Reinforcement Learning, Imitation Learning and Structured Policy Learning. 

The contributions to conversational \ac{QA} regard the annotation of existing datasets with information about ellipsis and coreferences, and with question rewriting to transform in-context questions into out-of-context questions. The generative models released with these annotations for the tasks of answer extraction, question generation and question rewriting; as well as the models implemented for predicting ellipsis and coreferences were also presented. I also briefly introduced the corpus \ac{KGConv} grounded in Wikidata. Moreover, I presented our work on graph embeddings in the hyperbolic space. Finally, I summarised the released resources such as datasets and frameworks, in which I contributed to their creation, annotation and development.

This document consolidates my own contributions as young researcher, the work of two PhD candidates supervised jointly with academics under the \ac{CIFRE} convention. I could also collaborate with academics in two \ac{ITN} projects: \ac{COBRA} and \ac{NL4XAI}. As head of the project \ac{DIANA} during $5$ years, I could also define the main workpackages of the project and supervise a dynamic team of researchers, developers, students in internship and apprenticeship. 

I would like to focus my future research in proposing a framework to evaluate LLMs performance in complex task-oriented dialogue. Moreover, I would like to explore interpretability, retrieval augmentation and semantically control to understand LLMs internally, to support grounding and to generate factual information. I would like to explore recent reasoning LLMs approach (e.g., ReAct, algorithm distillation, etc.) to study LLMs capabilities to make long-term decisions. It is wort noting that these research paths also cover multi-modal interactions.



\bibliographystyle{acl}

\bibliography{dialogs, dlearning, colingbiblio, genapprs, hyperbolic, imitation, gnn, anthology, mywork, convqa, cbt}

\end{document}